\pdfoutput=1
\documentclass[10pt,letterpaper]{article}

\usepackage[letterpaper,left=1.2in,right=1.2in,top=1in,bottom=1in,footskip=0.45in]{geometry}
\usepackage{iftex}
\ifPDFTeX
  \usepackage[T1]{fontenc}
  \usepackage[utf8]{inputenc}
  \usepackage[helvratio=0.9]{newtxtext}
  \usepackage{newtxmath}
\else
  \usepackage{newtxtext,newtxmath}
  \usepackage{fontspec}
\fi
\usepackage{amsmath}
\usepackage{graphicx}
\usepackage{booktabs}
\usepackage[table]{xcolor}
\usepackage{float}
\usepackage{array}
\usepackage{enumitem}
\usepackage{caption}
\usepackage{microtype}
\usepackage{fancyvrb}
\usepackage{titlesec}
\usepackage{fancyhdr}
\usepackage[most]{tcolorbox}
\usepackage{tikz}
\usetikzlibrary{positioning,arrows.meta,calc}
\usepackage[numbers,sort&compress,square]{natbib}
\usepackage[colorlinks=true,linkcolor=ohblue,citecolor=ohblue,urlcolor=ohblue]{hyperref}
\hypersetup{pdftitle={Openhagen Worlds: Calibrated Environments for Measuring and Training Scientific Discovery Agents},pdfauthor={Alexander Boesgaard Lorup},pdfsubject={Openhagen technical report}}
\usepackage[capitalise,noabbrev,nameinlink]{cleveref}

\definecolor{ohink}{HTML}{1D1D1B}
\definecolor{ohmuted}{HTML}{6B6A66}
\definecolor{ohcap}{HTML}{55544F}
\definecolor{ohrule}{HTML}{BDBCB6}
\definecolor{ohpanel}{HTML}{FAFAF7}
\definecolor{ohpanelrule}{HTML}{E3E2DD}
\definecolor{ohgroup}{HTML}{F1F0EC}
\definecolor{ohblue}{HTML}{1F5FAD}
\definecolor{ohbluefill}{HTML}{EEF4FC}
\definecolor{ohteal}{HTML}{0F766E}

\titleformat{\section}{\sffamily\bfseries\large\color{ohink}}{\thesection}{0.75em}{}
\titleformat{\subsection}{\sffamily\bfseries\normalsize\color{ohink}}{\thesubsection}{0.75em}{}
\titleformat{\paragraph}[runin]{\normalfont\bfseries}{}{0pt}{}
\titlespacing*{\section}{0pt}{16pt plus 3pt minus 2pt}{7pt}
\titlespacing*{\subsection}{0pt}{11pt plus 2pt minus 2pt}{5pt}
\titlespacing*{\paragraph}{0pt}{5pt plus 1pt minus 1pt}{0.7em}
\setlist{itemsep=2pt,topsep=3pt,parsep=0pt,leftmargin=2.2em}
\renewcommand{\arraystretch}{1.12}
\arrayrulecolor{ohrule}

\fancypagestyle{firstpage}{%
  \fancyhf{}%
  \fancyfoot[L]{\sffamily\footnotesize{\color{ohteal}\rule{5pt}{5pt}}\hspace{5pt}{\color{ohmuted}An Openhagen technical report}}%
  \fancyfoot[R]{\small\color{ohmuted}\thepage}}

\newcommand{\base}{\mbox{Qwen3.5-9B}}
\newcommand{\ours}{\mbox{Concise-9B}}
\newcommand{\qbig}{\mbox{Qwen3.8-27B}}
\newcommand{\flash}{\mbox{DeepSeek V4.1 Flash}}
\newcommand{\nar}{\mbox{Narrow-9B}}
\newcommand{\fin}{\mbox{Broad-9B}}
\newcommand{\yes}{\ensuremath{\bullet}}
\newcommand{\no}{\ensuremath{\circ}}
\newcommand{\dci}[3]{\ensuremath{#1}~[\ensuremath{#2},\allowbreak\ \ensuremath{#3}]}
\newcolumntype{L}[1]{>{\raggedright\arraybackslash}p{#1}}
\newcommand{\grouprow}[2]{\rowcolor{ohgroup}\multicolumn{#1}{@{}l}{\textsc{\footnotesize #2}}\\}
\newcommand{\capt}[1]{\textbf{\color{ohink}#1}}

\newcommand{\capCutPct}{27}
\newcommand{\capCutPctMin}{23}
\newcommand{\capCutPctMax}{39}
\newcommand{\capCutRungMin}{L1}
\newcommand{\capCutRungMax}{H3}
\newcommand{\capLogged}{1886}
\newcommand{\capReaskN}{1290}
\newcommand{\capReaskK}{233}
\newcommand{\capReaskRate}{0.18}
\newcommand{\capThinkN}{596}
\newcommand{\capThinkK}{413}
\newcommand{\capThinkRate}{0.69}
\newcommand{\oraclePhysFinal}{0.91}
\newcommand{\oracleChemFinal}{0.95}
\newcommand{\qPhysFinal}{0.03}
\newcommand{\qChemFinal}{0.18}
\newcommand{\solFrontierK}{71}
\newcommand{\solFrontierN}{84}
\newcommand{\solFrontierRate}{0.85}
\newcommand{\solFrontierLo}{0.75}
\newcommand{\solFrontierHi}{0.91}
\newcommand{\flashPilotFrontierK}{32}
\newcommand{\flashPilotFrontierRate}{0.38}
\newcommand{\qPilotFrontierK}{7}
\newcommand{\oraclePilotFrontierK}{79}

\newcommand{\solMinusFlash}{\dci{+0.46}{+0.36}{+0.57}}
\newcommand{\solFlashWins}{39}
\newcommand{\solFlashLosses}{0}
\newcommand{\solMinusQ}{\dci{+0.76}{+0.67}{+0.85}}
\newcommand{\solMinusOracle}{\dci{\ensuremath{-}0.10}{\ensuremath{-}0.18}{\ensuremath{-}0.01}}
\newcommand{\solHarderMinusOracle}{\dci{\ensuremath{-}0.02}{\ensuremath{-}0.14}{+0.10}}
\newcommand{\solLongMinusOracle}{\dci{\ensuremath{-}0.17}{\ensuremath{-}0.29}{\ensuremath{-}0.07}}
\newcommand{\solLthreeK}{8}
\newcommand{\solLthreeN}{14}
\newcommand{\oracleLthreeK}{13}
\newcommand{\solLthreeMinusOracle}{\dci{\ensuremath{-}0.36}{\ensuremath{-}0.64}{\ensuremath{-}0.14}}
\newcommand{\solRungKMin}{12}
\newcommand{\solRungKMax}{14}
\newcommand{\solRungN}{14}

\newcommand{\solCoreN}{45}
\newcommand{\flashPilotCoreK}{31}
\newcommand{\solPhysK}{17}
\newcommand{\solPhysN}{20}

\newcommand{\solChemK}{15}
\newcommand{\solChemN}{15}

\newcommand{\flashPilotPhysK}{1}
\newcommand{\flashPilotChemK}{6}
\newcommand{\solEpisodes}{164}
\newcommand{\solTokensPerDecision}{1,872}
\newcommand{\solModel}{\solname{} via the Codex CLI's ``ultra'' preset (server-reported reasoning effort xhigh), not the API protocol}
\newcommand{\solPilotScope}{one final instance ($o = 0$) per family and rung or level}
\newcommand{\fpPhysN}{320}
\newcommand{\fpPhysRounds}{16}

\newcommand{\fpPhysFlash}{0.13}
\newcommand{\fpPhysInvalid}{0.83}
\newcommand{\fpPhysCompleted}{0.73}
\newcommand{\fpPhysOracle}{0.90}
\newcommand{\fpPhysEmptyPct}{73}

\newcommand{\fpPhysMinusOracle}{\dci{\ensuremath{-}0.77}{\ensuremath{-}0.82}{\ensuremath{-}0.72}}
\newcommand{\fpPhysMinusQ}{\dci{+0.10}{+0.06}{+0.14}}
\newcommand{\fpPhysMinusConcise}{\dci{+0.12}{+0.08}{+0.16}}

\newcommand{\fpChemN}{360}

\newcommand{\fpChemFlash}{0.33}
\newcommand{\fpChemInvalid}{0.64}
\newcommand{\fpChemCompleted}{0.90}
\newcommand{\fpChemOracle}{0.95}
\newcommand{\fpChemEmptyPct}{93}

\newcommand{\fpChemMinusOracle}{\dci{\ensuremath{-}0.63}{\ensuremath{-}0.68}{\ensuremath{-}0.57}}
\newcommand{\fpChemMinusQ}{\dci{+0.15}{+0.10}{+0.20}}
\newcommand{\fpChemMinusConcise}{\dci{+0.29}{+0.24}{+0.35}}

\newcommand{\ladHoldSolMin}{0}
\newcommand{\ladHoldSolMax}{2}
\newcommand{\ladN}{8}
\newcommand{\ladHoldGateMin}{0.80}
\newcommand{\ladHoldGateMax}{1.00}
\newcommand{\ladStructSolMin}{7}
\newcommand{\ladStructSolMax}{8}
\newcommand{\ladHlFourSol}{8}
\newcommand{\ladHlFourN}{16}

\newcommand{\ladCcFourDiff}{\dci{\ensuremath{-}0.75}{\ensuremath{-}1.00}{\ensuremath{-}0.38}}
\newcommand{\ladCcFiveDiff}{\dci{\ensuremath{-}0.88}{\ensuremath{-}1.00}{\ensuremath{-}0.63}}
\newcommand{\ladHlFiveDiff}{\dci{\ensuremath{-}0.75}{\ensuremath{-}1.00}{\ensuremath{-}0.38}}
\newcommand{\audFailures}{35}

\newcommand{\audMalformed}{2}
\newcommand{\ccFiveVoneSol}{5}
\newcommand{\ccFiveVoneN}{8}
\newcommand{\audLong}{32}
\newcommand{\audStructure}{3}

\newcommand{\audFewerNotes}{29}
\newcommand{\audFullNotebook}{3}
\newcommand{\audFullFirstTurn}{51}
\newcommand{\tsHeld}{\dci{+0.120}{+0.046}{+0.194}}
\newcommand{\tsTrained}{\dci{+0.019}{\ensuremath{-}0.013}{+0.052}}
\newcommand{\tsHeldN}{216}
\newcommand{\tsTrainedN}{852}
\newcommand{\tsHeldBase}{0.394}
\newcommand{\tsHeldRtwo}{0.514}
\newcommand{\tsHeldFamN}{72}
\newcommand{\tsHeldCcBase}{24}
\newcommand{\tsHeldCcRtwo}{46}
\newcommand{\tsHeldMpBase}{47}
\newcommand{\tsHeldMpRtwo}{51}
\newcommand{\tsHeldAlBase}{14 of 72}
\newcommand{\tsHeldCluster}{\dci{+0.120}{+0.000}{+0.306}}

\newcommand{\tsGpqaRtwo}{179}
\newcommand{\tsGpqaN}{198}
\newcommand{\trConcise}{0.090}
\newcommand{\trNarrow}{0.103}
\newcommand{\trBroad}{0.148}
\newcommand{\trN}{837}
\newcommand{\hlConcise}{0.083}
\newcommand{\hlNarrow}{0.065}
\newcommand{\hlBroad}{0.167}
\newcommand{\trBN}{\dci{+0.045}{+0.019}{+0.072}}
\newcommand{\trBC}{\dci{+0.059}{+0.031}{+0.085}}
\newcommand{\hlBC}{\dci{+0.083}{+0.023}{+0.144}}
\newcommand{\invConcise}{0.090}
\newcommand{\invNarrow}{0.059}
\newcommand{\invBroad}{0.016}
\newcommand{\invBN}{\dci{-0.043}{-0.054}{-0.032}}
\newcommand{\invN}{2,181}
\newcommand{\gtConcise}{0.326}
\newcommand{\gtBroad}{0.804}
\newcommand{\gtInvConcise}{0.59}
\newcommand{\gtN}{46}
\newcommand{\gtBN}{\dci{+0.239}{+0.065}{+0.413}}
\newcommand{\gtBC}{\dci{+0.478}{+0.283}{+0.674}}
\newcommand{\elevenBC}{\dci{+0.034}{+0.009}{+0.059}}
\newcommand{\weakBN}{\dci{+0.026}{+0.007}{+0.046}}
\newcommand{\posMin}{0.014}
\newcommand{\posMax}{0.478}
\newcommand{\nPos}{ten}
\newcommand{\nNever}{five}
\newcommand{\neverMaxGain}{one}
\newcommand{\lcFamilies}{thirteen}
\newcommand{\lcOtherFamilies}{twelve}
\newcommand{\lcWorlds}{fourteen}
\newcommand{\lcRungs}{25}
\newcommand{\lcOracleMin}{0.83}
\newcommand{\lcOracleMax}{1.00}
\newcommand{\lcShortcutMax}{0.05}
\newcommand{\ccOracleMin}{0.78}
\newcommand{\ccOracleMax}{1.00}
\newcommand{\ccShortcutMax}{0.23}
\newcommand{\lcProbedRungs}{21}
\newcommand{\lcHold}{13}
\newcommand{\lcPartly}{3}
\newcommand{\lcNotAdmitted}{5}
\newcommand{\lcUnprobed}{circuit fault and reaction network}
\newcommand{\lcUnprobedRungs}{4}
\newcommand{\drugThreeSol}{3}
\newcommand{\drugThreeOracle}{6}
\newcommand{\drugThreeN}{8}
\newcommand{\drugThreeDiff}{\dci{\ensuremath{-}0.38}{\ensuremath{-}0.88}{+0.25}}
\newcommand{\chemFourSol}{1}
\newcommand{\chemFourN}{8}
\newcommand{\chemFourOracle}{8}
\newcommand{\chemFourDiff}{\dci{\ensuremath{-}0.88}{\ensuremath{-}1.00}{\ensuremath{-}0.63}}
\newcommand{\ilSol}{4}
\newcommand{\ilN}{8}

\newcommand{\ilDiff}{\dci{\ensuremath{-}0.50}{\ensuremath{-}0.88}{\ensuremath{-}0.13}}
\newcommand{\drugThreeFinalSol}{6}
\newcommand{\drugThreeFinalOracle}{26}
\newcommand{\drugThreeFinalDiff}{\dci{\ensuremath{-}0.63}{\ensuremath{-}0.81}{\ensuremath{-}0.44}}
\newcommand{\ilFinalSol}{15}
\newcommand{\ilFinalOracle}{32}
\newcommand{\ilFinalDiff}{\dci{\ensuremath{-}0.53}{\ensuremath{-}0.72}{\ensuremath{-}0.34}}
\newcommand{\lcAuditFailures}{147}
\newcommand{\lcAuditRead}{123}

\newcommand{\fTwoHone}{0.49}
\newcommand{\fTwoHtwo}{0.32}
\newcommand{\fTwoHthree}{0.27}
\newcommand{\fTwoLone}{0.47}
\newcommand{\fTwoLtwo}{0.32}
\newcommand{\fTwoLthree}{0.11}

\newcommand{\oLthree}{0.92}

\newcommand{\fHMin}{0.27}
\newcommand{\fHMax}{0.49}

\newcommand{\fLthree}{0.11}
\newcommand{\oMin}{0.91}
\newcommand{\oMax}{0.94}
\newcommand{\qMin}{0.04}
\newcommand{\qMax}{0.16}
\newcommand{\complMax}{0.007}
\newcommand{\fN}{1,886}
\newcommand{\fShared}{680}
\newcommand{\roundsHthree}{19}

\newcommand{\fMalformed}{10}
\newcommand{\qMalformed}{18}

\newcommand{\fWrongMin}{83}
\newcommand{\fWrongMax}{90}
\newcommand{\fUnsafe}{141}
\newcommand{\fUnsafeAL}{91}
\newcommand{\fTokMin}{91,000}
\newcommand{\fTokMax}{363,000}
\newcommand{\fUsd}{0.13}
\newcommand{\fUsdHone}{0.07}
\newcommand{\fUsdLthree}{0.27}
\newcommand{\fUsdCore}{0.037}
\newcommand{\fDecShared}{20.5}
\newcommand{\qDec}{19.6}
\newcommand{\fExpHone}{6.3}
\newcommand{\qExpHone}{5.8}
\newcommand{\fExpLtwo}{18.9}
\newcommand{\qExpLtwo}{16.4}
\newcommand{\fTimesQ}{3.9}
\newcommand{\fLthreeK}{27}
\newcommand{\fLthreeN}{252}
\newcommand{\qLthreeK}{5}
\newcommand{\qLthreeN}{112}
\newcommand{\gtLone}{1.00}
\newcommand{\gtLtwo}{0.96}
\newcommand{\gtLthree}{0.06}
\newcommand{\ccLone}{0.79}
\newcommand{\ccLthree}{0.06}
\newcommand{\stFirstMin}{0.81}
\newcommand{\stFirstMax}{0.85}
\newcommand{\stThirdLone}{0.55}
\newcommand{\stFifthLtwo}{0.49}
\newcommand{\stEighthLthree}{0.38}
\newcommand{\stReachLone}{340}
\newcommand{\stReachLtwo}{310}
\newcommand{\stReachLthree}{224}
\newcommand{\stNinth}{29}
\newcommand{\stLtwoRight}{107}
\newcommand{\stLtwoWrong}{203}
\newcommand{\stLtwoEarly}{103}
\newcommand{\stLtwoFinal}{46}
\newcommand{\stQReachLone}{110}
\newcommand{\stQReachLtwo}{93}
\newcommand{\stQReachLthree}{94}
\newcommand{\stQNinth}{9}
\newcommand{\exHoneCp}{0.58}
\newcommand{\exHtwoCc}{0.42}
\newcommand{\exLoneCp}{0.33}
\newcommand{\exLtwoCf}{0.00}
\newcommand{\exLthreeGt}{0.06}
\newcommand{\famMost}{causal intervention at H1; group testing at H1, L1 and L2; hidden law at L1; knockout network at H1 and H2}
\newcommand{\famRare}{automaton learning, circuit fault, heat inclusion}
\newcommand{\famRareN}{three}
\newcommand{\famRareMaxK}{12}
\newcommand{\famRareMinN}{133}
\newcommand{\dPooled}{\dci{+0.249}{+0.213}{+0.284}}
\newcommand{\dH}{\dci{+0.292}{+0.241}{+0.342}}
\newcommand{\dL}{\dci{+0.206}{+0.160}{+0.256}}
\newcommand{\dLthree}{\dci{+0.071}{+0.009}{+0.134}}

\newcommand{\stAbfMin}{0.15}
\newcommand{\stAbfMax}{0.19}
\newcommand{\cutCorePct}{14.9}
\newcommand{\cutCoreWith}{0.24}
\newcommand{\cutCoreWithout}{0.96}
\newcommand{\cutPhysPct}{19.2}
\newcommand{\cutPhysWith}{0.08}
\newcommand{\cutPhysWithout}{0.34}
\newcommand{\cutChemPct}{20.1}
\newcommand{\cutChemWith}{0.15}
\newcommand{\cutChemWithout}{0.90}
\newcommand{\solAuditCalls}{19,838}
\newcommand{\solAuditMaxTok}{46,457}
\newcommand{\rigidMaxDeg}{0.04}
\newcommand{\rigidReleases}{128}
\newcommand{\heatMaxMK}{19}
\newcommand{\heatCandidates}{2,916}
\newcommand{\heatNoiseMinMK}{80}
\newcommand{\qThinkMin}{3,813}
\newcommand{\qThinkMax}{4,039}
\newcommand{\qThinkFrontierMin}{3,877}
\newcommand{\qThinkFrontierMax}{3,999}

\newcommand{\fEpAll}{1,886}
\newcommand{\qEpAll}{680}
\newcommand{\roundsLthree}{18}
\newcommand{\nSharedH}{336}
\newcommand{\nSharedL}{344}
\newcommand{\fCoreAll}{0.58}
\newcommand{\fCoreCompleted}{0.93}
\newcommand{\fCoreInvalid}{0.37}
\newcommand{\clusterLthree}{\dci{+0.071}{\ensuremath{-}0.009}{+0.152}}

\newcommand{\physCutEpisodes}{350}
\newcommand{\ciSuccConcise}{0.111}
\newcommand{\ciSuccBroad}{0.139}
\newcommand{\ciSuccN}{72}
\newcommand{\routeConciseK}{5}
\newcommand{\routeConciseN}{23}
\newcommand{\routeConcise}{0.22}
\newcommand{\routeBroadK}{19}
\newcommand{\routeBroadN}{24}
\newcommand{\routeBroad}{0.79}
\newcommand{\routeDevMin}{0.07}
\newcommand{\routeDevMax}{0.16}
\newcommand{\routeDevN}{140}
\newcommand{\innerIdealMin}{0.44}
\newcommand{\innerIdealMax}{0.72}
\newcommand{\replayFlashSame}{129}
\newcommand{\replayFlashN}{130}
\newcommand{\replayFlashBytes}{49}
\newcommand{\replayQSame}{80}
\newcommand{\replayQN}{80}
\newcommand{\faN}{40}
\newcommand{\faCells}{six}
\newcommand{\faCommits}{32}

\newcommand{\fmaxN}{14}
\newcommand{\fmaxHoneK}{10}
\newcommand{\fmaxHoneDef}{8}
\newcommand{\fmaxHoneSol}{12}
\newcommand{\fmaxHoneOracle}{12}
\newcommand{\fmaxHoneMinusDef}{\dci{+0.14}{\ensuremath{-}0.14}{+0.43}}
\newcommand{\fmaxHoneSolMinus}{\dci{+0.14}{0.00}{+0.36}}
\newcommand{\fmaxHoneMinusOracle}{\dci{\ensuremath{-}0.14}{\ensuremath{-}0.36}{0.00}}
\newcommand{\fmaxHtwoK}{11}
\newcommand{\fmaxHtwoDef}{7}
\newcommand{\fmaxHtwoSol}{12}
\newcommand{\fmaxHtwoOracle}{13}
\newcommand{\fmaxHtwoMinusDef}{\dci{+0.29}{+0.07}{+0.57}}
\newcommand{\fmaxHtwoSolMinus}{\dci{+0.07}{\ensuremath{-}0.14}{+0.29}}
\newcommand{\fmaxHtwoMinusOracle}{\dci{\ensuremath{-}0.14}{\ensuremath{-}0.36}{0.00}}
\newcommand{\fmaxHthreeK}{8}
\newcommand{\fmaxHthreeDef}{4}
\newcommand{\fmaxHthreeSol}{13}
\newcommand{\fmaxHthreeOracle}{13}
\newcommand{\fmaxHthreeMinusDef}{\dci{+0.29}{+0.07}{+0.50}}
\newcommand{\fmaxHthreeSolMinus}{\dci{+0.36}{+0.14}{+0.64}}
\newcommand{\fmaxHthreeMinusOracle}{\dci{\ensuremath{-}0.36}{\ensuremath{-}0.64}{\ensuremath{-}0.14}}
\newcommand{\fmaxLoneK}{11}
\newcommand{\fmaxLoneDef}{6}
\newcommand{\fmaxLoneSol}{12}
\newcommand{\fmaxLoneOracle}{14}
\newcommand{\fmaxLoneMinusDef}{\dci{+0.36}{+0.14}{+0.64}}
\newcommand{\fmaxLoneSolMinus}{\dci{+0.07}{0.00}{+0.21}}
\newcommand{\fmaxLoneMinusOracle}{\dci{\ensuremath{-}0.21}{\ensuremath{-}0.43}{0.00}}
\newcommand{\fmaxLtwoK}{10}
\newcommand{\fmaxLtwoDef}{5}
\newcommand{\fmaxLtwoSol}{14}
\newcommand{\fmaxLtwoOracle}{14}
\newcommand{\fmaxLtwoMinusDef}{\dci{+0.36}{+0.07}{+0.64}}
\newcommand{\fmaxLtwoSolMinus}{\dci{+0.29}{+0.07}{+0.50}}
\newcommand{\fmaxLtwoMinusOracle}{\dci{\ensuremath{-}0.29}{\ensuremath{-}0.50}{\ensuremath{-}0.07}}
\newcommand{\fmaxLthreeK}{7}
\newcommand{\fmaxLthreeDef}{2}
\newcommand{\fmaxLthreeSol}{8}
\newcommand{\fmaxLthreeOracle}{13}
\newcommand{\fmaxLthreeMinusDef}{\dci{+0.36}{+0.14}{+0.64}}
\newcommand{\fmaxLthreeSolMinus}{\dci{+0.07}{\ensuremath{-}0.14}{+0.29}}
\newcommand{\fmaxLthreeMinusOracle}{\dci{\ensuremath{-}0.43}{\ensuremath{-}0.71}{\ensuremath{-}0.21}}
\newcommand{\fmaxHarderN}{42}
\newcommand{\fmaxHarderK}{29}
\newcommand{\fmaxHarderDef}{19}
\newcommand{\fmaxHarderSol}{37}
\newcommand{\fmaxHarderOracle}{38}
\newcommand{\fmaxHarderMinusDef}{\dci{+0.24}{+0.10}{+0.38}}
\newcommand{\fmaxHarderSolMinus}{\dci{+0.19}{+0.05}{+0.33}}
\newcommand{\fmaxHarderMinusOracle}{\dci{\ensuremath{-}0.21}{\ensuremath{-}0.36}{\ensuremath{-}0.10}}
\newcommand{\fmaxLongN}{42}
\newcommand{\fmaxLongK}{28}
\newcommand{\fmaxLongDef}{13}
\newcommand{\fmaxLongSol}{34}
\newcommand{\fmaxLongOracle}{41}
\newcommand{\fmaxLongMinusDef}{\dci{+0.36}{+0.19}{+0.52}}
\newcommand{\fmaxLongSolMinus}{\dci{+0.14}{+0.02}{+0.26}}
\newcommand{\fmaxLongMinusOracle}{\dci{\ensuremath{-}0.31}{\ensuremath{-}0.45}{\ensuremath{-}0.17}}
\newcommand{\fmaxAllN}{84}
\newcommand{\fmaxAllK}{57}
\newcommand{\fmaxAllDef}{32}
\newcommand{\fmaxAllSol}{71}
\newcommand{\fmaxAllOracle}{79}
\newcommand{\fmaxAllMinusDef}{\dci{+0.30}{+0.19}{+0.40}}
\newcommand{\fmaxAllSolMinus}{\dci{+0.17}{+0.07}{+0.26}}
\newcommand{\fmaxAllMinusOracle}{\dci{\ensuremath{-}0.26}{\ensuremath{-}0.36}{\ensuremath{-}0.18}}
\newcommand{\fmaxAllRate}{0.68}
\newcommand{\fmaxAllLo}{0.57}
\newcommand{\fmaxAllHi}{0.77}
\newcommand{\fmaxAllSolWins}{16}
\newcommand{\fmaxAllSolLosses}{2}
\newcommand{\fmaxCalls}{1,796}
\newcommand{\fmaxMaxCallTok}{170,299}
\newcommand{\fmaxUsdMin}{0.12}
\newcommand{\fmaxUsdMinRung}{H1}
\newcommand{\fmaxUsdMax}{0.53}
\newcommand{\fmaxUsdMaxRung}{L3}
\newcommand{\fmaxTokMin}{198,000}
\newcommand{\fmaxTokMax}{837,000}
\newcommand{\warm}{\mbox{Warm-9B}}
\newcommand{\labrl}{\mbox{LabRL-9B}}
\newcommand{\mixrl}{\mbox{MixRL-9B}}
\newcommand{\lrRepStart}{0.221}
\newcommand{\lrRepLabEight}{0.579}
\newcommand{\lrRepLabTwelve}{0.646}
\newcommand{\lrRepLab}{0.721}

\newcommand{\lrRepMixTwelve}{0.646}
\newcommand{\lrRepMix}{0.583}
\newcommand{\lrRepFlashLow}{0.446}
\newcommand{\lrRepFlashHigh}{0.371}
\newcommand{\lrRepLabMinusStart}{\dci{+0.500}{+0.400}{+0.596}}

\newcommand{\lrRepLabMinusFlashLow}{\dci{+0.275}{+0.150}{+0.404}}
\newcommand{\lrRepLabMinusFlashHigh}{\dci{+0.350}{+0.217}{+0.483}}
\newcommand{\lrRepMixMinusStart}{\dci{+0.362}{+0.283}{+0.446}}
\newcommand{\lrRepMixMinusFlashLow}{\dci{+0.138}{+0.017}{+0.254}}

\newcommand{\lrRepMixMinusLab}{\dci{-0.138}{-0.221}{-0.058}}

\newcommand{\lrRepMixTwelveMinusLabTwelve}{\dci{+0.000}{-0.067}{+0.071}}

\newcommand{\lrMcStart}{0.141}

\newcommand{\lrMcLab}{0.125}

\newcommand{\lrMcMixTwelve}{0.359}
\newcommand{\lrMcMix}{0.328}
\newcommand{\lrMcFlashLow}{0.359}
\newcommand{\lrMcFlashHigh}{0.094}
\newcommand{\lrMcLabMinusStart}{\dci{-0.016}{-0.078}{+0.062}}

\newcommand{\lrMcMixMinusStart}{\dci{+0.188}{+0.078}{+0.281}}
\newcommand{\lrMcMixMinusFlashLow}{\dci{-0.031}{-0.281}{+0.203}}

\newcommand{\lrMcMixMinusLab}{\dci{+0.203}{+0.078}{+0.328}}

\newcommand{\lrMcMixTwelveMinusLabTwelve}{\dci{+0.219}{+0.109}{+0.344}}

\newcommand{\lrRepWorlds}{30}
\newcommand{\lrMcWorlds}{8}
\newcommand{\lrPanelWorlds}{38}
\newcommand{\lrG}{8}
\newcommand{\lrRepEpisodes}{240}
\newcommand{\lrMcEpisodes}{64}
\newcommand{\lrRepStrata}{ten}
\newcommand{\lrRepPerStratum}{three}
\newcommand{\lrMcLevels}{two}
\newcommand{\lrMcPerLevel}{four}
\newcommand{\lrUpdates}{16}
\newcommand{\lrWorldsPerUpdate}{24}
\newcommand{\lrEpisodesPerUpdate}{192}
\newcommand{\lrEpisodesRun}{3,072}
\newcommand{\lrLR}{3\times 10^{-6}}
\newcommand{\lrWarmupUpdates}{2}
\newcommand{\lrClip}{0.2}
\newcommand{\lrDualClip}{3}
\newcommand{\lrRefCoef}{0.03}
\newcommand{\lrGradClip}{1.0}
\newcommand{\lrAdamBetaTwo}{0.95}
\newcommand{\lrAdamBetaOne}{0.9}
\newcommand{\lrMaskedTokens}{2}

\newcommand{\lrMaskNats}{5}
\newcommand{\lrTurnCap}{4,096}
\newcommand{\lrEpisodeCap}{16,384}
\newcommand{\lrContext}{40,960}
\newcommand{\lrMcLabPerUpdate}{16}
\newcommand{\lrMcMcPerUpdate}{8}
\newcommand{\lrTrainRepWorldsLab}{384}
\newcommand{\lrTrainRepWorldsMix}{256}
\newcommand{\lrMcWorldsMix}{128}

\newcommand{\lrTrainRepWorldsTwelve}{288}
\newcommand{\lrMixTrainRepWorldsTwelve}{192}
\newcommand{\lrHoursLab}{9.4}
\newcommand{\lrHoursMix}{8.9}
\newcommand{\lrTrainRewardFirst}{0.20}
\newcommand{\lrTrainRewardLast}{0.61}

\newcommand{\lrFlashLowCut}{24}
\newcommand{\lrFlashHighCut}{66}
\newcommand{\lrPanelTurnCapMc}{8,192}
\newcommand{\lrPoolWorlds}{1,726}
\newcommand{\lrWarmWorlds}{56}
\newcommand{\lrAnsStart}{0.813}
\newcommand{\lrRStart}{6}
\newcommand{\lrPXStart}{0.501}
\newcommand{\lrPStart}{0.403}
\newcommand{\lrXStart}{0.599}
\newcommand{\lrWFStart}{0.447}

\newcommand{\lrWFLabTwelve}{0.586}
\newcommand{\lrAnsLab}{0.976}
\newcommand{\lrRLab}{4}
\newcommand{\lrPXLab}{0.605}
\newcommand{\lrPLab}{0.478}
\newcommand{\lrXLab}{0.732}
\newcommand{\lrWFLab}{0.454}

\newcommand{\lrWFMixTwelve}{0.658}

\newcommand{\lrWFMix}{0.553}

\newcommand{\lrPXFlash}{0.849}

\newcommand{\lrWFFlash}{0.895}
\newcommand{\lrPXLabEightMinusStart}{\dci{+0.125}{+0.081}{+0.169}}

\newcommand{\lrPXLabTwelveMinusStart}{\dci{+0.127}{+0.086}{+0.169}}
\newcommand{\lrWFLabTwelveMinusStart}{\dci{+0.138}{+0.013}{+0.263}}
\newcommand{\lrPXLabMinusStart}{\dci{+0.104}{+0.062}{+0.146}}
\newcommand{\lrWFLabMinusStart}{\dci{+0.007}{-0.105}{+0.118}}
\newcommand{\lrPXMixEightMinusStart}{\dci{+0.093}{+0.051}{+0.137}}

\newcommand{\lrPXMixTwelveMinusStart}{\dci{+0.095}{+0.053}{+0.138}}
\newcommand{\lrWFMixTwelveMinusStart}{\dci{+0.211}{+0.086}{+0.336}}
\newcommand{\lrPXMixMinusStart}{\dci{+0.108}{+0.066}{+0.151}}
\newcommand{\lrWFMixMinusStart}{\dci{+0.105}{-0.013}{+0.230}}
\newcommand{\lrMatchedN}{505}
\newcommand{\lrMatchedLab}{0.642}
\newcommand{\lrMatchedStart}{0.645}
\newcommand{\lrRItems}{7}
\newcommand{\lrPXnpStart}{0.506}
\newcommand{\lrPXnpLabTwelve}{0.631}
\newcommand{\lrPXnpLabTwelveMinusStart}{\dci{+0.125}{+0.082}{+0.171}}
\newcommand{\lrPXnpLab}{0.620}
\newcommand{\lrPXnpLabMinusStart}{\dci{+0.114}{+0.070}{+0.158}}
\newcommand{\lrPXnpMixTwelve}{0.592}
\newcommand{\lrPXnpMixTwelveMinusStart}{\dci{+0.085}{+0.038}{+0.132}}
\newcommand{\lrPXnpMix}{0.610}
\newcommand{\lrPXnpMixMinusStart}{\dci{+0.104}{+0.058}{+0.151}}
\newcommand{\lrPowerSteps}{86}
\newcommand{\lrPowerBins}{51}
\newcommand{\lrPXnpItems}{581}
\newcommand{\lrPXMixMinusLab}{\dci{+0.004}{-0.035}{+0.045}}
\newcommand{\lrPXFlashMinusLab}{\dci{+0.245}{+0.206}{+0.284}}

\newcommand{\lrPXItems}{632}
\newcommand{\lrPItems}{360}
\newcommand{\lrXItems}{272}
\newcommand{\lrWFItems}{38}
\newcommand{\lrWFSteps}{152}
\newcommand{\lrWFClusters}{three}
\newcommand{\lrFlashMaxTokens}{24,000}
\newcommand{\lrEvalMaxTokens}{12,000}
\newcommand{\lrEvalTemp}{0.6}
\newcommand{\lrEvalTopP}{0.95}
\newcommand{\lrGpqaStart}{159}
\newcommand{\lrGpqaMedStart}{12,348}
\newcommand{\lrGpqaLab}{161}
\newcommand{\lrGpqaMedLab}{10,560}
\newcommand{\lrGpqaConcise}{155}

\newcommand{\lrGpqaConciseRepeat}{162}

\newcommand{\lrGpqaMedRatio}{86\%}
\newcommand{\lrGpqaN}{198}
\newcommand{\solMinusOracleFam}{\dci{\ensuremath{-}0.10}{\ensuremath{-}0.19}{0.00}}
\newcommand{\solLongMinusOracleFam}{\dci{\ensuremath{-}0.17}{\ensuremath{-}0.29}{\ensuremath{-}0.07}}
\newcommand{\fmaxAllMinusDefFam}{\dci{+0.30}{+0.15}{+0.44}}
\newcommand{\snapshot}{snapshot V4.1-Flash-1017}
\newcommand{\solname}{GPT-6.1 Sol}

\newcommand{\ohmark}{%
  \begin{tikzpicture}[baseline=-0.6ex]
    \fill[ohteal] (0,0) circle (0.36cm);
    \begin{scope}[shift={(-0.21cm,-0.16cm)},x=0.013125cm,y=0.013125cm]
      \fill[white] (0,16) arc[start angle=180,end angle=360,radius=16] -- (24,16)
                   arc[start angle=0,end angle=-180,radius=8] -- cycle;
      \fill[white] (8,16) -- (24,16) -- (16,24) -- (9.333,24) .. controls (8.597,24) and (8,23.403) .. (8,22.667) -- cycle;
    \end{scope}
  \end{tikzpicture}}

\begin{document}
\thispagestyle{firstpage}

\begin{tcolorbox}[enhanced,colback=ohpanel,colframe=ohpanelrule,boxrule=0.5pt,arc=7pt,left=16pt,right=16pt,top=16pt,bottom=12pt]
{\sffamily\bfseries\fontsize{17}{21.5}\selectfont\color{ohink}Openhagen Worlds: Calibrated Environments for\\ Measuring and Training Scientific Discovery Agents\par}
\vspace{9pt}
{\sffamily\bfseries\small\color{ohink}Alexander Boesgaard Lorup\par}
\vspace{2pt}
{\sffamily\small\color{ohmuted}Openhagen\par}
\vspace{11pt}
{\small
Agents that do science must choose informative experiments and reason from their results, but most discovery benchmarks cannot separate that skill from recall, unstated scoring rules or shortcut policies.
We introduce Openhagen Worlds, a suite of 27 procedurally generated environments (some with finite truth spaces), from classic puzzles of experimental reasoning to simulator-grounded physics and chemistry, in which an agent runs experiments under a budget and commits an answer that a deterministic verifier checks against a hidden truth.
Every observation states the complete scoring rule, and each of the 178 difficulty levels is calibrated: it is admitted only if, on 64 development instances, a scripted oracle succeeds on at least 0.75, the strongest plausible non-deductive policy on at most 0.25 and each gated shortcut on at most 0.35.
The 92 core levels span all 27 families and are sized to train small models; the 86 frontier rungs extend the 15 discovery worlds with harder questions and with campaigns of three, five and eight or more committed stages on one hidden system.
On the same \fmaxAllN{} withheld frontier instances, one per family and rung, DeepSeek V4.1 Flash solves \fmaxAllDef{} at its default reasoning effort with 16,384 tokens per call and \fmaxAllK{} at its maximum effort with up to 393,216 tokens per call (oracle: \fmaxAllOracle{}); the headroom that remains is largest in eight-stage campaigns.
On-policy expert iteration gives a 9B model modest gains: its core-tier success on trained families rises from \trConcise{} to \trBroad{}, partly through better action format, with no detectable loss on GPQA Diamond.
A hosted service plays the core tier's withheld instances.\par}
\vspace{9pt}
{\small\textbf{Date:} {\color{ohmuted}November 2026}\hspace{1.4em}\textbf{Evaluation service:} \href{https://eval.openhagen.ai}{\texttt{eval.openhagen.ai}}\hspace{1.4em}\href{https://eval.openhagen.ai/sample/}{\textbf{Explore 48 sample instances}}\hfill\raisebox{-0.5ex}{\ohmark}\par}
\end{tcolorbox}
\vspace{4pt}

\section{Introduction}
\label{sec:intro}

Reinforcement learning with verifiable rewards has made programmatic checks a central training signal for reasoning models \citep{lambert2024tulu3,deepseek2025r1}, and agents are now asked to do science: they are expected to choose informative experiments, take the results into account and commit to a conclusion.
For such agents an environment is only useful if its reward measures that skill.
The answer must be checked exactly against something the agent cannot look up, the conditions for success must be stated, no policy that skips the reasoning may score well, and the levels must stay hard as models improve.

Existing discovery environments cover a wide range of science and have shown how far agents are from autonomous discovery (\cref{tab:related}).
To our knowledge, none of them admits a level only after showing that scripted policies which skip the reasoning stay below a fixed success rate on it, and none shows the agent the complete rule it is judged by: tolerances, partial credit and judge rubrics stay inside the evaluator.
Most end each episode with a single answer, and, to our knowledge, none chains several irrevocably committed answers on one hidden system, with later stages depending on earlier ones, so none tests whether an agent can carry its own conclusions through a campaign.
And on the easiest levels, frontier models can already approach the ceiling: on NewtonBench, reasoning models reach 70--100\% symbolic accuracy in the easiest setting, while in the hardest only two of them exceed 6\% \citep{zheng2026newtonbench}.

To fill this gap we introduce Openhagen Worlds, a suite of 27 procedurally generated environment families with 178 calibrated difficulty levels (\cref{sec:construction}).
Each family hides a truth, publishes its full scoring rule in every observation and verifies a committed answer exactly or within a published tolerance.
The families range from classic puzzles that isolate experimental reasoning, such as group testing and the counterfeit-coin problem, to simulator-grounded physics and chemistry.
Each level is admitted by a calibration gate: on 64 development instances, a scripted oracle must succeed on at least 0.75 of them, while the strongest plausible non-deductive policy may succeed on at most 0.25 and every gated shortcut on at most 0.35; we call a level that passes the gate \emph{calibrated}.
The gate's measurements also show what each family tests: with the oracle interpreting the readings, a fixed sequence of every allowed experiment solves 0.98--1.00 of core mechanism-discrimination development instances, where the difficulty lies in the analysis, but only 0.02--0.28 of rule-induction ones, where the experiments must be chosen (\cref{tab:calibration}).
The 92 levels of the \emph{core tier} span all 27 families.
The 86 rungs of the \emph{frontier tier} extend the 15 discovery worlds with harder single questions (H1--H3) and with long-horizon campaigns of three, five and eight or more committed stages on one hidden system (L1--L3).
Evaluation instances come from a key that is never distributed, and a hosted service plays them.

Openhagen Worlds was built first as a training ground for smaller models.
Its core tier is sized so that models of a few billion parameters find first successes and a learning signal, with the aim of shaping them into specialised scientific reasoners for narrow domains, where a small model trained for one kind of experiment can be cheaper and more reliable than a general one.
That a frontier model approaches saturation on this tier is therefore expected: the pilot model solves all \solCoreN{} of its core instances, and Flash solves \fCoreCompleted{} of the core episodes it finishes, though \fCoreInvalid{} of its episodes end without a legal action because its turns are cut at the per-call token limit.
The frontier tier applies the same construction to harder rungs, admitted by the same gate.
A pilot with a stronger model at xhigh reasoning effort, below its ``max'' setting, solves most of that tier, and so does Flash at its maximum reasoning effort, so difficulty has to be a ladder that keeps climbing.
Probes on development instances show where it can climb: most longer campaigns admitted by the same gate hold the stronger model, while rungs that only add hidden structure do not (\cref{sec:ladder}).

We evaluate scripted references, two locally run models built on open-weight Qwen models and a fast frontier model through its API on the withheld instances (\cref{sec:results}).
The core tier separates the local models from the oracle and approaches saturation for frontier models.
On its discovery worlds, \flash{} solves \fCoreAll{} of its episodes; most of the others end without a legal action because its turns are cut at the per-call token limit, and of the episodes it finishes it solves \fCoreCompleted{}.
On the frontier tier, Flash solves \fHMin--\fHMax{} of its episodes on each of the five rungs H1--H3, L1 and L2, and \fLthree{} on the eight-stage campaigns (L3), where only a window of recent records stays visible; the oracle solves \oMin--\oMax{} on every rung.
\qbig{} reaches \qMin--\qMax{} there.
Most failed episodes of both models end on a committed wrong answer rather than on malformed output; for Flash, whose thinking reaches its per-call limit of 16,384 tokens on \capCutPct\% of frontier decisions, these endings do not separate reasoning errors from the budget, and episodes that end on a turn re-asked after the limit succeed on \capReaskRate{}, against \capThinkRate{} for the others.
Its frontier numbers therefore measure reasoning under that budget, and much of the difficulty is a budget effect: on the $o = 0$ frontier instances, one per family and rung, Flash at its maximum effort with up to 393,216 tokens per call solves \fmaxAllK{} of \fmaxAllN{}, against \fmaxAllDef{} under the cap and \fmaxAllOracle{} for the oracle (\cref{sec:flashmax}).
A pilot with \solModel{}, on \solPilotScope{}, succeeds on \solFrontierRate{} of frontier episodes, \solMinusFlash{} above Flash capped at 16,384 tokens per call but \fmaxAllSolMinus{} above Flash at its maximum effort, and stays below the oracle, \solMinusOracle{}, a gap that a bootstrap over families resolves only on the long-horizon rungs (\cref{app:cluster}).
The headroom that holds is therefore measured against the oracle, not against Flash's capped rates: Flash at its maximum effort stays below the oracle over all rungs (\fmaxAllMinusOracle{}), in the eight-stage campaigns the stronger model solves \solLthreeK{} of \solLthreeN{} (oracle \oracleLthreeK{}; \solLthreeMinusOracle{}) and Flash at its maximum effort \fmaxLthreeK{} of \fmaxN{}, and development-instance probes find more headroom in the longer campaigns of \cref{sec:longcampaigns}.

Beyond aggregate scores, we analyse the episodes and train on the suite (\cref{sec:analysis,sec:training}).
Campaign accuracy falls stage by stage, and Flash runs about as many experiments as the 27B and succeeds far more often, under a larger per-call budget.
On-policy expert iteration gives a 9B model modest gains on trained and held-out families, with no detectable regression in general reasoning on GPQA Diamond.
In a separate study on two laboratory families outside the suite (\cref{sec:labrl}), online reinforcement learning raises the same model's success on held-out replication worlds from \lrRepStart{} to \lrRepLab{} in one training run, above Flash with the same tools at low and default reasoning effort (\lrRepFlashLow{} and \lrRepFlashHigh{}) when Flash plays under the 9B's per-turn and per-episode token caps; with up to \lrFlashMaxTokens{} tokens per call, Flash stays well ahead on a broader benchmark of laboratory data analysis (\lrPXFlash{} against \lrPXLab{}), and the registered checkpoint fails one of that benchmark's registered guards. A run that splits its updates between replication and method-comparison worlds improves on both families.
To summarise, our contributions are as follows:
\begin{itemize}
  \item We present Openhagen Worlds, 27 calibrated, procedurally generated discovery environments (with finite truth spaces in some families; \cref{app:calibration}) with 178 difficulty levels in two tiers, each level admitted by a per-level calibration gate, with withheld evaluation instances, a hosted evaluation service and a public sample.
  \item We characterise the calibration and difficulty of both tiers, showing that every level separates deduction from every gated non-deductive policy, and that a stronger model at xhigh reasoning effort solves most of the frontier tier but stays well below the oracle on its longest campaigns.
  \item We evaluate a fast frontier model and locally run models built on open-weight Qwen models on the withheld instances, with a stronger frontier model in a pilot, and report success, failure modes, decisions, cost per episode and the effect of the per-call token budget.
  \item Through episode analysis and two training studies, we show how campaign errors compound, that on-policy expert iteration gives modest gains that partly reach held-out families, and that online reinforcement learning on laboratory worlds raises a 9B model's success on held-out worlds of its training family well above its starting point and above a fast frontier model held to the 9B's token caps, while that model stays well ahead at a larger budget on broader laboratory analysis.
\end{itemize}

\section{Related Work}
\label{sec:related}

\paragraph{Discovery environments.}
Text and simulated worlds have long been used to evaluate agents that experiment.
ScienceWorld poses elementary-school science tasks in an interactive text environment \citep{wang2022scienceworld}, and DiscoveryWorld asks agents to complete full discovery cycles across eight topics \citep{jansen2024discoveryworld}.
BoxingGym scores experimental design by expected information gain and model discovery by whether an explanation lets a second agent predict \citep{gandhi2025boxinggym}.
NewtonBench, PhysGym and DiscoverPhysics ask agents to recover physical laws from interactive experiments, often under shifted laws or withheld prior knowledge \citep{zheng2026newtonbench,chen2025physgym,wiemann2026discoverphysics}.
These benchmarks measure valuable skills, but to our knowledge none admits a level only after showing that non-deductive policies stay below a fixed success rate on it (ScienceWorld reports a random-action baseline and NewtonBench a test without observed data, neither used to admit levels), and none tests campaigns in which committed answers must be carried across stages.

\begin{table}[tbp]
\centering
\caption{\capt{Comparison of representative discovery and reasoning environments.}
Rows are grouped by their primary target. \emph{Experiments}: the agent chooses its own experiments over several turns. \emph{Training}: the work trains models on the environment. \emph{Gated levels}: every level is admitted only after scripted non-deductive policies are shown to stay below a fixed success threshold on it. \emph{Campaigns}: episodes that chain several separately and irrevocably committed sub-answers on one persistent hidden system under a pooled budget, later stages depending on earlier truths or commits. Entries for other works follow their published descriptions.}
\label{tab:related}
\footnotesize
\setlength{\tabcolsep}{5pt}
\begin{tabular}{@{}L{3.5cm}L{4.4cm}cccc@{}}
\toprule
\textbf{\sffamily Environment} & \textbf{\sffamily Primary emphasis} & \textbf{\sffamily Experiments} & \textbf{\sffamily Training} & \textbf{\sffamily Gated levels} & \textbf{\sffamily Campaigns} \\
\midrule
\grouprow{6}{Interactive science environments}
\textbf{ScienceWorld}~\citep{wang2022scienceworld} {\color{ohmuted}[2022]} & Elementary science tasks in a text world & \yes & \yes & \no & \no \\
\textbf{DiscoveryWorld}~\citep{jansen2024discoveryworld} {\color{ohmuted}[2024]} & End-to-end discovery tasks in eight topics & \yes & \no & \no & \no \\
\grouprow{6}{Experimental design and law discovery}
\textbf{BoxingGym}~\citep{gandhi2025boxinggym} {\color{ohmuted}[2025]} & Design scored by information gain & \yes & \no & \no & \no \\
\textbf{NewtonBench}~\citep{zheng2026newtonbench} {\color{ohmuted}[2026]} & Laws under counterfactual shifts & \yes & \no & \no & \no \\
\textbf{PhysGym}~\citep{chen2025physgym} {\color{ohmuted}[2025]} & Physics discovery with controlled priors & \yes & \no & \no & \no \\
\textbf{DiscoverPhysics}~\citep{wiemann2026discoverphysics} {\color{ohmuted}[2026]} & Altered laws of motion in N-body worlds & \yes & \no & \no & \no \\
\grouprow{6}{Environments for training}
\textbf{SciDisco}~\citep{xu2026scidisco} {\color{ohmuted}[2026]} & Data-analysis environments for turn-level RL & \no & \yes & \no & \no \\
\textbf{Reasoning Gym}~\citep{stojanovski2025reasoninggym} {\color{ohmuted}[2025]} & Procedural reasoning tasks & \no & \yes & \no & \no \\
\midrule
\rowcolor{ohbluefill}\textbf{\color{ohblue}Openhagen Worlds} & \textbf{Calibrated discovery across 27 families, two tiers} & \yes & \yes & \yes & \yes \\
\bottomrule
\end{tabular}
\end{table}

\paragraph{Environments for training.}
Procedural generation separates training from evaluation instances \citep{cobbe2020procgen}, and Reasoning Gym applies it to verifiable reasoning tasks for reinforcement learning \citep{stojanovski2025reasoninggym}.
\citet{xu2026scidisco} build SciDisco, discovery environments with hidden evidence graphs, and train a 14B model with turn-level reinforcement learning; \citet{kriz2026amortizing} train the reasoning behind experiment choice with an information-gain reward; and \citet{patel2026search} finds that a forward computation distils into a student while a backtracking search procedure does not.
Openhagen Worlds generates every instance, so the same families can be trained on and withheld, and it pairs held-out families with named trained relatives to test transfer.

\paragraph{Verifiable rewards and shortcuts.}
Benchmark builders remove or rewrite items that partial-input baselines or adversarial filters solve without the intended reasoning \citep{gururangan2018annotation,zellers2018swag,lebras2020adversarial}.
The calibration gate applies this idea per difficulty level, with scripted non-deductive policies in place of trained models.
A policy optimised against a reward can also satisfy it in unintended ways \citep{amodei2016concrete,skalse2022defining}, so every episode in Openhagen Worlds records the experiments it ran, and a training filter can check the route to an answer as well as the answer.
Canary strings let corpus builders exclude evaluation data \citep{srivastava2023bigbench}; because a generator distributed for training would otherwise also generate the test, we combine them with a private key for the evaluation instances.

\paragraph{Self-training.}
Expert iteration, which began as imitation of a search expert \citep{anthony2017thinking}, and rejection-sampling fine-tuning train a model on its own verified or highest-reward samples \citep{polu2022formal,zelikman2022star,yuan2023scaling,dong2023raft,singh2023beyond}; on-policy training typically forgets less than fine-tuning on another model's outputs \citep{shenfeld2025razor,chen2026retaining}, and, in controlled card-game and navigation tasks, reinforcement learning generalised further out of distribution than supervised fine-tuning \citep{chu2025sftmemorizes}.
We apply this route to multi-turn agentic environments, with a capped share of teacher episodes rendered in the student's own format.
Group-relative policy optimisation replaces a learned value function with the mean reward of several attempts at the same problem \citep{shao2024deepseekmath,liu2025understanding}; we use it for online reinforcement learning on laboratory worlds.

\section{Benchmark Construction}
\label{sec:construction}

\subsection{Construction pipeline}
\label{sec:pipeline}

\begin{figure}[tbp]
\centering
\begin{tikzpicture}[
  card/.style={draw=ohpanelrule,fill=white,rounded corners=3pt,minimum width=3.6cm,minimum height=3.55cm,inner sep=0pt},
  ttl/.style={font=\sffamily\bfseries\fontsize{7.6}{9}\selectfont,text=ohink,align=left,text width=3.15cm},
  dsc/.style={font=\sffamily\fontsize{6.3}{7.7}\selectfont,text=ohmuted,align=left,text width=3.15cm},
  num/.style={font=\sffamily\fontsize{6}{7}\selectfont,text=ohmuted},
  tag/.style={font=\sffamily\fontsize{5.8}{7}\selectfont,text=ohink},
  ico/.style={fill=ohink,rounded corners=1.5pt,minimum size=0.46cm,inner sep=0pt,text=white,font=\sffamily\bfseries\footnotesize}]
  \foreach \i/\n/\t/\g/\s/\d/\c in {
    0/01/DESIGN/?/{Hidden truth and\\published rule}/{A generator draws a hidden system; every observation states the objective, grammar, budget and complete scoring rule.}/ohink,
    1/02/VERIFY/{\checkmark}/{Exact verification\\and identifiability}/{The verifier checks the commit against the hidden truth; filters drop instances the allowed experiments cannot decide.}/ohink,
    2/03/GATE/{$\geq$}/{Calibration\\gate}/{Scripted policies play 64 development instances: oracle $\geq$ 0.75, naive $\leq$ 0.25, every shortcut $\leq$ 0.35.}/ohteal,
    3/04/FREEZE/\#/{Ladders and withheld\\instances}/{Admitted levels form ladders; final instances come from a private key that is never distributed.}/ohteal} {
    \node[card,anchor=north west] (c\i) at (\i*3.95,0) {};
    \draw[\c,line width=1.2pt] ([xshift=2pt]c\i.north west) -- ([xshift=-2pt]c\i.north east);
    \node[num,anchor=north west] at ([xshift=6pt,yshift=-6pt]c\i.north west) {\n};
    \node[tag,anchor=north east] at ([xshift=-6pt,yshift=-5.5pt]c\i.north east) {{\color{\c}$\bullet$}\,\t};
    \node[ico,anchor=north west] at ([xshift=6pt,yshift=-19pt]c\i.north west) {\g};
    \node[ttl,anchor=north west] at ([xshift=6pt,yshift=-38pt]c\i.north west) {\s};
    \node[dsc,anchor=north west] at ([xshift=6pt,yshift=-61pt]c\i.north west) {\d};
  }
  \foreach \i/\j in {0/1,1/2,2/3} {\draw[-{Stealth[length=3pt]},ohrule] ([yshift=-1.4cm]c\i.north east) -- ([yshift=-1.4cm]c\j.north west);}
  \node[font=\sffamily\fontsize{5.8}{7}\selectfont,text=ohmuted,anchor=south east] at ([yshift=3pt]c3.north east) {{\color{ohink}$\bullet$}\,per family\hspace{8pt}{\color{ohteal}$\bullet$}\,per level};
\end{tikzpicture}
\caption{\capt{The four-stage pipeline that admits every level of Openhagen Worlds.} Stages 01 and 02 are built once per family; stages 03 and 04 run for every difficulty level, in both tiers.}
\label{fig:pipeline}
\end{figure}
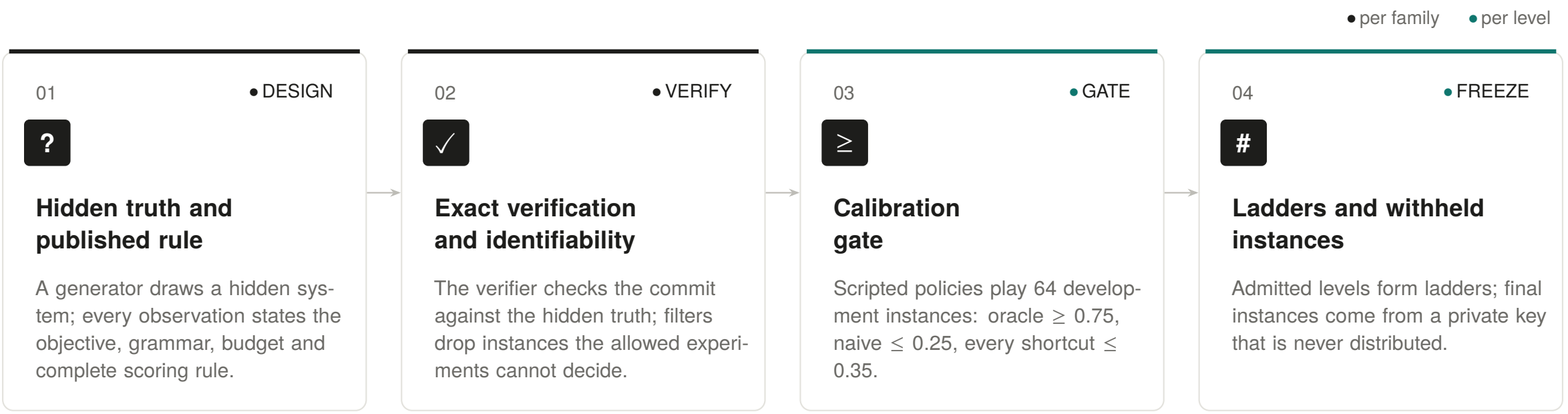

Every family and level passes through the same four-stage pipeline (\cref{fig:pipeline}).
The stages are run by code, not by annotators: a family is a generator, a verifier, a set of scripted policies and a datasheet, and a level is admitted or rejected by the gate's measurements alone.

\paragraph{Stage 1: Hidden truth and published rule.}
A family's generator draws a hidden system (a resistor network with a fault, a kinetic scheme, a causal graph with a confounder, a program with a bug) from its level's settings.
The agent sees a public observation that is re-rendered every turn: the objective, the objects and their limits, the action grammar, the experiment budget and its use, the history of readings and the complete scoring rule, including every tolerance and every condition that ends the episode.
Nothing that decides success is left unstated.

\paragraph{Stage 2: Exact verification and identifiability.}
The verifier scores a committed answer with reward 1 or 0.
Discrete answers must match exactly; answers with many spellings are scored by semantic equivalence (a rule over number triples on all 1,000 triples of its domain, a program fix by its output on every input, an automaton by behavioural equivalence, a molecule by constitutional identity); continuous answers have published tolerances; and design answers, such as a controller, a chromatographic method or a process, are verified by simulating them on the true system.
Identifiability filters reject instances whose required distinctions cannot be made with the allowed experiments, a failure \citet{shetty2026llmidea} document in DiscoverPhysics \citep{wiemann2026discoverphysics}.

\paragraph{Stage 3: The calibration gate.}
The gate runs scripted policies that read only the public observation.
The \emph{oracle} keeps the exact set of hypotheses consistent with the evidence, or a posterior over them, chooses the experiment that splits them best \citep{chaloner1995bayesian} and commits when one remains; it is an idealised reference, not a level models are expected to reach.
The \emph{naive policy} is the strongest plausible non-deductive policy for the family, such as probing in a fixed order or regressing on observational data.
\emph{Shortcut policies} are specific habits that could pass for skill, such as measuring at both safety limits, plus an answer prior that commits another instance's correct answer.
On the frontier tier the panel also includes the strategy that solves the next easier rung, a stage-by-stage policy that re-solves each stage of a campaign from scratch and, on the longest campaigns, an oracle without notes.
A level is admitted only if, on 64 development instances, the oracle succeeds on at least 0.75, the naive policy on at most 0.25 and every shortcut on at most 0.35, and only if truths, oracle replays and instance digests are identical across processes.
The \emph{exhaustive design}, where a family defines one, runs every allowed experiment in a fixed order and then applies the oracle's analysis; it is reported, not gated, and shows whether a family tests the choice of experiments or only the analysis of evidence.
On the frontier tier, the other reported policies are likewise deductive: each applies the oracle's exact analysis to a simpler design, such as the previous rung's design, a design for the current stage only, or a fixed set of probes, and some succeed on more than 0.35 of development instances, which shows that the rung's difficulty there lies in the analysis rather than in the design.
One reported habit is not deductive: on control plant's long-horizon rungs, tuning the outer loop as if the inner loop were perfect succeeds on \innerIdealMin--\innerIdealMax{} of development instances, because a fast inner loop often makes it good enough, and it is reported rather than gated.

\paragraph{Stage 4: Ladders and withheld instances.}
Admitted levels form a family's ladder.
An instance is a pure function of its family, split, level, index and key, so every episode can be replayed.
Training and development instances use separate keys, and the \emph{final instances} come from a third key that is never distributed; every model plays the same ones, and all comparisons are paired. Generators, level settings and verifiers are chosen on development instances; the final instances informed the design of the frontier tier only, which was tuned against a screen of one model on them (\cref{sec:families,sec:limitations}).
The hosted evaluation service plays exactly the core tier's final instances and reveals only the observation, reward, termination and success.

\paragraph{Contamination controls.}
One concern runs through every stage: a generator distributed for training must not also generate the test.
Beyond the private key, every distributed file carries a canary string and every observation an instance hash, famous instances (the twelve-coin puzzle, textbook rule-induction rules, named Bayesian networks and control plants) are excluded by construction, and a keyed filter rejects generated programs that match a protected program-execution benchmark (\cref{app:robustness}).

\subsection{Families, tiers and splits}
\label{sec:families}

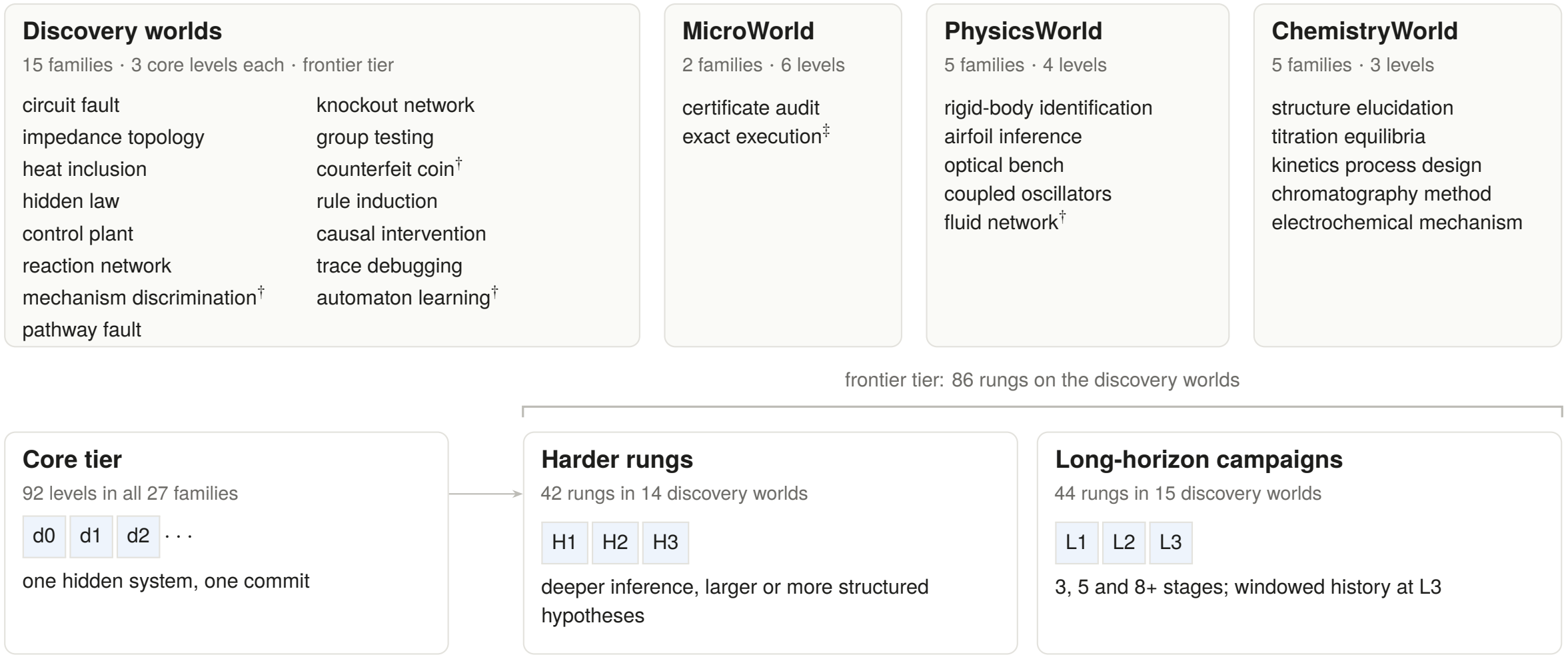
\begin{figure}[tbp]
\centering
\begin{tikzpicture}[
  grp/.style={draw=ohpanelrule,fill=ohpanel,rounded corners=3pt,inner sep=5pt,anchor=north west,font=\sffamily\fontsize{6.4}{8.1}\selectfont,text=ohink},
  tier/.style={draw=ohpanelrule,fill=white,rounded corners=3pt,inner sep=5pt,anchor=north west,font=\sffamily\fontsize{6.4}{8.1}\selectfont,text=ohink}]
  \newcommand{\gh}[2]{{\bfseries\fontsize{7.6}{9}\selectfont #1}\par\vspace{1pt}{\fontsize{6}{7.5}\selectfont\color{ohmuted}#2}\par\vspace{4pt}}
  \newcommand{\chip}[1]{{\setlength{\fboxsep}{1.6pt}\fcolorbox{ohpanelrule}{ohbluefill}{\strut\,#1\,}}}
  \newcommand{\topbox}[3]{\parbox[t][#2][t]{#1}{\raggedright #3}}
  \node[grp] (d) at (0,0) {\topbox{5.95cm}{3.05cm}{\gh{Discovery worlds}{15 families $\cdot$ 3 core levels each $\cdot$ frontier tier}%
    \begin{tabular}{@{}l@{\hspace{14pt}}l@{}}
    circuit fault & knockout network\\ impedance topology & group testing\\ heat inclusion & counterfeit coin$^\dagger$\\
    hidden law & rule induction\\ control plant & causal intervention\\ reaction network & trace debugging\\
    mechanism discrimination$^\dagger$ & automaton learning$^\dagger$\\ pathway fault & \\
    \end{tabular}}};
  \node[grp] (m) at (6.55,0) {\topbox{2.0cm}{3.05cm}{\gh{MicroWorld}{2 families $\cdot$ 6 levels}certificate audit\\ exact execution$^\ddagger$}};
  \node[grp] (p) at (9.15,0) {\topbox{2.65cm}{3.05cm}{\gh{PhysicsWorld}{5 families $\cdot$ 4 levels}rigid-body identification\\ airfoil inference\\ optical bench\\ coupled oscillators\\ fluid network$^\dagger$}};
  \node[grp] (c) at (12.4,0) {\topbox{2.7cm}{3.05cm}{\gh{ChemistryWorld}{5 families $\cdot$ 3 levels}structure elucidation\\ titration equilibria\\ kinetics process design\\ chromatography method\\ electrochemical mechanism}};
  \node[tier] (t0) at (0,-4.25) {\topbox{4.05cm}{1.85cm}{\gh{Core tier}{92 levels in all 27 families}\chip{d0}\,\chip{d1}\,\chip{d2}\,$\cdots$\par\vspace{3pt}one hidden system, one commit}};
  \node[tier] (t1) at (5.15,-4.25) {\topbox{4.55cm}{1.85cm}{\gh{Harder rungs}{42 rungs in 14 discovery worlds}\chip{H1}\,\chip{H2}\,\chip{H3}\par\vspace{3pt}deeper inference, larger or more structured hypotheses}};
  \node[tier] (t2) at (10.25,-4.25) {\topbox{4.85cm}{1.85cm}{\gh{Long-horizon campaigns}{44 rungs in 15 discovery worlds}\chip{L1}\,\chip{L2}\,\chip{L3}\par\vspace{3pt}3, 5 and 8+ stages; windowed history at L3}};
  \draw[ohrule,line width=0.6pt] ([yshift=4pt]t1.north west) -- ++(0,3pt) -| ([yshift=4pt]t2.north east);
  \node[font=\sffamily\fontsize{6.2}{7.5}\selectfont,text=ohmuted,anchor=south] at ([yshift=8pt]$(t1.north west)!0.5!(t2.north east)$) {frontier tier: 86 rungs on the discovery worlds};
  \draw[-{Stealth[length=3pt]},ohrule] ([yshift=-0.62cm]t0.north east) -- ([yshift=-0.62cm]t1.north west);
\end{tikzpicture}
\caption{\capt{The 27 families of Openhagen Worlds in four groups, and the two tiers.} Every family has a core ladder; the 15 discovery worlds also carry the frontier tier of harder (H) and long-horizon (L) rungs. \Cref{tab:suite} lists what each family hides, its experiments and how its readings are computed. $^\dagger$ held out of all training; $^\ddagger$ two levels use an evaluation-only grammar.}
\label{fig:suite}
\end{figure}

\paragraph{Families.}
The 27 families form four groups (\cref{fig:suite}; \cref{tab:suite} in \cref{app:suite}, where \cref{tab:terms} also defines the terms used throughout).
The fifteen \emph{discovery worlds} are the central families of the suite and the only ones with a frontier tier.
Four of them are classic puzzles of experimental reasoning (group testing, counterfeit coin, rule induction and automaton learning), and the others idealise laboratory and reasoning problems, from fault finding in circuits and kinetic models to causal inference and program repair; we treat all fifteen as tests of experimental reasoning rather than of domain science.
Two \emph{MicroWorld} families test long exact computation and evidence auditing.
The simulator-grounded science of the suite is in its last ten families: five \emph{PhysicsWorld} families wrap simulators (rigid-body dynamics, a panel method, ray transfer, modal and hydraulic models), and five \emph{ChemistryWorld} families model spectroscopy, equilibria, kinetics, chromatography and voltammetry.
Each family has a datasheet that records what is hidden, the grammar and scoring rule, the difficulty settings, the calibration, the instance space, contamination notes and the simulator with its known limits.
Two laboratory families outside the suite, used to train and evaluate a model by reinforcement learning, are described in \cref{sec:labrl}.

\paragraph{Core tier.}
Core levels, numbered d0, d1, \dots{} from the easiest, scale the size of the hypothesis space, the noise, the slack in the budget, the depth of the deduction or the length of the computation.
The discovery worlds have three core levels each, the MicroWorld families six, the PhysicsWorld families four and the ChemistryWorld families three: 92 levels in all.
Easier entry levels below them give small models their first successes during training and are not evaluated.

\paragraph{Frontier tier.}
The frontier tier adds six rungs to each discovery world.
Harder rungs (H1--H3) are harder, not noisier: no rung raises the noise or tightens a tolerance relative to the family's hardest core level.
Instead they enlarge or structure the hypothesis space or add an inference step the core levels never need, and each must defeat the strategy that solves the rung below it.
Long-horizon rungs (L1--L3) are \emph{campaigns}: three, five and eight or more stages on one persistent hidden system, each closed by a committed answer, under a pooled experiment budget that is too small to re-solve every stage from scratch.
Later stages depend on earlier truths and in some families on the agent's own earlier commits, such as its committed controller or patch, and an episode succeeds only if every stage is right.
L3 campaigns show only a window of recent records and give the agent a notebook, so it must decide what to write down.
Pooled over families, each ladder is graded, from an entry rung that a fast frontier model (\flash{}) solves about half the time to upper rungs that it solves far less often. That grading was a design target: the tier was built after Flash's results on the core tier, aiming for Flash success of about 0.10--0.40 at each family's top rung, and the target was read from screens of Flash on final instances. Rung settings were chosen on development instances, and no rung setting, generator or verifier changed after any model played the final instances; after the first screen, however, the frontier families' action parsers (a lenient second reading of a rejected reply), part of the campaign wording, the budget display of one family and the re-ask after a cut turn were revised, and the same final instances were played again.
Of the 90 possible rungs, 86 are built and admitted; pathway fault has only L1 and L3, and only its L1 is evaluated here.

\paragraph{Held-out families and splits.}
Three discovery worlds are held out of all training, each with a close trained relative: mechanism discrimination (reaction network, hidden law), counterfeit coin (group testing) and automaton learning (rule induction, trace debugging); fluid network is held out among the PhysicsWorld families, and two exact-execution levels use a construction that no training level contains.
Final instances number 24 per family and level on both tiers (2,208 core instances).

\subsection{Episodes and protocol}
\label{sec:episode}

\begin{figure}[tbp]
\centering
\begin{tcolorbox}[enhanced,colback=ohpanel,colframe=ohpanelrule,boxrule=0.4pt,arc=4pt,left=6pt,right=6pt,top=4pt,bottom=4pt,
  title={Group testing, core level d0 (illustrative episode)},colbacktitle=ohgroup,coltitle=ohink,fonttitle=\sffamily\bfseries\footnotesize,toptitle=2pt,bottomtitle=2pt]
\begin{Verbatim}[fontsize=\scriptsize,commandchars=\\\{\}]
\textbf{Observation} (abridged; re-rendered in full every turn)
  samples:       S01, S02, ..., S32
  objective:     exactly one sample is contaminated; a pooled test of any set of samples
                 is positive if and only if the set contains the contaminated sample
  grammar:       TEST samples=<comma-separated sample labels, e.g. S01,S02>; at most 5
                 tests. Then: COMMIT sample=<label>.
  success rule:  the COMMIT succeeds only if it names the contaminated sample; an unknown
                 sample or a test beyond the budget ends the episode at zero
  budget:        5 tests, 0 used; 8 laboratory turns
\textbf{Episode}
  1  ACTION: TEST samples=S01,S02,...,S16   ->  negative
  2  ACTION: TEST samples=S17,S18,...,S24   ->  positive
  3  Next I will test S17 to S20.           ->  not executed: no legal action (legal forms:
                                                TEST samples=<labels>, COMMIT sample=<label>)
  4  ACTION: TEST samples=S17,S18,S19,S20   ->  negative
  5  ACTION: TEST samples=S21,S22           ->  positive
  6  ACTION: TEST samples=S21               ->  negative
  7  ACTION: COMMIT sample=S22
\textbf{Verifier}  correct sample; reward 1
\end{Verbatim}
\end{tcolorbox}
\caption{\capt{The episode format.} On the easiest core level of group testing, 32 samples and $\lceil\log_2 32\rceil = 5$ pooled tests leave room only for a halving plan. Turn 3 shows how the protocol treats a reply without a legal action. The excerpt was constructed to show the format; \cref{app:samples} shows recorded oracle episodes from the public sample.}
\label{fig:episode}
\end{figure}

An agent receives the public observation and takes one action per turn (\cref{fig:episode}): an experiment, a sandboxed Python analysis of the readings (in families that offer the tool), or a COMMIT, which submits the answer and, outside campaigns, ends the episode.
Actions are accepted as bare commands, JSON objects or tool calls, and equivalent spellings of a command are accepted, so that interface conventions are not what is measured.
A reply without a legal action is returned with the parser's reason and the exact legal form and costs one turn, and two such replies in a row end the episode.
An experiment outside the safe limits or beyond the budget ends the episode at zero, as the observation states.
We call an episode that ends on replies without a legal action an \emph{invalid ending}.

\section{Benchmark Analysis}
\label{sec:benchmark}

\subsection{Calibration}
\label{sec:calibration}

\begin{figure}[tbp]
\centering
\includegraphics[width=\linewidth]{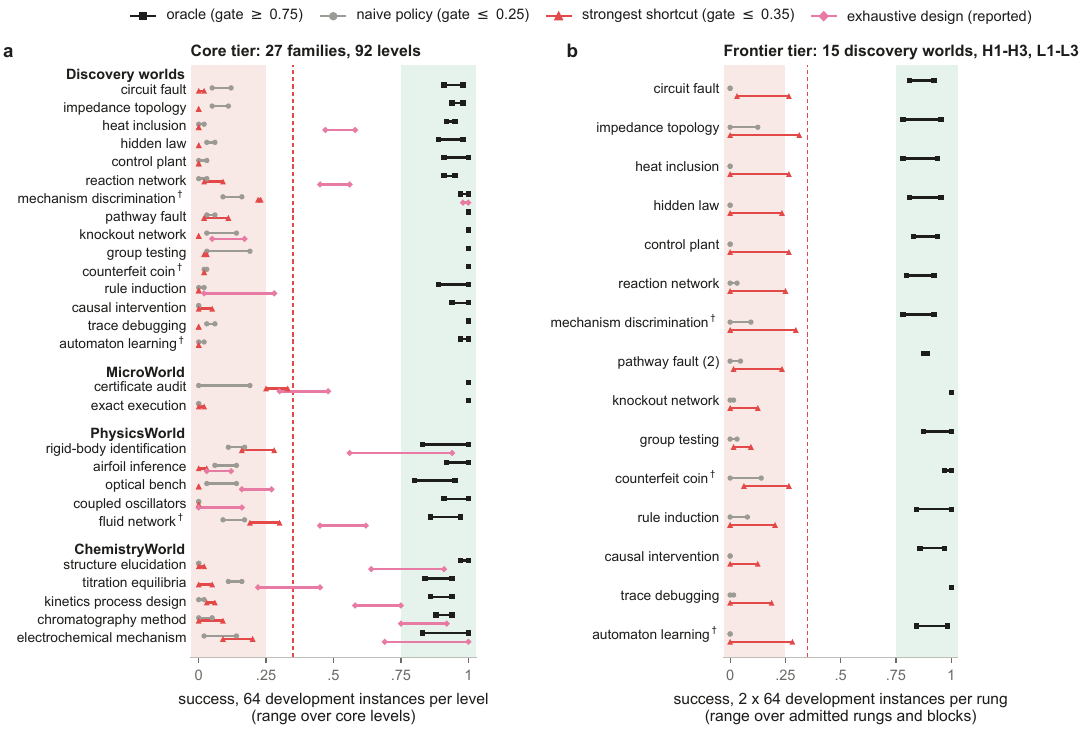}
\caption{\capt{The calibration gate on both tiers.} Each bar spans a policy's success over a family's levels, on 64 development instances per level: the core levels (a; first of two blocks for the PhysicsWorld and ChemistryWorld families) and the admitted frontier rungs over both development blocks (b; pathway fault has two rungs). Shaded bands mark the oracle gate (green, $\geq 0.75$) and the naive gate (red, $\leq 0.25$); the dashed line is the shortcut gate (0.35). Per-level values are in \cref{app:calibration}. $^\dagger$ held out of all training.}
\label{fig:calsum}
\end{figure}

All 178 levels pass the gate (\cref{fig:calsum}).
On the 92 core levels, the oracle succeeds on 0.80--1.00 of development instances, the naive policy on at most 0.19 and the strongest shortcut on at most 0.33; the ten PhysicsWorld and ChemistryWorld families also pass on a second independent block.
On the 86 frontier rungs, every rung passes on two blocks: the oracle succeeds on 0.78--1.00, the naive policy on at most 0.14, every gated shortcut on at most 0.31, and copying the grammar's example answer on none.

\paragraph{The exhaustive design separates experiment choice from analysis.}
Where it fails, the evidence has to be chosen: knockout network (0.05--0.17), rule induction (0.02--0.28), airfoil inference (0.03--0.12), coupled oscillators (0.00--0.16) and the optical bench (0.16--0.27) cannot be solved by running experiments in a fixed order.
Where it succeeds often, the bottleneck is the analysis: mechanism discrimination (0.98--1.00), chromatography method (0.75--0.92), electrochemical mechanism (0.69--1.00) and structure elucidation (0.64--0.91).
Families in between, such as heat inclusion (0.47--0.58) and reaction network (0.45--0.56), need both, and each datasheet reports where its family lies, so that a study of experiment choice can weight the families where the exhaustive design fails.

\paragraph{Campaigns cannot be solved one stage at a time.}
On every long-horizon rung, the stage-by-stage policy, which runs the family's single-question oracle afresh at each stage with the pooled budget split evenly, stays below the shortcut gate (at most 0.27), while the oracle that reuses earlier records clears 0.78.
On L3 the oracle without notes also stays below the gate.
The campaigns therefore reward carrying evidence forward, which no core level tests.

\subsection{Difficulty and headroom}
\label{sec:difficulty}

\begin{table}[tbp]
\centering
\caption{\capt{Core levels and the two frontier ladders on the discovery worlds.} Flash: \flash{} (\snapshot{}). Success on final instances, pooled over the levels or rungs of each row; the core row is the 360 instances (8 per family and level) that every system played. Concise-9B was evaluated on the core tier only. On the frontier rows, the 27B column covers the 8 instances per family and rung that it played (\nSharedH{} and \nSharedL{}). Per-rung values are in \cref{tab:main}.}
\label{tab:tiers}
\footnotesize
\setlength{\tabcolsep}{4.2pt}
\begin{tabular}{@{}llcccccccL{3.5cm}@{}}
\toprule
\textbf{\sffamily Ladder} & \textbf{\sffamily Rungs} & \textbf{\sffamily Families} & \textbf{\sffamily Levels} & \textbf{\sffamily $n$} & \textbf{\sffamily Oracle} & \textbf{\sffamily Flash} & \textbf{\sffamily 27B} & \textbf{\sffamily 9B} & \textbf{\sffamily What makes it hard} \\
\midrule
Core & d0--d2 & 15 & 45 & 360 & 0.978 & 0.583 & 0.358 & 0.094 & one hidden system, one commit \\
Harder & H1--H3 & 14 & 42 & 938 & 0.916 & 0.367 & 0.083 & -- & larger or more structured hypotheses, deeper inference \\
Long-horizon & L1--L3 & 15 & 44 & 948 & 0.929 & 0.319 & 0.090 & -- & 3, 5 and 8+ committed stages on one system \\
\bottomrule
\end{tabular}
\end{table}

\begin{table}[tbp]
\centering
\caption{\capt{Example tasks for each rung, with the success of \flash{} (\snapshot{}) on that family's final instances} (8 per core level; 24 per frontier rung, 18 on L3). Tasks are paraphrased from the observations and simplified to fit the space. The last column names the main difficulty driver.}
\label{tab:examples}
\footnotesize
\setlength{\tabcolsep}{4pt}
\begin{tabular}{@{}llL{6.3cm}cL{3.4cm}@{}}
\toprule
\textbf{\sffamily Rung} & \textbf{\sffamily Family} & \textbf{\sffamily Task (simplified)} & \textbf{\sffamily Flash} & \textbf{\sffamily Main difficulty driver} \\
\midrule
d0 & group testing & ``Find the one contaminated sample among 32 with at most 5 pooled tests.'' & 1.00 & a halving plan at the information limit \\
d2 & rule induction & ``Identify a hidden rule over number triples by membership tests; the rule is scored on all 1,000 triples.'' & 0.00 & a large rule grammar; tests must split it \\
H1 & control plant & ``Identify a plant with dead time and a possible zero, then commit a controller that meets a specification a zero-free design misses.'' & \exHoneCp{} & an inference step the core levels never need \\
H2 & counterfeit coin & ``Find up to two counterfeit coins among 19, each heavy or light, within the ternary weighing limit.'' & \exHtwoCc{} & deviations can cancel on one pan \\
L1 & control plant & ``Identify the actuator loop and commit its controller; then identify the outer process through that loop and tune it.'' & \exLoneCp{} & later experiments run through the agent's own commits \\
L2 & circuit fault & ``Five successive faults on one board; each repair restores one part, and the healthy parts keep their true values.'' & \exLtwoCf{} & early readings make later small drifts detectable \\
L3 & group testing & ``An outbreak on a sample rack over eight rounds; older pool results leave the visible history, and a notebook keeps what the agent writes.'' & \exLthreeGt{} & deciding what to record (L1: \gtLone{}, L2: \gtLtwo{}) \\
\bottomrule
\end{tabular}
\end{table}

For the models in \cref{tab:tiers}, the core tier and the two frontier ladders are well separated.
On the core tier the oracle succeeds on 0.978 of the 360 instances that every system played, \flash{} on 0.583 and \qbig{} on 0.358; on the harder and long-horizon rungs the oracle stays at \oMin--\oMax{} while Flash falls to the values in \cref{tab:tiers}.
Difficulty is not simply the number of decisions an episode takes.
The examples in \cref{tab:examples} show that the driver is the kind of work required: core levels need one deduction at the information limit, harder rungs need an inference step the core levels never ask for, and campaigns need evidence that was gathered, and in L3 written down, stages earlier.
Group testing makes the point: Flash solves \gtLtwo{} of its five-stage outbreak campaigns (L2) and \gtLthree{} of its eight-stage ones, once older results leave the window.

\section{Experiment Setup}
\label{sec:setup}

We evaluate three kinds of systems on the withheld final instances:
\begin{itemize}
  \item \textbf{Scripted references}: the gate's oracle and naive policy, which read only the public observation and play the same instances as the models.
  \item \textbf{Local models}: \ours{}, which is \base{} \citep{qwen2026qwen35} after a post-training stage that shortens its reasoning at matched accuracy \citep{lorup2026concise}, and \qbig{} \citep{qwen2026qwen38}, a reasoning model of the same family. Both run locally with thinking on, a 4,096-token thinking cap per decision with a forced close, and sampling at temperature 1.0, top-$p$ 0.95 and top-$k$ 20.
  \item \textbf{API model}: \flash{} \citep{deepseek2026v41flash}, a fast frontier model, in its \snapshot{}, served under the \texttt{deepseek-flash} API name since 17 October 2026, with thinking on at the provider's default reasoning effort (``high''; \citealp{deepseek2026thinking}) and a limit of 16,384 tokens per call; on the pilot's frontier instances it also plays at its maximum effort with up to 393,216 tokens per call (\cref{sec:flashmax}).
  \item \textbf{Pilot}: \solModel{}, on \solPilotScope{}. Every call ran through Codex CLI 0.166.0-alpha.5.1 with its top preset, ``ultra'' (maximum reasoning with automatic task delegation), for which the provider's server reports reasoning effort ``xhigh'' in trace probes of the same CLI build; the call logs record the requested preset, not the server-side effort. The model also accepts a higher single-agent effort, ``max'', which the pilot did not use. Codex's multi-agent features were turned off, but this CLI build still offered the model its collaboration (sub-agent) tools; no stored call record of the pilot or the probes shows them used. Each decision sends the same messages as the API protocol in one stateless call with every other tool disabled, and the model's final message is the turn's action; no per-call token cap of ours applies, so there is no re-ask (\cref{app:eval}).
\end{itemize}

\begin{table}[tbp]
\centering
\caption{\capt{Every comparator setting and its budget.} Decision: one turn of an episode. The oracle and naive policy read only the public observation, as the models do.}
\label{tab:settings}
\footnotesize
\setlength{\tabcolsep}{4pt}
\begin{tabular}{@{}L{2.7cm}L{1.9cm}L{5.2cm}L{3.0cm}@{}}
\toprule
\textbf{\sffamily System} & \textbf{\sffamily Reasoning effort} & \textbf{\sffamily Setting and budget} & \textbf{\sffamily Instances} \\
\midrule
Scripted oracle, naive policy & -- & Deterministic policies; no model budget & every instance the models play \\
\ours{}, \nar{}, \fin{}, \qbig{} & thinking on & Local; at most 4,096 thinking tokens per decision, then a forced close; temperature 1.0, top-$p$ 0.95, top-$k$ 20; context 20,480 tokens & core tier, 24 per family and level; \qbig{} also 8 per family and frontier rung \\
\flash{} (\snapshot{}) & ``high'', the provider's default & API; thinking on; at most 16,384 tokens per call. Core tier, PhysicsWorld and ChemistryWorld: a call cut at the budget returns no action. Frontier tier: one thinking-off re-ask that carries the cut reasoning. \capCutPct\% of its frontier decisions are cut at the budget & core tier, 8 per family and level; frontier tier, up to 24 per family and rung (\roundsHthree{} on H3, \roundsLthree{} on L3); PhysicsWorld and ChemistryWorld (\cref{tab:othergroups}) \\
\flash{} (\snapshot{}), maximum budget & ``max'' & API; thinking on; at most 393,216 tokens per call, no re-ask; no call reached the limit & $o = 0$ of each family on every frontier rung, the pilot's frontier instances (\cref{tab:flashmax}) \\
\solname{} (pilot and probes) & Codex's top preset, ``ultra''; server-reported xhigh, below the single-agent ``max'' & Codex CLI 0.166.0-alpha.5.1; multi-agent features off, collaboration tools still offered, no use found in the call records; no per-call token cap of ours, and no call truncated in \solAuditCalls{} calls (largest \solAuditMaxTok{} output tokens); every tool disabled; one stateless call per decision; not the API protocol & one final instance per family and rung or level; 8 or 16 development instances per candidate rung \\
\midrule
\multicolumn{4}{@{}l}{\emph{Laboratory worlds (\cref{sec:labrl}; \cref{app:labrl})}} \\
\warm{}, \labrl{}, \mixrl{} & thinking on & Local; panel: at most \lrTurnCap{} tokens per turn (\lrPanelTurnCapMc{} in method comparison) and \lrEpisodeCap{} per episode, temperature 1.0 without truncation; P+X and workflow: at most \lrEvalMaxTokens{} tokens per call, temperature \lrEvalTemp{}, top-$p$ \lrEvalTopP{} & the \lrPanelWorlds{} panel worlds, \lrG{} episodes each; P+X and workflow development items, one attempt each \\
\flash{} & ``low'' (reference) and ``high'' on the panel; ``high'' otherwise & API; thinking on; panel: the local models' tools, prompts, verifier and per-turn caps, counted in their tokenizer; P+X and workflow: at most \lrFlashMaxTokens{} tokens per call & as for the local models \\
\bottomrule
\end{tabular}
\end{table}

\Cref{tab:settings} lists every comparator setting with its budget.

\paragraph{Implementation details.}
Every system sees the same observations and plays under the protocol of \cref{sec:episode}.
On the frontier tier, a turn whose API call reaches the token limit gets one re-ask with thinking off that carries the cut turn's reasoning and asks for the action line only, and the reply is the turn's action.
Like the local models' forced close, the re-ask thus keeps the thinking so far in context.
\flash{}'s thinking reaches the limit on \capCutPctMin--\capCutPctMax\% of frontier-tier decisions per rung, so the per-call budget shapes these episodes (\cref{sec:budget}).
A frontier episode may take as many decisions as its lab-action turns plus Python runs.
Local models run at a context length of 20,480 tokens; on the frontier tier, the longest prompt in the gate's oracle and naive transcripts is 5,580 tokens.
\qbig{} plays every frontier rung; \ours{} plays the core tier, where it is the starting point of the training study.

\subsection{Evaluation metrics}
\label{sec:metrics}

\paragraph{Success.}
Let $\mathcal{I}$ be a set of final instances and $r_i \in \{0, 1\}$ the verifier's reward for a system's episode on instance $i$.
A system's success on $\mathcal{I}$ is
\begin{equation}
  S(\mathcal{I}) = \frac{1}{|\mathcal{I}|}\sum_{i \in \mathcal{I}} r_i ,
\end{equation}
pooled over families with equal instance counts per family and level, so that no family dominates.
On campaigns, $r_i = 1$ only if every stage commit and the final commit are right; per-stage verdicts are kept for analysis.

\paragraph{Completed-episode success.}
We also report success among episodes that do not end invalid, $S_c = \sum_i r_i / \sum_i v_i$ with $v_i = 1$ for a valid ending, which separates reasoning from failures to produce an answer.

\paragraph{Paired differences and intervals.}
Because every system plays the same instances, two systems $A$ and $B$ are compared by the paired difference
\begin{equation}
  \Delta_{A,B} = \frac{1}{|\mathcal{I}|}\sum_{i \in \mathcal{I}} \bigl(r_i^A - r_i^B\bigr),
\end{equation}
with a 95\% percentile interval from 10,000 bootstrap resamples of instances; single rates carry Wilson intervals.
Intervals describe these families and instances, not new ones.
A difference is \emph{resolved} when its interval excludes zero, and we say a system beats another only then.
Because episodes of one family may be correlated, \cref{app:cluster} repeats every headline comparison with a bootstrap over whole families, computed exactly by enumerating every resample (no Monte Carlo error); all stay resolved except Flash minus the 27B on L3 alone, the held-out training gains of both models, which rest on three families, and the pilot model's gap to the oracle over all frontier rungs, which stays resolved on the long-horizon rungs.
With 72 episodes per family, a single rate near 0.5 has an interval of about $\pm 0.11$.

\paragraph{Headroom and cost.}
\emph{Headroom} is oracle success minus model success on the same instances.
For the API model we report the mean cost per episode at the provider's list prices, including cached-input discounts and the peak-hour rate for calls made in peak hours, under its capped budget of 16,384 tokens per call, and the mean number of decisions per episode.

\section{Results}
\label{sec:results}

\begin{table}[tbp]
\centering
\caption{\capt{Success on the final instances of the discovery worlds, by tier and rung.} API model: \flash{} (\snapshot{}). The core column is the 360 instances that every system played (8 per family and level). Frontier rungs: \flash{} and the oracle on 24 instances per family (the first \roundsHthree{} on H3 and \roundsLthree{} on L3, the rounds every family finished; \fN{} episodes), \qbig{} on 8 per family. Decisions: mean turns per frontier episode, for Flash over its \fEpAll{} episodes and for the 27B over its \qEpAll{}; on the instances both played, Flash averages \fDecShared{} (\cref{sec:analysis}). Cost: mean cost per episode on the core / frontier tier at the provider's list prices, including the peak-hour rate for calls made in peak hours, under the 16,384-token cap, the latter with every turn paid. The best model in each column is in bold. \flash{} minus \qbig{} on the instances both played is resolved on the core tier, \dci{+0.225}{+0.169}{+0.278}, and pooled over the frontier tier, \dPooled{} (\fShared{} instances).}
\label{tab:main}
\footnotesize
\setlength{\tabcolsep}{3.6pt}
\begin{tabular}{@{}lcccccccccc@{}}
\toprule
 & \textbf{\sffamily Core} & \multicolumn{3}{c}{\textbf{\sffamily Harder}} & \multicolumn{3}{c}{\textbf{\sffamily Long-horizon}} & \textbf{\sffamily Frontier} & \textbf{\sffamily Decisions} & \textbf{\sffamily Cost (\$)} \\
\cmidrule(lr){3-5}\cmidrule(lr){6-8}
\textbf{\sffamily System} & d0--d2 & H1 & H2 & H3 & L1 & L2 & L3 & pooled & & core / frontier \\
\midrule
\addlinespace[2pt]\multicolumn{11}{@{}l}{\emph{Scripted reference}} \\
\quad Oracle & 0.978 & 0.926 & 0.911 & 0.910 & 0.942 & 0.923 & 0.921 & 0.923 & -- & -- \\
\addlinespace[2pt]\multicolumn{11}{@{}l}{\emph{Local models (thinking $\leq$ 4,096 tokens per decision)}} \\
\quad Concise-9B & 0.094 & -- & -- & -- & -- & -- & -- & -- & -- & -- \\
\quad Qwen3.8-27B & 0.358 & 0.161 & 0.036 & 0.054 & 0.117 & 0.107 & 0.045 & 0.087 & 19.6 & -- \\
\addlinespace[2pt]\multicolumn{11}{@{}l}{\emph{API model (thinking on, $\leq$ 16,384 tokens per call)}} \\
\quad DeepSeek V4.1 Flash & \textbf{0.583} & \textbf{0.491} & \textbf{0.321} & \textbf{0.267} & \textbf{0.467} & \textbf{0.318} & \textbf{0.107} & \textbf{0.343} & 19.7 & 0.037 / 0.132 \\
\bottomrule
\end{tabular}
\end{table}

\begin{figure}[tbp]
\centering
\includegraphics[width=\linewidth]{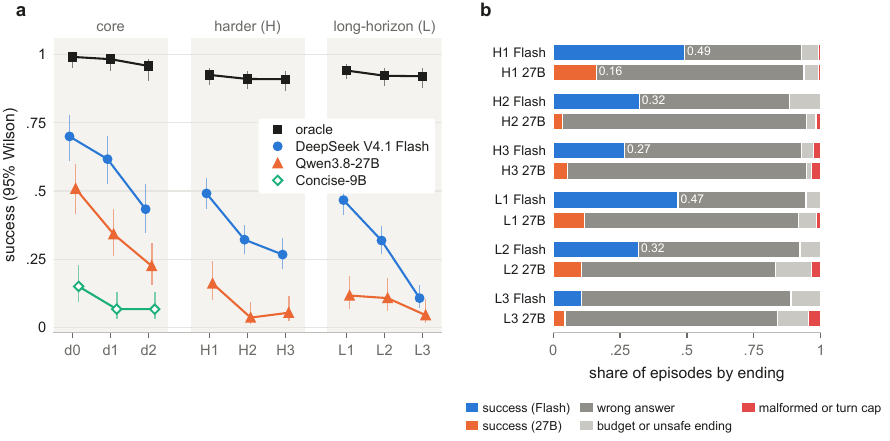}
\caption{\capt{Success by rung and how frontier episodes end.} (a) Pooled success on the final instances of the discovery worlds, with 95\% Wilson intervals; markers are offset slightly so that intervals do not overlap. (b) The share of each frontier rung's episodes that succeed, commit a wrong answer, end on a stated budget or safety rule, or end on malformed replies or the turn cap.}
\label{fig:tiers}
\end{figure}

\paragraph{The core tier separates the local models and approaches saturation for frontier models.}
On the core tier's 360 shared instances, \flash{} succeeds on 0.583 and \qbig{} on 0.358, a resolved difference of \dci{+0.225}{+0.169}{+0.278}, against 0.978 for the oracle and 0.094 for \ours{} (\cref{tab:main}).
Flash ends 0.37 of these episodes invalid, almost all on two turns in a row at its token limit; of the episodes it completes, it solves 0.925.
These core-tier rates are confounded with Flash's per-call budget and may understate it: \cutCorePct\% of its core decisions are cut at the cap, and episodes with a cut turn succeed on \cutCoreWith{} against \cutCoreWithout{} without one; the same holds on PhysicsWorld (\cutPhysPct\% of decisions cut over all \physCutEpisodes{} finished episodes; \cutPhysWith{} against \cutPhysWithout) and ChemistryWorld (\cutChemPct\%; \cutChemWith{} against \cutChemWithout).
When it finishes, rule induction (0.43 of completed episodes) and pathway fault (0.67) hold it back most.
On all 1,080 core instances of the discovery worlds, the 27B succeeds on 0.358 of the trained families' episodes and 0.394 of the held-out families', against an oracle at 0.964 and 0.995, a headroom of about 0.60; it ends fewer than 0.05 of them invalid, so that headroom is not a format effect (\cref{app:baselines}), though the 27B's thinking is near its 4,096-token cap on most decisions (a mean of \qThinkMin--\qThinkMax{} tokens per decision).
In the simulator-grounded PhysicsWorld and ChemistryWorld families, the 27B succeeds on \qPhysFinal{} and \qChemFinal{} of final instances, against \oraclePhysFinal{} and \oracleChemFinal{} for the scripted oracle on the same instances (\cref{tab:othergroups}).
Flash, in the core tier's setting, is also far below the oracle on both groups.
On the instances it played (\fpPhysN{} PhysicsWorld episodes, the first \fpPhysRounds{} of each level's 24, and all \fpChemN{} ChemistryWorld episodes), it succeeds on \fpPhysFlash{} and \fpChemFlash{}, against \fpPhysOracle{} and \fpChemOracle{} for the oracle, paired differences of \fpPhysMinusOracle{} and \fpChemMinusOracle{}; it is above the 27B by \fpPhysMinusQ{} and \fpChemMinusQ{} on the same instances.
Most of its failures are invalid endings: \fpPhysInvalid{} and \fpChemInvalid{} of its episodes end on replies without a legal action, \fpPhysEmptyPct\% and \fpChemEmptyPct\% of those on two replies in a row with no content at all, as happens when a turn is cut at the per-call limit, and it solves \fpPhysCompleted{} and \fpChemCompleted{} of the episodes it completes.
The \solname{} pilot, by contrast, solves most of both groups at one instance per family and level (\cref{tab:pilot}).

\paragraph{Flash on the frontier tier, under its capped budget.}
On up to 24 instances per family and rung (\roundsHthree{} on H3, \roundsLthree{} on L3), Flash succeeds on \fTwoHone, \fTwoHtwo{} and \fTwoHthree{} of the harder rungs' episodes and on \fTwoLone, \fTwoLtwo{} and \fTwoLthree{} of the campaigns', while the oracle stays at \oMin--\oMax{} on the same instances (\cref{fig:tiers}a).
Pooled over families, each ladder is graded: Flash solves about half of its entry rung's episodes (\fTwoHone{} on H1, \fTwoLone{} on L1), and pooled success falls with every rung above it, though not in every family (\cref{tab:frontier-h,tab:frontier-l}).
Flash's completed-episode success is within \complMax{} of its success on every rung, because re-asks almost always return an action; the per-call budget still shapes these episodes, as the next paragraph shows.
On the \fShared{} instances that both models played (8 per family and rung), Flash beats the 27B by \dPooled, by \dH{} on the harder and by \dL{} on the long-horizon rungs, and the difference is resolved on every rung at the episode level, down to \dLthree{} on L3; a bootstrap over families does not resolve L3 alone, \clusterLthree{} (\cref{app:cluster}).

\paragraph{Part of the difficulty may be the per-call budget.}
\label{sec:budget}
Flash's thinking reaches the 16,384-token limit on \capCutPct\% of frontier decisions, from \capCutPctMin\% on \capCutRungMin{} to \capCutPctMax\% on \capCutRungMax{}.
Of the \capLogged{} frontier episodes with a complete decision log, \capReaskN{} end on a turn answered by the thinking-off re-ask; they succeed on \capReaskRate{} (\capReaskK{} of \capReaskN{}), against \capThinkRate{} (\capThinkK{} of \capThinkN{}) for episodes whose final decision was a full thinking turn.
Hard instances also cause long thinking, so the split is correlational and does not show how much a larger budget would recover.
We therefore read Flash's frontier results as reasoning under a fixed per-call budget; the next paragraph measures the budget effect on one instance per family and rung.
The pilot model runs without a per-call cap of ours; over its \solEpisodes{} episodes the Codex CLI reported a mean of \solTokensPerDecision{} output tokens per decision, reasoning included; token counts are not directly comparable across models, and the comparison with Flash is between a model without our cap and one under it.

\paragraph{At its maximum effort and a large per-call budget, Flash solves two thirds of the pilot's frontier instances.}
\label{sec:flashmax}
On the pilot's frontier instances, the $o = 0$ instance of each family on every rung, \flash{} at reasoning effort ``max'' with up to 393,216 tokens per call and no re-ask (\cref{tab:settings}) succeeds on \fmaxAllK{} of \fmaxAllN{} episodes (\fmaxAllRate{}, Wilson interval [\fmaxAllLo{}, \fmaxAllHi{}]), against \fmaxAllDef{} at its default effort under the 16,384-token cap, a paired difference of \fmaxAllMinusDef{} (\cref{tab:flashmax}; over families, \fmaxAllMinusDefFam{}, \cref{tab:cluster}).
The difference is \fmaxHarderMinusDef{} on the harder and \fmaxLongMinusDef{} on the long-horizon rungs, and it is resolved on every rung except H1, where capped Flash already solves \fmaxHoneDef{} of \fmaxN{}.
None of its \fmaxCalls{} calls reached the larger limit (the largest returned \fmaxMaxCallTok{} completion tokens), and every episode ended on a committed answer.
It stays below the oracle, \fmaxAllMinusOracle{} over all rungs; the gap is resolved on H3, L2 and L3 and largest on L3, \fmaxLthreeMinusOracle{}.
The pilot model stays ahead, \fmaxAllSolMinus{} over all rungs (\fmaxAllSolWins{} instances won, \fmaxAllSolLosses{} lost), against \solMinusFlash{} for capped Flash, so at its maximum effort Flash closes about two thirds of its gap to the pilot model and about half of its gap to the oracle.
Per rung, the pilot model's lead is resolved on H3, \fmaxHthreeSolMinus{}, and L2, \fmaxLtwoSolMinus{}; on the eight-stage campaigns the two solve \fmaxLthreeSol{} and \fmaxLthreeK{} of \fmaxN{}, \fmaxLthreeSolMinus{}.
At list prices the maximum setting costs from about \$\fmaxUsdMin{} per episode on \fmaxUsdMinRung{} to \$\fmaxUsdMax{} on \fmaxUsdMaxRung{}, with means of \fmaxTokMin{} to \fmaxTokMax{} completion tokens per episode.
Flash's capped frontier rates therefore understate it substantially, and much of the frontier tier's difficulty for Flash is its budget; the headroom that holds is measured against the oracle. With one instance per family and rung, these \fmaxAllN{} episodes measure the size of the budget effect on the pilot's instances, not Flash's tier-wide rates at its maximum effort, which we have not measured.

\begin{table}[tbp]
\centering
\caption{\capt{Flash at its maximum reasoning effort on the frontier tier.} The $o = 0$ final instance of each of the 14 families on every frontier rung, the frontier instances of the pilot. Max: \flash{} (\snapshot{}) at reasoning effort ``max'', at most 393,216 tokens per call, no re-ask. Default: \flash{} at its default effort under the frontier protocol (16,384 tokens per call, re-ask after a cut turn), on the same instances. Sol: the pilot of \cref{tab:pilot} (\solModel{}). Successes over instances; differences paired over instances with 95\% bootstrap intervals.}
\label{tab:flashmax}
\footnotesize
\setlength{\tabcolsep}{3.5pt}
\begin{tabular}{@{}lrcccclll@{}}
\toprule
\textbf{\sffamily Rung} & \textbf{\sffamily $n$} & \textbf{\sffamily Max} & \textbf{\sffamily Default} & \textbf{\sffamily Sol} & \textbf{\sffamily Oracle} & \textbf{\sffamily Max $-$ default} & \textbf{\sffamily Sol $-$ max} & \textbf{\sffamily Max $-$ oracle} \\
\midrule
H1 & \fmaxN{} & \fmaxHoneK{} & \fmaxHoneDef{} & \fmaxHoneSol{} & \fmaxHoneOracle{} & \fmaxHoneMinusDef{} & \fmaxHoneSolMinus{} & \fmaxHoneMinusOracle{} \\
H2 & \fmaxN{} & \fmaxHtwoK{} & \fmaxHtwoDef{} & \fmaxHtwoSol{} & \fmaxHtwoOracle{} & \fmaxHtwoMinusDef{} & \fmaxHtwoSolMinus{} & \fmaxHtwoMinusOracle{} \\
H3 & \fmaxN{} & \fmaxHthreeK{} & \fmaxHthreeDef{} & \fmaxHthreeSol{} & \fmaxHthreeOracle{} & \fmaxHthreeMinusDef{} & \fmaxHthreeSolMinus{} & \fmaxHthreeMinusOracle{} \\
L1 & \fmaxN{} & \fmaxLoneK{} & \fmaxLoneDef{} & \fmaxLoneSol{} & \fmaxLoneOracle{} & \fmaxLoneMinusDef{} & \fmaxLoneSolMinus{} & \fmaxLoneMinusOracle{} \\
L2 & \fmaxN{} & \fmaxLtwoK{} & \fmaxLtwoDef{} & \fmaxLtwoSol{} & \fmaxLtwoOracle{} & \fmaxLtwoMinusDef{} & \fmaxLtwoSolMinus{} & \fmaxLtwoMinusOracle{} \\
L3 & \fmaxN{} & \fmaxLthreeK{} & \fmaxLthreeDef{} & \fmaxLthreeSol{} & \fmaxLthreeOracle{} & \fmaxLthreeMinusDef{} & \fmaxLthreeSolMinus{} & \fmaxLthreeMinusOracle{} \\
\midrule
H1--H3 & \fmaxHarderN{} & \fmaxHarderK{} & \fmaxHarderDef{} & \fmaxHarderSol{} & \fmaxHarderOracle{} & \fmaxHarderMinusDef{} & \fmaxHarderSolMinus{} & \fmaxHarderMinusOracle{} \\
L1--L3 & \fmaxLongN{} & \fmaxLongK{} & \fmaxLongDef{} & \fmaxLongSol{} & \fmaxLongOracle{} & \fmaxLongMinusDef{} & \fmaxLongSolMinus{} & \fmaxLongMinusOracle{} \\
All & \fmaxAllN{} & \fmaxAllK{} & \fmaxAllDef{} & \fmaxAllSol{} & \fmaxAllOracle{} & \fmaxAllMinusDef{} & \fmaxAllSolMinus{} & \fmaxAllMinusOracle{} \\
\bottomrule
\end{tabular}
\end{table}

\paragraph{A stronger model solves most of the frontier tier (pilot).}
In the pilot (\solModel{}, on \solPilotScope{}; \cref{tab:pilot}), \solname{} succeeds on \solFrontierK{} of \solFrontierN{} frontier episodes (\solFrontierRate{}, Wilson interval [\solFrontierLo{}, \solFrontierHi{}]), against \flashPilotFrontierK{} for Flash, \qPilotFrontierK{} for the 27B and \oraclePilotFrontierK{} for the oracle on the same instances.
Against Flash capped at 16,384 tokens per call at its default effort, the difference is \solMinusFlash{}, with \solFlashWins{} paired instances won and \solFlashLosses{} lost, a gap that mixes model strength with Flash's budget: against Flash at its maximum effort on the same instances it shrinks to \fmaxAllSolMinus{}, about a third of it (\cref{sec:flashmax}). It exceeds the 27B by \solMinusQ{} and stays below the oracle, \solMinusOracle{}, a gap that a bootstrap over families, which allows for correlated rungs of one family, does not resolve (\solMinusOracleFam{}; \cref{tab:cluster}).
It solves \solRungKMin{}--\solRungKMax{} of the \solRungN{} instances on each rung from H1 to L2, and on the harder rungs together its difference from the oracle is not resolved, \solHarderMinusOracle{}.
On the long-horizon rungs it is below the oracle, \solLongMinusOracle{}, and the bootstrap over families resolves that gap (\solLongMinusOracleFam{}); it comes mostly through L3, where it solves \solLthreeK{} of \solLthreeN{} campaigns against \oracleLthreeK{} for the oracle, \solLthreeMinusOracle{}.
It also solves all \solCoreN{} core-level instances of the discovery worlds, against \flashPilotCoreK{} for Flash, and \solPhysK{} of \solPhysN{} PhysicsWorld and \solChemK{} of \solChemN{} ChemistryWorld instances, against \flashPilotPhysK{} and \flashPilotChemK{} for Flash.
A stronger model at xhigh reasoning effort therefore solves most of the frontier tier, and Flash at its maximum effort solves \fmaxAllRate{} of the same instances; the headroom that remains is in the eight-stage campaigns and in the gap to the oracle.
With one instance per family and rung, these are pooled estimates with wide intervals, and we draw no per-family conclusion from them.

\begin{table}[tbp]
\centering
\caption{\capt{Pilot: \solModel{}, on \solPilotScope{}.} Successes over instances (rate) for each system on the same final instances; Sol: Wilson 95\% interval for \solname{}. Flash: \flash{} (\snapshot{}) under the API protocol of \cref{sec:setup}, on PhysicsWorld and ChemistryWorld from its own final-instance run; 27B: \qbig{}. The oracle is shown on the frontier tier. One pathway-fault L1 instance that could not be reproduced on the pilot's machine is excluded. Sol minus Flash (capped at 16,384 tokens per call, default effort) on the frontier tier: \solMinusFlash{}; minus the 27B: \solMinusQ{}; minus the oracle: \solMinusOracle{} (family bootstrap: \solMinusOracleFam{}; \cref{tab:cluster}).}
\label{tab:pilot}
\footnotesize
\setlength{\tabcolsep}{5pt}
\begin{tabular}{@{}lrccccc@{}}
\toprule
\textbf{\sffamily Cut} & \textbf{\sffamily $n$} & \textbf{\sffamily Sol} & \textbf{\sffamily Sol, 95\% CI} & \textbf{\sffamily Flash} & \textbf{\sffamily 27B} & \textbf{\sffamily Oracle} \\
\midrule
Frontier tier, all rungs & 84 & 71/84 (0.85) & [0.75, 0.91] & 32/84 (0.38) & 7/84 (0.08) & 79/84 (0.94) \\
\quad H1 & 14 & 12/14 (0.86) & [0.60, 0.96] & 8/14 (0.57) & 4/14 (0.29) & 12/14 (0.86) \\
\quad H2 & 14 & 12/14 (0.86) & [0.60, 0.96] & 7/14 (0.50) & 0/14 (0.00) & 13/14 (0.93) \\
\quad H3 & 14 & 13/14 (0.93) & [0.69, 0.99] & 4/14 (0.29) & 1/14 (0.07) & 13/14 (0.93) \\
\quad L1 & 14 & 12/14 (0.86) & [0.60, 0.96] & 6/14 (0.43) & 1/14 (0.07) & 14/14 (1.00) \\
\quad L2 & 14 & 14/14 (1.00) & [0.78, 1.00] & 5/14 (0.36) & 1/14 (0.07) & 14/14 (1.00) \\
\quad L3 & 14 & 8/14 (0.57) & [0.33, 0.79] & 2/14 (0.14) & 0/14 (0.00) & 13/14 (0.93) \\
\addlinespace
Core levels, discovery worlds & 45 & 45/45 (1.00) & [0.92, 1.00] & 31/45 (0.69) & 18/45 (0.40) & -- \\
PhysicsWorld core levels & 20 & 17/20 (0.85) & [0.64, 0.95] & 1/20 (0.05) & 0/20 (0.00) & -- \\
ChemistryWorld core levels & 15 & 15/15 (1.00) & [0.80, 1.00] & 6/15 (0.40) & 5/15 (0.33) & -- \\
\bottomrule
\end{tabular}
\end{table}

\begin{figure}[tbp]
\centering
\includegraphics[width=\linewidth]{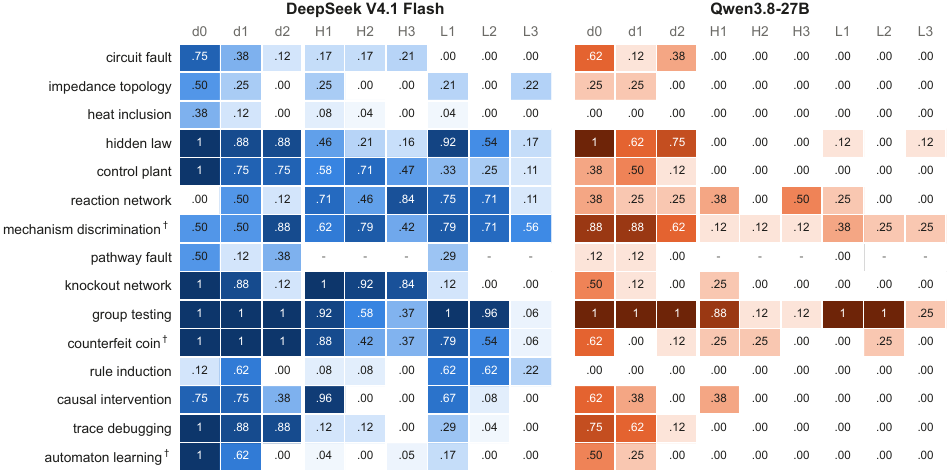}
\caption{\capt{Success per family and rung.} Final instances: 8 per cell on the core levels; on the frontier rungs 24 per cell for \flash{} (\roundsHthree{} on H3, \roundsLthree{} on L3) and 8 for \qbig{}. A dash marks a rung that does not exist or was not evaluated: pathway fault has only L1 and L3, and its L3 is not evaluated here. Every cell's oracle success is in \cref{tab:frontier-h,tab:frontier-l}. $^\dagger$ held out of all training.}
\label{fig:heatmap}
\end{figure}

\paragraph{Long memory is the hardest axis.}
On L3, Flash succeeds on \fLthreeK{} of \fLthreeN{} episodes and the 27B on \qLthreeK{} of \qLthreeN{}, against an oracle at \oLthree{}.
The L3 campaigns are not harder per stage than L2: they are longer, and older records leave the window.
Several families that the models handle at L1 or L2 collapse at L3, including group testing (Flash \gtLone{} at L1, \gtLtwo{} at L2 and \gtLthree{} at L3) and counterfeit coin (Flash \ccLone{} at L1, \ccLthree{} at L3; \cref{fig:heatmap}).
L3 is also the rung where the pilot's stronger model falls furthest below the oracle (\cref{tab:pilot}).

\paragraph{How frontier episodes end.}
Across the frontier tier, \fWrongMin--\fWrongMax\% of Flash's failures per rung are committed wrong answers, and most of the rest end on a safety rule that the observation states (\cref{fig:tiers}b).
Only \fMalformed{} of its \fN{} frontier episodes, and \qMalformed{} of the 27B's \fShared{}, end on malformed replies or the turn cap.
Because more than a quarter (\capCutPct\%) of Flash's decisions are cut at its per-call budget (above), these endings describe its episodes under that budget and do not by themselves separate reasoning errors from truncation.
Rule endings concentrate in automaton learning, where a query beyond the remaining symbol allowance ends the episode, as the observation says (\fUnsafeAL{} of Flash's \fUnsafe{} unsafe endings).

\paragraph{Flash runs about as many experiments as the 27B and succeeds far more often.}
On the instances both played, Flash and the 27B take almost the same number of decisions per frontier episode (\fDecShared{} and \qDec) and run similar numbers of experiments on most rungs (for example \fExpHone{} and \qExpHone{} on H1, \fExpLtwo{} and \qExpLtwo{} on L2), yet Flash succeeds \fTimesQ{} times as often.
The difference lies mostly in what each model infers from comparable evidence, under different budgets: the 27B's thinking averages \qThinkFrontierMin--\qThinkFrontierMax{} of its 4,096 tokens per decision on the frontier tier.
Flash spends \fTokMin--\fTokMax{} completion tokens per frontier episode, at a mean cost of about \$\fUsd{} per episode at list prices (about \$\fUsdHone{} on H1 and \$\fUsdLthree{} on L3), against \$\fUsdCore{} per core episode.

\paragraph{Families differ more than tiers.}
Per family, frontier difficulty ranges from rungs that Flash solves on more than nine in ten instances (\famMost) to \famRareN{} families where it succeeds on at most \famRareMaxK{} of its \famRareMinN{} frontier episodes (\famRare; \cref{fig:heatmap}).
The 27B succeeds on no frontier episode in eight of the 15 families, including circuit fault and heat inclusion, which the oracle solves on most of the same instances (\cref{tab:frontier-h,tab:frontier-l}).

\section{Long-Horizon Campaigns}
\label{sec:longcampaigns}

The frontier tier shows that a stronger model at xhigh reasoning effort solves most single questions and short campaigns but falls well below the oracle on the eight-stage ones (\solLthreeK{} of \solLthreeN{} against \oracleLthreeK{} of \solLthreeN{}).
This section probes longer campaigns built by the same construction and admitted by the same gate.
Every result here is a probe on \emph{development} instances with the pilot model (\solModel{}), eight or sixteen per rung, with the scripted oracle on the same instances, except two coupled rungs that were later also played on final instances (\cref{sec:coupled}); the rungs are candidates, not part of the evaluated suite.

\subsection{Which rungs hold a stronger model}
\label{sec:ladder}

Because the pilot model solves most of the frontier tier, we probed candidate rungs above it with the same model and protocol on \emph{development} instances, eight per rung (sixteen for rungs near the band edge), with the scripted oracle on the same instances (\cref{tab:ladder}).
Every candidate first passed the calibration gate of \cref{sec:pipeline} on two blocks of 64 development instances.
The protocol extends a rung at 3--5 of 8 by eight more instances and judges it on all sixteen.
This was done for the rungs of \cref{tab:ladder} and for three long campaigns of \cref{tab:longcampaigns}; four later rungs at 3--5 of 8 (heat inclusion L5, drug pipeline C2 and C3, and instrument-in-the-loop C3) were not extended and are reported at eight instances, so their status is less certain.
A rung holds the pilot model when it solves at most 0.4 of the probe instances, the target band; a rung solved on more than 0.6 is not admitted, and one between 0.4 and 0.6 holds it only partly.
The protocol also prescribes a pre-specified harder setting and a new probe for a partly-holding rung; only hidden law L4 was rebuilt that way (its 14-law version in \cref{sec:lx}), and the other partly-holding rungs are reported as probed.

\paragraph{Long campaigns hold the stronger model.}
Two long-horizon rungs of counterfeit coin (L4, 13 batches from 6 mints; L5, 20 batches from 10 mints) and an L5 campaign of hidden law (a 16-law cascade whose last answer needs constants of early laws) are admitted by the gate with the oracle at \ladHoldGateMin{}--\ladHoldGateMax{} per block, and they hold the pilot model at \ladHoldSolMin{} to \ladHoldSolMax{} of \ladN{} probe instances, below the oracle on the same instances on each (counterfeit coin L4 \ladCcFourDiff{}, L5 \ladCcFiveDiff{}; hidden law L5 \ladHlFiveDiff{}).
They combine an exact pooled budget, a history window of about a stage and a half, a notebook just large enough for a consolidated plan, and late questions about early facts that no later stage can re-measure.
Hidden law L4, with 12 laws, is solved on \ladHlFourSol{} of \ladHlFourN{}, above the target band; its harder version with 14 laws is in \cref{sec:lx}.

\paragraph{Added hidden structure does not.}
Rungs that add hidden structure to a single question, such as factor combinations read through a drifting instrument (hidden law H4 and H5) or a stated number of faulty assay results (group testing H4 and H5, with one or two wrong results), pass the gate but are solved on \ladStructSolMin{} or \ladStructSolMax{} of \ladN{} probe instances, so they are not admitted.
With a stated model class and a Python tool, the pilot model runs the oracle's own exhaustive structure search and sequential design; without the tool, it designs error-correcting pools unaided.

Existing frontier rungs can be probed the same way: rule induction's H1 and H2 hold the pilot model only partly and its H3 holds it (\cref{tab:ladder}).

\paragraph{Failures are memory and planning, not format.}
An audit of all \audFailures{} failed probe episodes of the candidate rungs found no invalid ending; \audMalformed{} malformed replies cost a turn and no budget.
The audit includes a first version of the counterfeit-coin L5 rung, which the pilot model solved on \ccFiveVoneSol{} of \ccFiveVoneN{} probe instances, above 0.6, so it was not admitted and was redesigned as the L5 rung of \cref{tab:ladder}.
The \audLong{} failures in long campaigns all used the whole pooled budget, and \audFewerNotes{} of them wrote fewer notes than the notebook holds; the other \audFullNotebook{} filled it only from turn \audFullFirstTurn{} on, after the readings they needed had left the window.
They re-derived forgotten facts until the budget ran out or carried early constants too loosely to answer the late questions.
The \audStructure{} failures on the structure rungs are ties: the pilot model's own record left two explanations equally likely, and it chose the wrong one.
Long campaigns that make memory and planning bind are therefore where frontier headroom remains; these probes are development-instance evidence about candidate rungs, not results on the evaluated suite.

\begin{table}[tbp]
\centering
\caption{\capt{Candidate rungs above the frontier tier, probed on development instances} (not final instances) with \solModel{}. Gate: the oracle's success on the two calibration blocks of 64 development instances; every candidate is admitted by the gate. Sol and Oracle: successes on the same probe instances. Holds: the pilot model solves at most 0.4 of them (the target band); partly: between 0.4 and 0.6; not admitted: more than 0.6. Rule induction's existing rungs were probed on two blocks of eight. Differences are paired over the probe instances with 95\% bootstrap intervals.}
\label{tab:ladder}
\footnotesize
\setlength{\tabcolsep}{4pt}
\begin{tabular}{@{}llcccll@{}}
\toprule
\textbf{\sffamily Family} & \textbf{\sffamily Rung} & \textbf{\sffamily Gate (oracle)} & \textbf{\sffamily Sol} & \textbf{\sffamily Oracle} & \textbf{\sffamily Sol $-$ oracle} & \textbf{\sffamily Status} \\
\midrule
\multicolumn{7}{@{}l}{\emph{Long campaigns (L4: 12--13 stages; L5: 16--20 stages)}} \\
counterfeit coin & L4 & 1.00 / 1.00 & 2/8 & 8/8 & \dci{\ensuremath{-}0.75}{\ensuremath{-}1.00}{\ensuremath{-}0.38} & holds \\
counterfeit coin & L5 & 1.00 / 1.00 & 1/8 & 8/8 & \dci{\ensuremath{-}0.88}{\ensuremath{-}1.00}{\ensuremath{-}0.63} & holds \\
hidden law & L4 & 0.92 / 0.97 & 8/16 & 14/16 & \dci{\ensuremath{-}0.38}{\ensuremath{-}0.69}{\ensuremath{-}0.06} & partly \\
hidden law & L5 & 0.84 / 0.80 & 0/8 & 6/8 & \dci{\ensuremath{-}0.75}{\ensuremath{-}1.00}{\ensuremath{-}0.38} & holds \\
\addlinespace
hidden law & H4 & 0.92 / 0.91 & 8/8 & 8/8 & \dci{0.00}{0.00}{0.00} & not admitted \\
hidden law & H5 & 0.92 / 0.80 & 8/8 & 8/8 & \dci{0.00}{0.00}{0.00} & not admitted \\
group testing, one wrong result & H4 & 1.00 / 0.98 & 7/8 & 8/8 & \dci{\ensuremath{-}0.13}{\ensuremath{-}0.38}{0.00} & not admitted \\
group testing, one wrong result & H5 & 0.98 / 0.94 & 8/8 & 8/8 & \dci{0.00}{0.00}{0.00} & not admitted \\
group testing, two wrong results & H4 & 1.00 / 1.00 & 7/8 & 8/8 & \dci{\ensuremath{-}0.13}{\ensuremath{-}0.38}{0.00} & not admitted \\
group testing, two wrong results & H5 & 0.91 / 0.95 & 7/8 & 7/8 & \dci{0.00}{\ensuremath{-}0.38}{+0.38} & not admitted \\
\addlinespace[2pt]\multicolumn{7}{@{}l}{\emph{Existing frontier rungs}} \\
rule induction & H1 & 0.91 / 0.92 & 8/16 & 14/16 & \dci{\ensuremath{-}0.38}{\ensuremath{-}0.69}{\ensuremath{-}0.06} & partly \\
rule induction & H2 & 0.89 / 0.91 & 7/16 & 14/16 & \dci{\ensuremath{-}0.44}{\ensuremath{-}0.69}{\ensuremath{-}0.13} & partly \\
rule induction & H3 & 0.84 / 0.91 & 4/16 & 14/16 & \dci{\ensuremath{-}0.63}{\ensuremath{-}0.88}{\ensuremath{-}0.38} & holds \\
\bottomrule
\end{tabular}
\end{table}

\subsection{Long campaigns for \lcWorlds{} discovery worlds}
\label{sec:lx}
Following that pattern, long-campaign rungs extend \lcOtherFamilies{} more discovery worlds, and hidden law gains a harder L4 with 14 laws, \lcFamilies{} families in all (L4 with 12 stages, 14 for hidden law, and L5 with 16, or four and five epochs where a family changes regime): one persistent hidden system, a pooled budget at the oracle's exact need, a window of about one and a half stages, a notebook just large enough for a consolidated plan, and a final question about early facts that no later stage can re-measure, such as the original system before an announced change, a part no longer on the bench, or the first version of a program.
All \lcRungs{} rungs pass the calibration gate on both development blocks, with the oracle at \lcOracleMin{}--\lcOracleMax{} and every gated shortcut, including each family's coupling shortcut, at most \lcShortcutMax{}.
Of the \lcProbedRungs{} rungs probed with the pilot model, \lcHold{} hold it (at most 0.4 of the probe instances), \lcPartly{} hold it only partly (0.4--0.6) and \lcNotAdmitted{} are not admitted (more than 0.6), so harder versions of those families are needed (\cref{tab:longcampaigns}); probes on the \lcUnprobedRungs{} rungs of \lcUnprobed{} are not reported here.

\subsection{Coupled multi-engine campaigns}
\label{sec:coupled}
The coupled campaigns chain the suite's simulators into one project.
A chemical-process campaign carries a compound from titration and kinetics through chromatography to a charge plan, and a drug-pipeline campaign carries a candidate from binding and potency assays through pharmacokinetics to a dose, each stage computed by the same laboratory engine and each later stage depending on numbers committed or measured earlier.
Their rungs C1--C4 pass the gate with the oracle at \ccOracleMin{}--\ccOracleMax{} per block and every gated shortcut at most \ccShortcutMax{}, including a policy that uses only one engine and one that re-measures instead of remembering.
The pilot model solves most of the shorter rungs; the longest drug-pipeline rung (C3) holds it at \drugThreeSol{} of \drugThreeN{} instances, against \drugThreeOracle{} for the oracle, a difference that eight instances do not resolve (\drugThreeDiff{}), and the four-lot chemical-process rung (C4) at \chemFourSol{} of \chemFourN{}, against \chemFourOracle{} (\chemFourDiff{}).
An instrument-in-the-loop campaign, in which an optics calibration feeds a spectrophotometric titration and kinetics, holds it only partly, at \ilSol{} of \ilN{} (\ilDiff{}).
These two rungs and drug pipeline C2 were probed on eight instances only (\cref{sec:ladder}); drug pipeline C3 and the instrument-in-the-loop rung were later also played by the pilot model, with the same rung code, on 32 final instances each, outside the evaluated suite: it solves \drugThreeFinalSol{} of 32 drug-pipeline campaigns, against \drugThreeFinalOracle{} for the oracle (\drugThreeFinalDiff{}), inside the target band, and \ilFinalSol{} of 32 instrument-in-the-loop campaigns, against \ilFinalOracle{} (\ilFinalDiff{}), in the partly-holding band, consistent with their probe status.
All \lcAuditFailures{} failed probe episodes of the long and coupled campaigns in \cref{tab:longcampaigns} ended on a committed answer, none on an invalid ending; reading \lcAuditRead{} of them (every coupled-campaign failure and up to ten per long-campaign probe) found, as in \cref{sec:ladder}, values that left the window without being noted, or were noted without their numbers, and no format failure.
Long campaigns in which memory and planning bind are therefore where the strongest model we tested still fails; these probes show the direction of the ladder, not results on the evaluated suite.

\begin{table}[tbp]
\centering
\caption{\capt{Long-horizon campaigns probed on development instances} (not final instances) with \solModel{}, eight instances per rung (sixteen where a rung near the band edge was probed again; four rungs at 3--5 of 8, heat inclusion L5, drug pipeline C2 and C3 and instrument-in-the-loop C3, were not probed again, \cref{sec:ladder}). Gate: the oracle's success on the two calibration blocks of 64 development instances; every rung is admitted by the gate. Sol and Oracle: successes on the same probe instances, pooled over two blocks of eight where a rung near the band edge was probed again; differences paired with 95\% bootstrap intervals. Status as in \cref{tab:ladder}.}
\label{tab:longcampaigns}
\footnotesize
\setlength{\tabcolsep}{4pt}
\begin{tabular}{@{}llcccll@{}}
\toprule
\textbf{\sffamily Family} & \textbf{\sffamily Rung} & \textbf{\sffamily Gate (oracle)} & \textbf{\sffamily Sol} & \textbf{\sffamily Oracle} & \textbf{\sffamily Sol $-$ oracle} & \textbf{\sffamily Status} \\
\midrule
\multicolumn{7}{@{}l}{\emph{Long campaigns on discovery worlds}} \\
automaton learning & L4 & 1.00 / 1.00 & 7/16 & 16/16 & \dci{\ensuremath{-}0.56}{\ensuremath{-}0.81}{\ensuremath{-}0.31} & partly \\
automaton learning & L5 & 0.98 / 1.00 & 7/16 & 16/16 & \dci{\ensuremath{-}0.56}{\ensuremath{-}0.81}{\ensuremath{-}0.31} & partly \\
causal intervention & L4 & 0.97 / 0.94 & 2/8 & 8/8 & \dci{\ensuremath{-}0.75}{\ensuremath{-}1.00}{\ensuremath{-}0.38} & holds \\
causal intervention & L5 & 0.91 / 0.88 & 0/8 & 7/8 & \dci{\ensuremath{-}0.88}{\ensuremath{-}1.00}{\ensuremath{-}0.63} & holds \\
control plant & L4 & 0.83 / 0.89 & 6/16 & 14/16 & \dci{\ensuremath{-}0.50}{\ensuremath{-}0.75}{\ensuremath{-}0.19} & holds \\
control plant & L5 & 0.89 / 0.88 & 1/8 & 7/8 & \dci{\ensuremath{-}0.75}{\ensuremath{-}1.00}{\ensuremath{-}0.38} & holds \\
group testing & L4 & 1.00 / 1.00 & 2/8 & 8/8 & \dci{\ensuremath{-}0.75}{\ensuremath{-}1.00}{\ensuremath{-}0.38} & holds \\
group testing & L5 & 1.00 / 1.00 & 1/8 & 8/8 & \dci{\ensuremath{-}0.88}{\ensuremath{-}1.00}{\ensuremath{-}0.63} & holds \\
heat inclusion & L4 & 0.88 / 0.83 & 7/8 & 8/8 & \dci{\ensuremath{-}0.13}{\ensuremath{-}0.38}{0.00} & not admitted \\
heat inclusion & L5 & 0.84 / 0.84 & 4/8 & 6/8 & \dci{\ensuremath{-}0.25}{\ensuremath{-}0.63}{0.00} & partly \\
hidden law & L4 (14 laws) & 0.89 / 0.89 & 0/8 & 7/8 & \dci{\ensuremath{-}0.88}{\ensuremath{-}1.00}{\ensuremath{-}0.63} & holds \\
impedance topology & L4 & 1.00 / 1.00 & 7/8 & 8/8 & \dci{\ensuremath{-}0.13}{\ensuremath{-}0.38}{0.00} & not admitted \\
impedance topology & L5 & 1.00 / 0.98 & 8/8 & 8/8 & \dci{0.00}{0.00}{0.00} & not admitted \\
knockout network & L4 & 1.00 / 1.00 & 0/8 & 8/8 & \dci{\ensuremath{-}1.00}{\ensuremath{-}1.00}{\ensuremath{-}1.00} & holds \\
knockout network & L5 & 1.00 / 1.00 & 0/8 & 8/8 & \dci{\ensuremath{-}1.00}{\ensuremath{-}1.00}{\ensuremath{-}1.00} & holds \\
mechanism discrimination & L4 & 0.94 / 0.97 & 8/8 & 8/8 & \dci{0.00}{0.00}{0.00} & not admitted \\
mechanism discrimination & L5 & 0.97 / 0.98 & 6/8 & 7/8 & \dci{\ensuremath{-}0.13}{\ensuremath{-}0.50}{+0.25} & not admitted \\
rule induction & L4 & 1.00 / 1.00 & 0/8 & 8/8 & \dci{\ensuremath{-}1.00}{\ensuremath{-}1.00}{\ensuremath{-}1.00} & holds \\
rule induction & L5 & 1.00 / 1.00 & 0/8 & 8/8 & \dci{\ensuremath{-}1.00}{\ensuremath{-}1.00}{\ensuremath{-}1.00} & holds \\
trace debugging & L4 & 1.00 / 1.00 & 2/8 & 8/8 & \dci{\ensuremath{-}0.75}{\ensuremath{-}1.00}{\ensuremath{-}0.38} & holds \\
trace debugging & L5 & 1.00 / 1.00 & 0/8 & 8/8 & \dci{\ensuremath{-}1.00}{\ensuremath{-}1.00}{\ensuremath{-}1.00} & holds \\
\addlinespace[2pt]\multicolumn{7}{@{}l}{\emph{Coupled multi-engine campaigns}} \\
chemical process & C1 & 1.00 / 0.97 & 8/8 & 8/8 & \dci{0.00}{0.00}{0.00} & not admitted \\
chemical process & C2 & 1.00 / 0.98 & 7/8 & 8/8 & \dci{\ensuremath{-}0.13}{\ensuremath{-}0.38}{0.00} & not admitted \\
chemical process & C3 & 0.97 / 1.00 & 6/8 & 8/8 & \dci{\ensuremath{-}0.25}{\ensuremath{-}0.63}{0.00} & not admitted \\
chemical process & C4 & 0.95 / 0.92 & 1/8 & 8/8 & \dci{\ensuremath{-}0.88}{\ensuremath{-}1.00}{\ensuremath{-}0.63} & holds \\
drug pipeline & C1 & 1.00 / 0.98 & 7/8 & 8/8 & \dci{\ensuremath{-}0.13}{\ensuremath{-}0.38}{0.00} & not admitted \\
drug pipeline & C2 & 0.95 / 0.97 & 5/8 & 8/8 & \dci{\ensuremath{-}0.38}{\ensuremath{-}0.75}{\ensuremath{-}0.13} & not admitted \\
drug pipeline & C3 & 0.78 / 0.91 & 3/8 & 6/8 & \dci{\ensuremath{-}0.38}{\ensuremath{-}0.88}{+0.25} & holds \\
instrument in the loop & C3 & 1.00 / 1.00 & 4/8 & 8/8 & \dci{\ensuremath{-}0.50}{\ensuremath{-}0.88}{\ensuremath{-}0.13} & partly \\
\bottomrule
\end{tabular}
\end{table}

\section{Episode Analysis}
\label{sec:analysis}

We analyse the frontier episodes of Flash and the 27B to see where campaigns go wrong.

\begin{figure}[tbp]
\centering
\includegraphics[width=\linewidth]{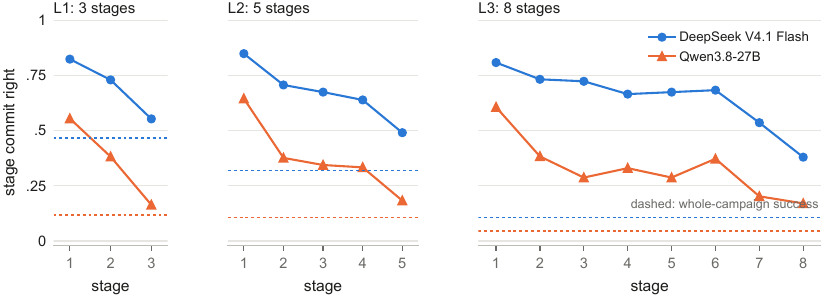}
\caption{\capt{Campaign accuracy by stage.} The share of campaigns, among those that reached their final commit, whose $k$-th stage commit is right; dashed lines mark whole-campaign success. Campaigns reaching the final commit on L1, L2 and L3: \stReachLone, \stReachLtwo{} and \stReachLthree{} for Flash (24 instances per family; 18 on L3); \stQReachLone, \stQReachLtwo{} and \stQReachLthree{} for the 27B (8 per family). L3 campaigns with a ninth stage (\stNinth{} for Flash, \stQNinth{} for the 27B) are not shown beyond stage eight.}
\label{fig:stages}
\end{figure}

\paragraph{Campaign accuracy falls stage by stage.}
Among campaigns that reach their final commit, Flash gets the first stage right in \stFirstMin--\stFirstMax{} of them on every long-horizon rung, but only \stThirdLone{} of third stages on L1, \stFifthLtwo{} of fifth stages on L2 and \stEighthLthree{} of eighth stages on L3 (\cref{fig:stages}).
The 27B follows a similar curve, lower at every stage.
Later stages are not harder in isolation; they depend on earlier truths, on records the agent must keep and, in some families, on the agent's own earlier commits.

\paragraph{Most failed campaigns go wrong before the last stage.}
Of Flash's \stReachLtwo{} L2 campaigns that reach the final commit, \stLtwoRight{} are fully right; of the other \stLtwoWrong, \stLtwoEarly{} first go wrong at stage 1 or 2 and \stLtwoFinal{} at the final stage.
Campaigns whose only wrong stage is the last make up \stAbfMin--\stAbfMax{} of those that reach the final commit on each long-horizon rung.
Errors therefore mostly start early and propagate, rather than accumulating only in the final integration step.
Flash's episodes are read under its per-call cap; a cut turn can make any stage's commit wrong.

\paragraph{Failed episodes replay exactly.}
An audit of \faN{} failed Flash episodes in \faCells{} family rungs without a success, from an earlier run of the frontier tier with the same families and verifiers, found no scoring, parser or simulator fault: replays reproduce every recorded reading, action and verdict, and each of the \faCommits{} committed answers re-scores as wrong, so these failures are not artefacts of scoring, parsing or simulation.

\section{Training on the Worlds}
\label{sec:training}

Because the rewards are exact and gated, the suite can also be used for training.
We train \ours{} on the core tier by on-policy expert iteration and evaluate it on the core tier's final instances, and then train the same model by online reinforcement learning on two laboratory families outside the suite (\cref{sec:labrl}).

\subsection{Method}
\label{sec:method}

In each round the policy plays training-split instances several times, keeps its successful episodes and fine-tunes on every decision of those episodes, its own thinking and action \citep{polu2022formal,zelikman2022star,singh2023beyond}; malformed turns inside rewarded episodes are never targets.
Training has two phases.
The \emph{narrow phase} runs three rounds on the twelve trained discovery worlds with the policy's own episodes only, ending an episode on its first reply without a legal action, and yields \nar{}.
The \emph{broad phase} starts from \nar{}, runs four rounds under the standard protocol on the 23 families that are not held out, and yields \fin{}.
It balances the kept episodes across families, samples every level including the entry levels, and adds a capped share of successful episodes from \qbig{} and \flash{}, rendered in the student's chat format; it also trains on a laboratory data-analysis family outside the suite, built on the same statistics engine as the laboratory families of \cref{sec:labrl} but distinct from them and not evaluated here (the reinforcement learning of \cref{sec:labrl} starts from \ours{}, not from \fin{}).
Episode counts and optimisation settings are in \cref{app:training}; the exact composition of the training mixture is withheld.
Training comparisons use the final instances whose hidden system does not appear among the training draws.
Four criteria for the broad phase were fixed before its evaluation, all relative to \nar{} except the last: fewer invalid endings; higher success on the trained discovery worlds, with a paired lower bound above zero; the same on the six families where \nar{} succeeded on fewer than 0.10 of episodes; and GPQA Diamond \citep{rein2023gpqa} within eight questions of \ours{}, with median length at least 70\% of \ours{}'s.
All four pass; the comparisons with \ours{} were not pre-specified.

\subsection{Results}
\label{sec:trainresults}

\begin{table}[tbp]
\centering
\caption{\capt{Training results on the core tier's final instances.} Differences are paired by episode with 95\% bootstrap intervals (exploratory per family); discovery-world rows sorted by \fin{} $-$ \nar{}. ChemistryWorld (360 episodes) and the four trained PhysicsWorld families (384) are pooled; exact execution and the certificate audit (144 each) and fluid network (96) are single families. Heat inclusion, knockout network and rule induction score 0.000 for every model and are omitted. On the trained families, \fin{} $-$ \nar{} is the pre-specified comparison. $^*$ \nar{} below 0.10. $^\dagger$ held out of all training. $^\ddagger$ includes the two evaluation-only levels.}
\label{tab:results}
\footnotesize
\setlength{\tabcolsep}{3.4pt}
\begin{tabular}{@{}lccccc@{}}
\toprule
\textbf{\sffamily Family} & \textbf{\sffamily \ours{}} & \textbf{\sffamily \nar{}} & \textbf{\sffamily \fin{}} & \textbf{\sffamily \fin{} $-$ \nar{}} & \textbf{\sffamily \fin{} $-$ \ours{}} \\
\midrule
group testing & 0.326 & 0.565 & 0.804 & \dci{+0.239}{+0.065}{+0.413} & \dci{+0.478}{+0.283}{+0.674} \\
circuit fault & 0.097 & 0.111 & 0.208 & \dci{+0.097}{-0.014}{+0.208} & \dci{+0.111}{+0.000}{+0.222} \\
control plant$^*$ & 0.083 & 0.000 & 0.097 & \dci{+0.097}{+0.042}{+0.167} & \dci{+0.014}{-0.069}{+0.097} \\
hidden law & 0.181 & 0.111 & 0.194 & \dci{+0.083}{-0.042}{+0.208} & \dci{+0.014}{-0.097}{+0.125} \\
trace debugging & 0.167 & 0.194 & 0.250 & \dci{+0.056}{-0.042}{+0.153} & \dci{+0.083}{-0.028}{+0.194} \\
pathway fault$^*$ & 0.097 & 0.083 & 0.125 & \dci{+0.042}{-0.056}{+0.125} & \dci{+0.028}{-0.069}{+0.125} \\
impedance topology$^*$ & 0.000 & 0.000 & 0.014 & \dci{+0.014}{+0.000}{+0.042} & \dci{+0.014}{+0.000}{+0.042} \\
reaction network & 0.097 & 0.181 & 0.181 & \dci{+0.000}{-0.111}{+0.111} & \dci{+0.083}{-0.014}{+0.181} \\
causal intervention & 0.111 & 0.153 & 0.139 & \dci{-0.014}{-0.111}{+0.097} & \dci{+0.028}{-0.083}{+0.139} \\
\textbf{twelve trained families} & \textbf{0.090} & \textbf{0.103} & \textbf{0.148} & \dci{\boldsymbol{+0.045}}{+0.019}{+0.072} & \dci{\boldsymbol{+0.059}}{+0.031}{+0.085} \\
\midrule
mechanism discrimination$^\dagger$ & 0.153 & 0.083 & 0.250 & \dci{+0.167}{+0.056}{+0.278} & \dci{+0.097}{-0.042}{+0.236} \\
counterfeit coin$^\dagger$ & 0.097 & 0.111 & 0.250 & \dci{+0.139}{+0.028}{+0.264} & \dci{+0.153}{+0.042}{+0.278} \\
automaton learning$^\dagger$ & 0.000 & 0.000 & 0.000 & 0.000 & 0.000 \\
\textbf{three held-out families} & \textbf{0.083} & \textbf{0.065} & \textbf{0.167} & \dci{\boldsymbol{+0.102}}{+0.051}{+0.157} & \dci{\boldsymbol{+0.083}}{+0.023}{+0.144} \\
\midrule
ChemistryWorld & 0.031 & 0.025 & 0.056 & \dci{+0.031}{+0.008}{+0.056} & \dci{+0.025}{+0.000}{+0.050} \\
PhysicsWorld, trained & 0.010 & 0.005 & 0.008 & \dci{+0.003}{-0.005}{+0.010} & \dci{-0.003}{-0.016}{+0.010} \\
exact execution$^\ddagger$ & 0.618 & 0.562 & 0.611 & \dci{+0.049}{-0.014}{+0.111} & \dci{-0.007}{-0.056}{+0.035} \\
certificate audit & 0.049 & 0.132 & 0.188 & \dci{+0.056}{-0.021}{+0.132} & \dci{+0.139}{+0.069}{+0.208} \\
fluid network$^\dagger$ & 0.010 & 0.010 & 0.000 & \dci{-0.010}{-0.031}{+0.000} & \dci{-0.010}{-0.031}{+0.000} \\
\midrule
\multicolumn{6}{@{}l}{\emph{on the 1,053 compared instances}} \\
invalid endings & 0.095 & 0.049 & 0.013 & & \\
experiments per episode & 2.69 & 2.78 & 3.06 & & \\
\bottomrule
\end{tabular}
\end{table}

\begin{figure}[tbp]
\centering
\includegraphics[width=\linewidth]{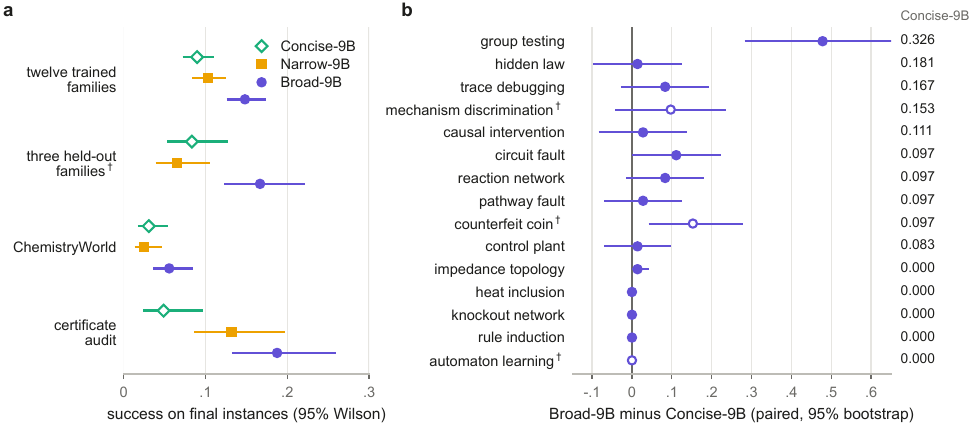}
\caption{\capt{Training results, from \cref{tab:results}.} (a) Pooled success on the final instances, with 95\% Wilson intervals (unpaired; the paired differences are in \cref{tab:results}). (b) Per-family paired difference \fin{} minus \ours{} with 95\% bootstrap intervals for the fifteen discovery worlds, sorted by \ours{}'s success (right column). Open markers and $^\dagger$ mark held-out families.}
\label{fig:training}
\end{figure}

\paragraph{Expert iteration gives modest gains.}
On the twelve trained families (\trN{} final instances), \fin{} succeeds on \trBroad{}, against \trNarrow{} for \nar{}, \trBN{} (pre-specified), and \trConcise{} for \ours{}, \trBC{} (\cref{tab:results,fig:training}).
On the three held-out families, it succeeds on \hlBroad{}, against \hlConcise{} for \ours{}, \hlBC{}, and \hlNarrow{} for \nar{}; the gain comes from mechanism discrimination (0.083 to 0.250 over \nar{}) and counterfeit coin (0.111 to 0.250), while automaton learning stays at zero, so a bootstrap over the three families does not resolve it (\cref{app:cluster}).
In absolute terms the gains are small: after training, the model still fails on more than four in five core-tier episodes of both the trained and the held-out families.
Outside the discovery worlds, ChemistryWorld rose from 0.031 to 0.056 (not resolved) and the certificate audit from 0.049 to 0.188 (\dci{+0.139}{+0.069}{+0.208} over \ours{}); PhysicsWorld and exact execution did not move.

\paragraph{Part of the gain is action format.}
Over the \invN{} final instances of the comparison, invalid endings fell from \invConcise{} for \ours{} and \invNarrow{} for \nar{} to \invBroad{} for \fin{}, \invBN{} against \nar{}.
The largest family gain, group testing (\gtConcise{} to \gtBroad{} on its \gtN{} final instances; \gtBN{} over \nar{} and \gtBC{} over \ours{}), is mostly format: \ours{} ends \gtInvConcise{} of those episodes invalid.
On the other eleven trained families, \fin{} is still \elevenBC{} above \ours{}, and every result is read with its invalid-ending rate.

\paragraph{Gains appear where the model already succeeds.}
All \nPos{} discovery worlds where \ours{} succeeds at all have positive point estimates, from \posMin{} to \posMax{} (\cref{fig:training}b).
The \nNever{} where it never succeeds gain at most \neverMaxGain{} episode, as success filtering predicts: such a family supplies none of the policy's own episodes at its core levels.
Headroom does not explain where the gains fall: pathway fault, with a headroom of 0.82 against the 27B, gains 0.028, and heat inclusion, with 0.92, gains nothing.
In the last round, rule induction succeeded on three quarters of its entry-level episodes and on none at its core levels, so the hardest families still need a first success at the levels that are evaluated.

\paragraph{On-policy expert iteration at 27B (low-rank adapter, own episodes, two rounds).}
A lighter variant of the recipe, applied to \qbig{}, trains a low-rank adapter on the model's own successful episodes only, without a narrow phase or a teacher share, and was read after two rounds.
On the three held-out discovery worlds, success rises from \tsHeldBase{} for the base model to \tsHeldRtwo{} (\tsHeldN{} paired final instances), \tsHeld{} at the episode level; the gain comes from counterfeit coin (\tsHeldCcBase{} to \tsHeldCcRtwo{} of \tsHeldFamN{}) and mechanism discrimination (\tsHeldMpBase{} to \tsHeldMpRtwo{}), while automaton learning stays at \tsHeldAlBase{}, so a bootstrap over the three families does not resolve it, \tsHeldCluster{}.
On the twelve trained families the difference is not resolved, \tsTrained{} (\tsTrainedN{}), and GPQA Diamond stays at \tsGpqaRtwo{} of \tsGpqaN{}, as for the base model.

\paragraph{No detectable loss of general reasoning.}
On GPQA Diamond, \fin{} answers 156 of 198 questions correctly, against 155 for \ours{} and 157 for \nar{}, with a median generation length 98\% of \ours{}'s (\cref{tab:gpqa}).
No GPQA outcome was used in training.

\paragraph{Episode records support route checks.}
\label{sec:skip}
In causal intervention, success changed little (\ciSuccConcise{} for \ours{}, \ciSuccBroad{} for \fin{}, of \ciSuccN{} final instances each), but the route to the joint-effect answer did: the share of joint-effect episodes answered from observational estimates, without an intervention, rose from \routeConcise{} (\routeConciseK{} of \routeConciseN) to \routeBroad{} (\routeBroadK{} of \routeBroadN).
Such estimates land inside the published tolerance on \routeDevMin--\routeDevMax{} of development instances (about \routeDevN{} per level), well below the shortcut gate but close to the policy's own success on that question.
Because every episode records its experiments, a training filter can also require the family's target experiment, which we recommend where a policy's success is low.

\subsection{Reinforcement learning on laboratory worlds}
\label{sec:labrl}

Expert iteration learns only from successful episodes, and on the suite its gains are small.
We therefore also train \ours{} by online reinforcement learning, which learns from failed attempts as well as successful ones, on two laboratory families outside the suite, in which the agent analyses an experiment's data with a statistics engine and commits decisions that a verifier checks.

\paragraph{Laboratory worlds.}
A \emph{replication} world gives the agent a scientist's brief: a claimed treatment effect and the smallest difference that matters; the lab's own experiment as a data table, with its design and a quality-control log; a budget for new experiments, priced per experimental unit and per measurement; and the lab's decision standard.
The design states how animals, wells or donors were assigned to conditions, whether whole containers such as cages or plates were assigned, and how repeated measurements nest within units.
The standard maps the 95\% interval of the treatment effect, estimated at the experimental unit \citep{hurlbert1984pseudoreplication,lazic2010problem} after exactly the exclusions that the log documents, to one of four decisions on the claim (retain, revise, reject or insufficient evidence), and it requires the final decision to rest on an independent replication; when the design does not determine the unit or an exclusion, the required decision is insufficient evidence, with a statement of what cannot be determined.
Within seven tool calls the agent analyses the table, commits a decision on the claim, sizes and runs a replication within the budget and analyses its data, and then replies with a final decision.
Its tools are those of a laboratory data-analysis assistant: table inspection, validated statistical analyses such as t tests at a chosen unit and mixed models, design checks, power calculations, a calculator, the replication experiment and the decision commit; a malformed or refused call returns its reason.
A deterministic verifier recomputes every analysis that a decision cites and accepts the episode only if each decision follows the lab's rule from an analysis at the right unit with the documented exclusions, the replication fits the budget, and every number in the answer comes from a tool result or from the brief.
Worlds span \lrRepStrata{} design strata, from one measurement per unit and repeated measurements through treatments assigned to containers and nested labels that restart in every container to quality-control judgements, replication sizing and designs that leave the unit undetermined.
A \emph{method-comparison} world, built from the same laboratory engine as the coupled campaigns of \cref{sec:coupled}, is a two-stage campaign on a hidden analyser: within a sample budget the agent chooses how many patient samples to measure with the new and a reference method and over what range, locks the analysis set with any exclusions the data justify, and commits whether the two methods are interchangeable at two medical decision points, with the bias at each, under the laboratory's Deming-regression procedure \citep{linnet1993evaluation}.
It has eight tool calls, and its verifier checks the commit against the hidden analyser and accepts it only if the locked data cover and settle both decision points.
Both families state their decision rules and the checks on an answer in the brief.
They are training environments outside the benchmark suite, built in the same way, with a hidden truth, a stated rule and a deterministic verifier, but not part of its calibrated tiers.

\begin{table}[tbp]
\centering
\caption{\capt{Reinforcement learning on laboratory worlds: success on the held-out panel.} Success per world, the mean over \lrG{} paired-seed episodes, averaged over the \lrRepWorlds{} replication and the \lrMcWorlds{} method-comparison worlds (\lrRepEpisodes{} and \lrMcEpisodes{} episodes per system), none of which occurs in training. Trained on: distinct training worlds up to that checkpoint (MC: method comparison). \flash{} plays the same worlds through the same tools, prompts, verifier and per-turn caps. Differences are paired by world, with 95\% intervals from 5,000 bootstrap resamples of worlds. The best system in each column is in bold. One training seed per run; the checkpoints after 16 updates are the registered ones.}
\label{tab:labrl}
\footnotesize
\setlength{\tabcolsep}{5pt}
\begin{tabular}{@{}llcc@{}}
\toprule
\textbf{\sffamily System} & \textbf{\sffamily Trained on} & \textbf{\sffamily Replication} & \textbf{\sffamily Method comparison} \\
\midrule
\grouprow{4}{Starting model}
\warm{} & -- & 0.221 & 0.141 \\
\grouprow{4}{Reinforcement learning, one training seed}
\labrl{}, after 8 updates & 192 lab & 0.579 & 0.172 \\
\labrl{}, after 12 updates & 288 lab & 0.646 & 0.141 \\
\labrl{}, after 16 updates & 384 lab & \textbf{0.721} & 0.125 \\
\mixrl{}, after 8 updates & 128 lab + 64 MC & 0.508 & 0.219 \\
\mixrl{}, after 12 updates & 192 lab + 96 MC & 0.646 & \textbf{0.359} \\
\mixrl{}, after 16 updates & 256 lab + 128 MC & 0.583 & 0.328 \\
\grouprow{4}{API model, same worlds, tools and caps}
\flash{}, low effort & -- & 0.446 & \textbf{0.359} \\
\flash{}, default effort & -- & 0.371 & 0.094 \\
\midrule
\grouprow{4}{Paired differences, 95\% bootstrap over worlds}
\labrl{} $-$ \warm{} & & \dci{+0.500}{+0.400}{+0.596} & \dci{-0.016}{-0.078}{+0.062} \\
\mixrl{} $-$ \warm{} & & \dci{+0.362}{+0.283}{+0.446} & \dci{+0.188}{+0.078}{+0.281} \\
\labrl{} $-$ \flash{}, low & & \dci{+0.275}{+0.150}{+0.404} & \dci{-0.234}{-0.375}{-0.062} \\
\mixrl{} $-$ \flash{}, low & & \dci{+0.138}{+0.017}{+0.254} & \dci{-0.031}{-0.281}{+0.203} \\
\mixrl{} $-$ \labrl{} & & \dci{-0.138}{-0.221}{-0.058} & \dci{+0.203}{+0.078}{+0.328} \\
\bottomrule
\end{tabular}
\end{table}

\paragraph{Method.}
Each update draws \lrWorldsPerUpdate{} worlds that no earlier update used, plays each \lrG{} times with the current policy and takes one optimiser step on all \lrEpisodesPerUpdate{} episodes.
The reward is binary and terminal, 1 when the verifier accepts the episode and 0 otherwise, without length shaping.
An episode's advantage is its reward minus the mean reward of its group of \lrG{} attempts at the same world \citep{shao2024deepseekmath}, without division by the group's standard deviation \citep{liu2025understanding}.
The loss clips the importance ratio \citep{schulman2017ppo}, bounds negative-advantage tokens with a dual clip \citep{ye2020mastering} and adds a per-token penalty on the divergence from the starting model; it is divided by a fixed token count, so that an episode's length does not change the weight of its tokens, and only the model's own tokens are trained, not observations or tool results.
All parameters are trained at a learning rate of $\lrLR$ for \lrUpdates{} updates, with sampling at temperature 1.0 without truncation and at most \lrTurnCap{} tokens per turn and \lrEpisodeCap{} per episode.
The starting model, \warm{}, is \ours{} after a short supervised warm-up on turns from its own episodes in training replication worlds.
Two runs start from \warm{} with the same seed.
\labrl{} trains on \lrWorldsPerUpdate{} replication worlds per update (\lrTrainRepWorldsLab{} in all).
\mixrl{} trains on \lrMcLabPerUpdate{} of those replication worlds per update, in the same order and with the same environment seeds, plus \lrMcMcPerUpdate{} method-comparison worlds (\lrTrainRepWorldsMix{} and \lrMcWorldsMix{} in all), so that it replaces a third of \labrl{}'s replication practice with method comparison.
The runs took \lrHoursLab{} and \lrHoursMix{} hours on one B200 GPU; \cref{app:labrl} gives every setting.

\paragraph{Registration and evaluation.}
The design, its checkpoints and a rule for reading the results were registered before the runs began, and the registration treats this round of training as exploratory.
Its primary comparison, \mixrl{} against \labrl{} on a locked test set of complete laboratory workflows built from public replication data, has not been read and is not reported here.
We report the registered secondary evaluations: a held-out panel of the two training families, a benchmark of laboratory data analysis, a development set of complete workflows and GPQA Diamond.
The panel holds \lrRepWorlds{} replication worlds, \lrRepPerStratum{} from each of the \lrRepStrata{} design strata, and \lrMcWorlds{} method-comparison worlds, \lrMcPerLevel{} at each of \lrMcLevels{} levels, drawn from the families' development splits.
It is disjoint from training: no panel world, scenario, parent study or task content occurs among the \lrPoolWorlds{} training-split worlds that the runs and the warm-up drew from, which include every world either run trained on (\cref{app:labrl}).
Every system plays each panel world \lrG{} times with the same environment and sampling seeds; success per world is the mean over its episodes, averaged over worlds, and differences are paired by world.
The checkpoints after \lrUpdates{} updates are the registered ones; those after 8 and 12 updates were evaluated and not selected on.
For reference, \flash{} plays the panel through the same tools, prompts, verifier and per-turn caps, counted in the local models' tokens, at low reasoning effort, the setting fixed before its run as its reference reading, and at its default effort (``high''), at which more of its turns reach the cap (\lrFlashHighCut{} capped turns on the panel, against \lrFlashLowCut{} at low effort).
These comparisons with Flash are not part of the registration.

\begin{figure}[tbp]
\centering
\includegraphics[width=\linewidth]{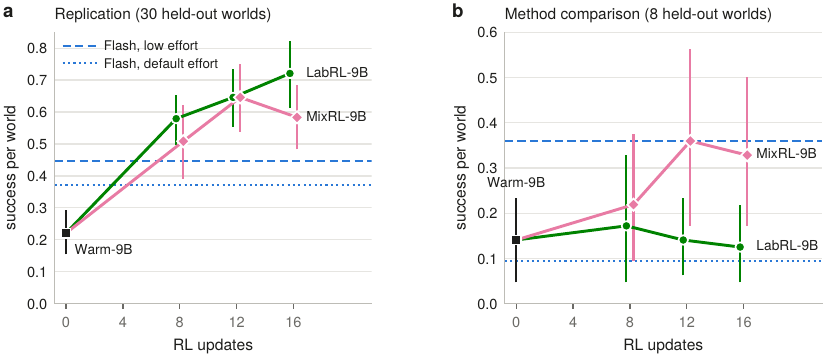}
\caption{\capt{Held-out panel success over training, from \cref{tab:labrl}.} (a) Replication and (b) method comparison: success per world for \labrl{} and \mixrl{} after 8, 12 and 16 updates, both starting from \warm{} at 0 updates, with 95\% bootstrap intervals over worlds; markers are offset slightly so that intervals do not overlap. The horizontal lines mark \flash{} at low and at default reasoning effort on the same worlds.}
\label{fig:labrl}
\end{figure}

\paragraph{Reinforcement learning raises held-out replication success above Flash under the 9B's caps.}
On the \lrRepWorlds{} held-out replication worlds, \labrl{} succeeds on \lrRepLab{}, against \lrRepStart{} for \warm{}, \lrRepLabMinusStart{}, and its point estimate rises at each evaluated checkpoint: \lrRepLabEight{}, \lrRepLabTwelve{} and \lrRepLab{} after 8, 12 and 16 updates (\cref{tab:labrl}; \cref{fig:labrl}a).
It is above \flash{} at low effort (\lrRepFlashLow{}), \lrRepLabMinusFlashLow{}, and at its default effort (\lrRepFlashHigh{}), \lrRepLabMinusFlashHigh{}.
Both Flash readings hold it to the 9B's caps of \lrTurnCap{} tokens per turn and \lrEpisodeCap{} per episode, counted in the 9B's tokenizer; at its default effort Flash reaches the turn cap more often (\lrFlashHighCut{} against \lrFlashLowCut{} capped turns) and scores lower than at low effort, so these readings understate Flash, which leads at a larger budget on the broader evaluations below.
In training, the mean reward per update rose from \lrTrainRewardFirst{} at the first update to \lrTrainRewardLast{} at the last.

\paragraph{Each run improves the family it trains on.}
\labrl{} does not move on method comparison (\lrMcStart{} to \lrMcLab{}, \lrMcLabMinusStart{}).
\mixrl{} rises there to \lrMcMixTwelve{} after 12 updates and \lrMcMix{} after 16, \lrMcMixMinusStart{} over \warm{}, and its difference from \flash{} at low effort (\lrMcFlashLow{}) is not resolved, \lrMcMixMinusFlashLow{}; on replication it reaches \lrRepMix{}, \lrRepMixMinusStart{} over \warm{} and \lrRepMixMinusFlashLow{} over Flash at low effort.
After 12 updates, \mixrl{} matches \labrl{} on replication (\lrRepMixTwelve{} each; \lrRepMixTwelveMinusLabTwelve{}) with \lrMixTrainRepWorldsTwelve{} replication worlds against \lrTrainRepWorldsTwelve{}, and leads on method comparison, \lrMcMixTwelveMinusLabTwelve{}.
At the registered checkpoint after 16 updates, it is lower on replication, \lrRepMixMinusLab{}, and higher on method comparison, \lrMcMixMinusLab{} (\cref{fig:labrl}); with one training seed per run, these checkpoints do not separate a cost of mixing the families from variation between checkpoints.
The panel measures learning in the training families on held-out worlds; the broader evaluations below and the locked test address other laboratory work.

\begin{table}[tbp]
\centering
\caption{\capt{Broader evaluations of the laboratory runs.} P+X: a first-party benchmark of laboratory data analysis through the same statistics engine, on its \lrPXItems{} development items (P: \lrPItems{} multi-step projects; X: \lrXItems{} experiment-design decisions), one attempt each, scored as the macro mean of the two suites. Answered: share of P+X items answered in the required format within the token limit. R: the benchmark's \lrRItems{} real-data items solved. Workflow: the mean step score on \lrWFItems{} development workflows built from public replication data, one attempt each. GPQA: correct answers of \lrGpqaN{} under the protocol of \cref{app:gpqa}. \flash{} at its default reasoning effort; -- not evaluated. P+X minus \warm{} (paired by item, 95\% bootstrap intervals): \labrl{} \lrPXLabEightMinusStart{}, \lrPXLabTwelveMinusStart{} and \lrPXLabMinusStart{} after 8, 12 and 16 updates; \mixrl{} \lrPXMixEightMinusStart{}, \lrPXMixTwelveMinusStart{} and \lrPXMixMinusStart{}; \mixrl{} $-$ \labrl{} after 16 updates: \lrPXMixMinusLab{}.}
\label{tab:labrlbroad}
\footnotesize
\setlength{\tabcolsep}{6pt}
\begin{tabular}{@{}lccccccc@{}}
\toprule
\textbf{\sffamily System} & \textbf{\sffamily P+X} & \textbf{\sffamily P} & \textbf{\sffamily X} & \textbf{\sffamily Answered} & \textbf{\sffamily R} & \textbf{\sffamily Workflow} & \textbf{\sffamily GPQA} \\
\midrule
\warm{} & 0.501 & 0.403 & 0.599 & 0.813 & 6 & 0.447 & 159 \\
\labrl{}, after 8 updates & 0.626 & 0.481 & 0.772 & 0.948 & 6 & 0.474 & -- \\
\labrl{}, after 12 updates & 0.628 & 0.492 & 0.765 & 0.964 & 6 & 0.586 & -- \\
\labrl{}, after 16 updates & 0.605 & 0.478 & 0.732 & 0.976 & 4 & 0.454 & 161 \\
\mixrl{}, after 8 updates & 0.595 & 0.450 & 0.739 & 0.916 & 6 & 0.480 & -- \\
\mixrl{}, after 12 updates & 0.596 & 0.419 & 0.772 & 0.918 & 6 & 0.658 & -- \\
\mixrl{}, after 16 updates & 0.609 & 0.461 & 0.757 & 0.949 & 5 & 0.553 & -- \\
\flash{}, default effort & 0.849 & 0.794 & 0.904 & 0.998 & 6 & 0.895 & -- \\
\bottomrule
\end{tabular}
\end{table}

\paragraph{On broader laboratory analysis, the gain comes from more reliable finishing, and Flash stays ahead.}
On a first-party benchmark of laboratory data analysis answered through the same statistics engine (P+X; \cref{tab:labrlbroad}), \labrl{} raises the score from \lrPXStart{} to \lrPXLab{}, \lrPXLabMinusStart{}, with gains in both suites (projects \lrPStart{} to \lrPLab{}, design decisions \lrXStart{} to \lrXLab{}); after 12 updates its gain is \lrPXLabTwelveMinusStart{}, and \mixrl{} gains \lrPXMixMinusStart{}.
The gain comes from finishing: the share of items answered in the required format within the token limit rises from \lrAnsStart{} to \lrAnsLab{}, and on the \lrMatchedN{} items that both models answered, accuracy is \lrMatchedLab{} against \lrMatchedStart{}.
The registered checkpoint of \labrl{}, after 16 updates, fails one of the benchmark's registered guards against the starting model: it solves \lrRLab{} of the \lrRItems{} real-data items (R) against \lrRStart{} for \warm{}, and the guard allows at most one fewer; its other guards hold. Every registered guard holds for \labrl{} after 12 updates and for \mixrl{} after 16.
Without the power-calculation items, which the registration reports separately, the differences from \warm{} are \lrPXnpLabMinusStart{} for \labrl{}, \lrPXnpLabTwelveMinusStart{} after 12 updates and \lrPXnpMixMinusStart{} for \mixrl{} (\cref{app:labrl}).
With up to \lrFlashMaxTokens{} tokens per call at its default effort, Flash stays well ahead on P+X, at \lrPXFlash{} against \lrPXLab{} for \labrl{}, \lrPXFlashMinusLab{}.
On the development set of a complete-workflow evaluation, \lrWFItems{} workflows built from the public data of three experiments in one study of the Reproducibility Project: Cancer Biology \citep{errington2021investigating}, each asking in turn for the experimental unit, in most workflows an exclusion decision, whether the claim holds and the sample size of the next experiment, \labrl{} scores \lrWFLab{} against \lrWFStart{} for \warm{} (\lrWFLabMinusStart{}), and the checkpoints after 12 updates score higher: \lrWFLabTwelve{} for \labrl{} (\lrWFLabTwelveMinusStart{}) and \lrWFMixTwelve{} for \mixrl{} (\lrWFMixTwelveMinusStart{}), against \lrWFMix{} for \mixrl{} after 16 (\lrWFMixMinusStart{}); Flash scores \lrWFFlash{}.
These intervals resample workflows, which come from three source experiments, so they understate the uncertainty of this small set.
Both sets are development splits that informed earlier design choices, and none of the 9B models was trained on their items.

\paragraph{General reasoning.}
On GPQA Diamond under the protocol of \cref{app:gpqa}, scored as in \cref{tab:gpqa}, \warm{} answers \lrGpqaStart{} of \lrGpqaN{} questions and \labrl{} \lrGpqaLab{}, with a median generation of \lrGpqaMedLab{} tokens, \lrGpqaMedRatio{} of \warm{}'s \lrGpqaMedStart{}. The two are equal within the run-to-run variation of this measurement: two runs of \ours{} under the same protocol answered \lrGpqaConcise{} (\cref{tab:gpqa}) and \lrGpqaConciseRepeat{}.

\section{Limitations}
\label{sec:limitations}

The families idealise laboratory and reasoning problems, and calibration cannot exclude shortcuts that nobody has written, nor rule out that a route below the gate still shapes training (\cref{sec:skip}).
The classic-puzzle families measure experimental reasoning in a pure form, not domain science; the suite's simulator-grounded science is in PhysicsWorld and ChemistryWorld, whose simulators have the known limits recorded in their datasheets.
The frontier tier was tuned against the one API model it reports: its target band was a range of Flash success, read from screens of Flash on final instances, and its action parsers, part of its campaign wording, one budget display and the re-ask after a cut turn were revised after Flash's first screen of those instances, which were then played again (\cref{sec:families}). Selecting rungs against Flash biases Flash's absolute rates on that tier downwards and favours other models in comparisons with Flash; confirming Flash's frontier rates on an untouched block of final instances is future work.
Flash's tier-wide results use the provider's default reasoning effort and 16,384 tokens per call and understate it: on the \fmaxAllN{} $o = 0$ frontier instances, one per family and rung, its maximum effort with up to 393,216 tokens per call raises its successes from \fmaxAllDef{} to \fmaxAllK{}, below the pilot model's \fmaxAllSol{} and the oracle's \fmaxAllOracle{}, and its maximum-effort rates on the other frontier instances and on the core tier are not measured. The headroom we claim is against the oracle, with the pilot model at xhigh effort as the strongest system tested; it lies mainly in the eight-stage campaigns, where neither that model nor Flash at maximum effort solves more than \solLthreeK{} of \solLthreeN{}, and, in development-instance probes, in the longer campaigns.
The candidate rungs above the frontier tier were probed on development instances with one model at xhigh effort and eight or sixteen instances per rung, and two of them also on 32 final instances; four rungs at 3--5 of 8 were not extended to sixteen as the probe protocol prescribes, and of the partly-holding rungs only one was rebuilt harder, so the probes show a direction for the ladder rather than results on the evaluated suite.
Flash's PhysicsWorld run covers the first \fpPhysRounds{} of 24 final instances per level, paired with the other systems on those instances.
The pilot plays one final instance per family and rung through the Codex CLI rather than the API protocol, without our per-call token cap, so its rates are pooled estimates with wide intervals that support no per-family claim.
The tier's main results rest on one API model at up to 24 instances per family and rung (\roundsHthree{} on H3, \roundsLthree{} on L3), one locally run model at 8, whose thinking is near its 4,096-token cap on most decisions (a mean of \qThinkMin--\qThinkMax{} tokens per decision), and a one-instance pilot of a third model; a broader model comparison is a natural next step.
Scores depend on the evaluation budgets: 4,096 thinking tokens per decision for the local models and 16,384 tokens per call for the API model, which reaches that limit on \capCutPct\% of its frontier decisions; episodes that end on a re-asked turn succeed far less often, and its maximum effort with a larger budget raises Flash's success on one instance per family and rung from \flashPilotFrontierRate{} to \fmaxAllRate{} (\cref{sec:flashmax}), so much of the frontier tier's difficulty for Flash is a budget effect.

The training study is one model and one run.
Its intervals cover the sampling of evaluation episodes, not variation across training runs, and its two phases differ in family mix, teacher share, protocol and number of rounds, so their separate contributions are not identified.
GPQA Diamond with one response per question detects a large loss of general reasoning but not a small one, and transfer rests on three held-out discovery worlds, two of which improve in point estimate, and one held-out PhysicsWorld family.

The reinforcement-learning study (\cref{sec:labrl}) trains one seed per run, and its intervals describe the panel's worlds and episodes, not variation across training runs.
Its panel holds \lrRepWorlds{} and \lrMcWorlds{} held-out worlds of the two training families, so it measures learning within those families; the registered test on locked real-data workflows has not been read, and the broader evaluations use development splits with one attempt per item, on which the gain carries over through more reliable finishing, with accuracy on the items both models answered unchanged, while Flash stays ahead.
The laboratory families are training environments outside the benchmark suite, built in the same way but not part of its calibrated tiers, and we make no claim that the trained models learned to identify the experimental unit from the stated design; the locked workflow test addresses that.
Flash's reference setting on the panel is its low reasoning effort; at its default effort its success is lower on both families (\lrRepFlashHigh{} and \lrMcFlashHigh{}), and the comparison with it is not part of the registration. Both panel readings hold Flash to the 9B's per-turn and per-episode token caps, which bind it more at its default effort, so the panel understates it; with up to \lrFlashMaxTokens{} tokens per call it leads on the broader evaluations. The registered checkpoint of \labrl{} fails one registered guard of the broader benchmark (R).

\section{Conclusion}
\label{sec:conclusion}

We presented Openhagen Worlds, a suite of calibrated, procedurally generated discovery environments (with finite truth spaces in some families) in which hidden truths are checked exactly, every scoring rule is published and a gate at every level holds the naive policy to at most 0.25 and every gated shortcut to at most 0.35 of development instances while the oracle succeeds on at least 0.75.
Its core tier separates the local models from the oracle and approaches saturation for frontier models; on both tiers, Flash's tier-wide results are measured under a capped per-call budget and understate it: at its maximum effort it solves \fmaxAllK{} rather than \fmaxAllDef{} of \fmaxAllN{} frontier instances, one per family and rung, still below the oracle's \fmaxAllOracle{}.
In a pilot, a stronger model at xhigh reasoning effort solves most of the frontier tier, and Flash at its maximum effort closes about two thirds of the gap to it (\fmaxAllK{} against \fmaxAllSol{} of \fmaxAllN{}; \fmaxLthreeK{} against \fmaxLthreeSol{} of \fmaxN{} on the eight-stage campaigns), leaving headroom mainly in those campaigns and in the gap to the oracle, so the ladder has to keep climbing; development-instance probes show that most longer campaigns admitted by the same gate, which make memory and planning bind, hold that model where added hidden structure does not.
Under Flash's per-call cap, failed episodes mostly end on wrong answers rather than malformed output, and campaign errors start early and compound.
On-policy expert iteration on the suite gives a 9B model modest gains, resolved on the trained families and carried by two of three held-out families, without a detectable loss on GPQA Diamond, and its episode records let a training filter check how each answer was reached.
In a separate study, online reinforcement learning on laboratory worlds outside the suite raises a 9B model's success on held-out replication worlds from \lrRepStart{} to \lrRepLab{} in one training run, above Flash held to the 9B's token caps but not above Flash at a larger budget on broader laboratory analysis (\lrPXFlash{} against \lrPXLab{}), and its registered checkpoint fails one registered guard.
Beyond single worlds, we have composed the families into multi-laboratory campaigns, in which samples and measurements pass between laboratories, runs fail or drift, and requests change partway through; training runs that included these campaigns were accompanied by gains on held-out laboratory tasks, and we will report a controlled comparison separately.
We hope the suite serves as a calibrated measure of experimental reasoning; we plan to extend the frontier tier's evaluation to more models and to the full set of withheld instances at maximum-effort budgets, to admit the longer campaigns as rungs of the evaluated suite, and to read the locked workflow test of the laboratory runs and repeat them over training seeds.

\section*{Availability}
Openhagen Worlds, including the datasheet of every family and the frontier tier, is available to research and industry partners under licence; the multi-laboratory campaigns are available on the same terms.
A hosted evaluation service (\url{https://eval.openhagen.ai}; its front page and \url{https://eval.openhagen.ai/evaluation} document the API, and access tokens are issued on request to \href{mailto:hello@openhagen.ai}{\texttt{hello@openhagen.ai}}) plays exactly the core tier's 2,208 final instances with the same family code and protocol, and reveals only the observation, reward, termination and success, so any model can be scored on the instances behind \cref{tab:results,tab:baselines} without access to the generators.
Played through the service, the scripted oracle and naive policy reproduce all 30 of their per-family final-instance rates on the discovery worlds.
A frozen sample of 48 development-split instances from two families, trace debugging and causal intervention, each with its observation, hidden truth and a verified gold answer, and one scripted-oracle episode per family, is released with this report under CC BY 4.0, as ancillary files of this preprint and at \url{https://eval.openhagen.ai/sample/} (\cref{app:samples}).
The keys of the final instances and of the protected-structure filter are held by Openhagen and never distributed.

\section*{Acknowledgements}
The author is not a native English speaker and used AI writing assistance to help write this paper; AI coding assistants were also used to implement and run experiments under the author's direction.
The research, including its questions, design, decisions and conclusions, is the author's own, and the author takes full responsibility for the content.

\bibliographystyle{plainnat}
\bibliography{references}

\clearpage
\appendix
\crefalias{section}{appendix}
\titleformat{\section}{\sffamily\bfseries\large\color{ohink}}{\thesection}{0.75em}{}

\section{Sample episodes from the public sample}
\label{app:samples}

The public sample holds eight development-split instances per core level of two families, each with its observation, hidden truth and a verified gold commit, and one scripted-oracle episode per family.
Its four files, \texttt{README.md}, \texttt{MANIFEST.json} (record counts and SHA-256 digests), \texttt{instances.jsonl} and \texttt{oracle\_episodes.jsonl}, ship as ancillary files of this preprint and at \url{https://eval.openhagen.ai/sample/}, under CC BY 4.0.
We show the first level-0 instance of each family, with its observation abridged only where marked, and the oracle's recorded episode on it.

\paragraph{Causal intervention.}
The agent must find the direct causes of a target in a hidden linear structural causal model with an unobserved common cause, using at most four single-variable interventions.

\begin{tcolorbox}[enhanced,colback=ohpanel,colframe=ohpanelrule,boxrule=0.4pt,arc=4pt,left=6pt,right=6pt,top=4pt,bottom=4pt,
  title={Observation (abridged)},colbacktitle=ohgroup,coltitle=ohink,fonttitle=\sffamily\bfseries\footnotesize,toptitle=2pt,bottomtitle=2pt]
\begin{Verbatim}[fontsize=\scriptsize]
variables:   growth, humidity, pressure, turbidity        target: growth
candidates:  humidity, pressure, turbidity
intervention_range:  humidity [3.5, 13.3]  pressure [6.6, 18.2]  growth [10.6, 33.2]
                     turbidity [16.2, 27.8]
model:       hidden linear structural causal model over the observed variables: each variable =
             intercept + sum of coefficient x (its direct causes) + Gaussian noise; the causal graph
             is a random DAG (names carry no information); an unobserved common cause may drive
             several variables; every direct effect on growth has |coefficient| >= 0.6.
             INTERVENE X=v sets X to v in a fresh batch of 20 samples (X's own causes are cut; the
             other variables respond through the graph)
objective:   find the set of direct causes of growth: the variables that appear with a nonzero
             coefficient in growth's own structural equation (not causes acting only through other
             variables, not effects of growth, not merely correlated variables)
success_rule: the COMMIT succeeds only if the committed set equals the set of direct causes exactly;
             an unknown variable, a value outside the intervention range or an intervention beyond
             the budget ends the episode at zero
grammar:     INTERVENE <variable>=<value>; at most 4 interventions.
             Then: COMMIT causes=<comma-separated variable names> (or COMMIT causes=none).
history:     observational record, n = 200: means growth 21.88, humidity 8.38, pressure 12.39,
             turbidity 21.94, and the 4 x 4 covariance matrix
\end{Verbatim}
\end{tcolorbox}

\begin{tcolorbox}[enhanced,colback=ohpanel,colframe=ohpanelrule,boxrule=0.4pt,arc=4pt,left=6pt,right=6pt,top=4pt,bottom=4pt,
  title={Scripted-oracle episode},colbacktitle=ohgroup,coltitle=ohink,fonttitle=\sffamily\bfseries\footnotesize,toptitle=2pt,bottomtitle=2pt]
\begin{Verbatim}[fontsize=\scriptsize]
1  INTERVENE humidity=13.3    ->  means: growth 28.57, pressure 17.90, turbidity 26.89 (n = 20)
2  INTERVENE pressure=18.2    ->  means: growth 29.41, humidity 8.61, turbidity 27.01
3  INTERVENE turbidity=27.8   ->  means: growth 21.29, humidity 8.23, pressure 12.05
4  COMMIT causes=pressure     ->  correct; reward 1
\end{Verbatim}
\end{tcolorbox}

Setting humidity moves growth, but only through pressure: the hidden truth is humidity $\to$ pressure $\to$ growth and pressure $\to$ turbidity, with the unobserved cause driving growth and humidity, so an observational regression would credit humidity as well.

\paragraph{Trace debugging.}
The agent must find the one buggy line of a short program and its fix, using at most five runs or line traces.

\begin{tcolorbox}[enhanced,colback=ohpanel,colframe=ohpanelrule,boxrule=0.4pt,arc=4pt,left=6pt,right=6pt,top=4pt,bottom=4pt,
  title={Observation (abridged)},colbacktitle=ohgroup,coltitle=ohink,fonttitle=\sffamily\bfseries\footnotesize,toptitle=2pt,bottomtitle=2pt]
\begin{Verbatim}[fontsize=\scriptsize]
program:     def f(c, w):
              1|  bonus = 0
              2|  gap = 0
              3|  for j from 2 to c + 3:
              4|      if j % 3 == 0:
              5|          gap = gap - c - c
              6|  gap = gap + w * bonus
              7|  return gap + bonus * 2
input_domain: f(c, w) with every argument an integer from 0 to 6
examples:    f(0, 1) = 0   f(1, 2) = -2   f(1, 3) = -2   f(2, 5) = -4   f(4, 1) = -16   f(6, 0) = -24
bug:         exactly one line differs from the intended program, by one edit: an operator swapped
             within its kind, an integer constant off by one, or a variable read replaced by another
             variable defined at that line; some examples may already pass
tools:       RUN: the shown program's output. TRACE: the variables defined at line n just before
             each run of line n (first 12 runs)
success_rule: the COMMIT succeeds only if the program with that line's expression replaced by the
             fix returns the intended value for every input of the domain; an input outside the
             domain, a variable not defined at that line or an action beyond the budget ends the
             episode at zero
grammar:     RUN input=(<int>, <int>) or TRACE input=(<int>, <int>) line=<n>; at most 5 in total.
             Then: COMMIT line=<n> fix=<that line's expression>
\end{Verbatim}
\end{tcolorbox}

\begin{tcolorbox}[enhanced,colback=ohpanel,colframe=ohpanelrule,boxrule=0.4pt,arc=4pt,left=6pt,right=6pt,top=4pt,bottom=4pt,
  title={Scripted-oracle episode},colbacktitle=ohgroup,coltitle=ohink,fonttitle=\sffamily\bfseries\footnotesize,toptitle=2pt,bottomtitle=2pt]
\begin{Verbatim}[fontsize=\scriptsize]
1  TRACE input=(6, 0) line=4   ->  line 4 ran 8 times, j = 2, 3, ..., 9; gap 0 -> -12 -> -24
2  COMMIT line=3 fix=2 to c + 2 ->  correct; reward 1
\end{Verbatim}
\end{tcolorbox}

The trace shows the loop running to $j = 9 = c + 3$, one step past the intended bound; with the loop ending at $c + 2$, every example holds.

\section{Suite details}
\label{app:suite}

\begin{table}[H]
\centering
\caption{\capt{The 27 families.} \emph{Recover}: what the agent must deduce. \emph{Experiments}: what it can do before committing. \emph{Backing}: how readings are computed. Each group heading gives its number of core levels per family. $^\dagger$ held out of all training; $^\ddagger$ two levels use an evaluation-only grammar.}
\label{tab:suite}
\scriptsize
\setlength{\tabcolsep}{2.2pt}
\renewcommand{\arraystretch}{1.05}
\begin{tabular}{@{}L{3.0cm}L{5.6cm}L{3.2cm}L{2.5cm}@{}}
\toprule
\textbf{\sffamily Family} & \textbf{\sffamily Recover} & \textbf{\sffamily Experiments} & \textbf{\sffamily Backing} \\
\midrule
\grouprow{4}{Discovery worlds: twelve trained, three held out; three core levels each}
circuit fault & the open, shorted or drifted resistor of a network & voltage and current probes & nodal analysis \\
impedance topology & an R/L/C network's topology and element values & impedance at chosen frequencies & complex impedance \\
heat inclusion & position, size and contrast of an insert in a heated plate & thermocouples, heated edge & finite volumes \\
hidden law & a power law's exact exponents, then an extrapolation & measurements in safe ranges & closed form \\
control plant & a plant with dead time, or a controller meeting a spec & step and sine experiments & exact discretisation \\
reaction network & a kinetic scheme among candidates and a rate constant & batch runs & closed form \\
mechanism discrimination$^\dagger$ & one of four nested mechanisms, then an extrapolation & measurements in a safe range & closed form \\
pathway fault & the faulty step of a curated kinetic model & time courses and boluses & libRoadRunner \\
knockout network & every synthetic-lethal pair of a signalling network & single and double knockouts & Boolean reachability \\
group testing & the contaminated sample, within $\lceil\log_2 N\rceil$ tests & pooled tests & exact \\
counterfeit coin$^\dagger$ & the off-weight coin and its direction, at the bound & balance weighings & exact \\
rule induction & a hidden rule over number triples, by equivalence & membership tests & exact \\
causal intervention & direct causes or a joint effect under confounding & interventions & linear causal model \\
trace debugging & the buggy line of a program and its fix & runs and line traces & interpreter \\
automaton learning$^\dagger$ & a hidden Mealy machine with an error state & membership, equivalence queries & exact \\
\grouprow{4}{MicroWorld: six core levels each}
certificate audit & the revoked root and corrupted feed of a certificate graph & direct verifications & exact \\
exact execution$^\ddagger$ & three exact values of a long generated program & none; one commit & two interpreters \\
\grouprow{4}{PhysicsWorld simulator families: four core levels each}
rigid-body identification & payload, damping, friction, offset and spring of a hinge & releases and holds & rigid-body dynamics \\
airfoil inference & lift slope, zero-lift angle, Reynolds-dependent stall; design & wind-tunnel runs & panel method \\
optical bench & a sealed lens train's focal properties, then a set-up & images at chosen positions & ray transfer \\
coupled oscillators & clamps, broken links, walls and stiffnesses of a cart chain & impact tests & modal model \\
fluid network$^\dagger$ & a burst, crack, draw or blockage, its site and size & readings under head, valves & hydraulic solver \\
\grouprow{4}{ChemistryWorld: three core levels each}
structure elucidation & one constitutional isomer of a molecular formula & spectra under a credit budget & RDKit isomer bank \\
titration equilibria & the acid--base system, concentrations and p$K_a$ values & pH, indicator endpoints & charge balance \\
kinetics process design & an Arrhenius network, then a process meeting a spec & isothermal batch runs & closed form \\
chromatography method & retention of a mixture, then a method separating every pair & gradient runs & retention model \\
electrochemical mechanism & an E, EC, CE or ECE mechanism and its rate constant & cyclic voltammetry & integral-equation CV \\
\bottomrule
\end{tabular}
\end{table}

\begin{table}[H]
\centering
\caption{\capt{Terms used in the paper,} with the section that defines them.}
\label{tab:terms}
\footnotesize
\setlength{\tabcolsep}{4pt}
\begin{tabular}{@{}L{3.6cm}L{9.0cm}l@{}}
\toprule
\textbf{\sffamily Term} & \textbf{\sffamily Meaning} & \textbf{\sffamily Sec.} \\
\midrule
family, level, d0--d5 & One environment type; one of its core difficulty settings, numbered from the easiest. & \ref{sec:families} \\
core tier, frontier tier & The 92 core levels of all 27 families; the 86 harder and long-horizon rungs of the 15 discovery worlds. & \ref{sec:families} \\
rung, H1--H3, L1--L3 & A frontier level: harder single questions (H) or campaigns of 3, 5 and 8+ stages (L). & \ref{sec:families} \\
campaign, stage commit & A long-horizon episode on one hidden system; a committed sub-answer that closes one stage. & \ref{sec:families} \\
oracle, naive, shortcut policy & The gate's scripted policies: a planning, deducing oracle; the strongest plausible non-deductive policy; specific non-reasoning habits, including an answer prior. & \ref{sec:pipeline} \\
exhaustive design & Every allowed experiment in a fixed order, then the oracle's analysis; reported, not gated. & \ref{sec:pipeline} \\
development, final instances & Instances for calibration and design (64 per level and block) / withheld evaluation instances from a private key. & \ref{sec:pipeline} \\
invalid ending & An episode that ends on replies without a legal action. & \ref{sec:episode} \\
completed-episode success & Success among episodes that do not end invalid. & \ref{sec:metrics} \\
resolved & A paired difference whose 95\% interval excludes zero. & \ref{sec:metrics} \\
headroom & Oracle success minus model success on the same final instances. & \ref{sec:metrics} \\
\ours{}, \nar{}, \fin{} & The 9B model before training, after the narrow phase and after the broad phase. & \ref{sec:method} \\
\warm{}, \labrl{}, \mixrl{} & \ours{} after a laboratory warm-up, then after reinforcement learning on replication worlds, or on replication and method-comparison worlds. & \ref{sec:labrl} \\
forced close, re-ask & At the thinking cap, the thinking is closed and an answer requested (local models); at the API's token limit, one thinking-off request for the action that carries the cut reasoning (API model). & \ref{sec:setup} \\
\bottomrule
\end{tabular}
\end{table}

\section{Calibration details}
\label{app:calibration}

\begin{figure}[H]
\centering
\includegraphics[width=\linewidth]{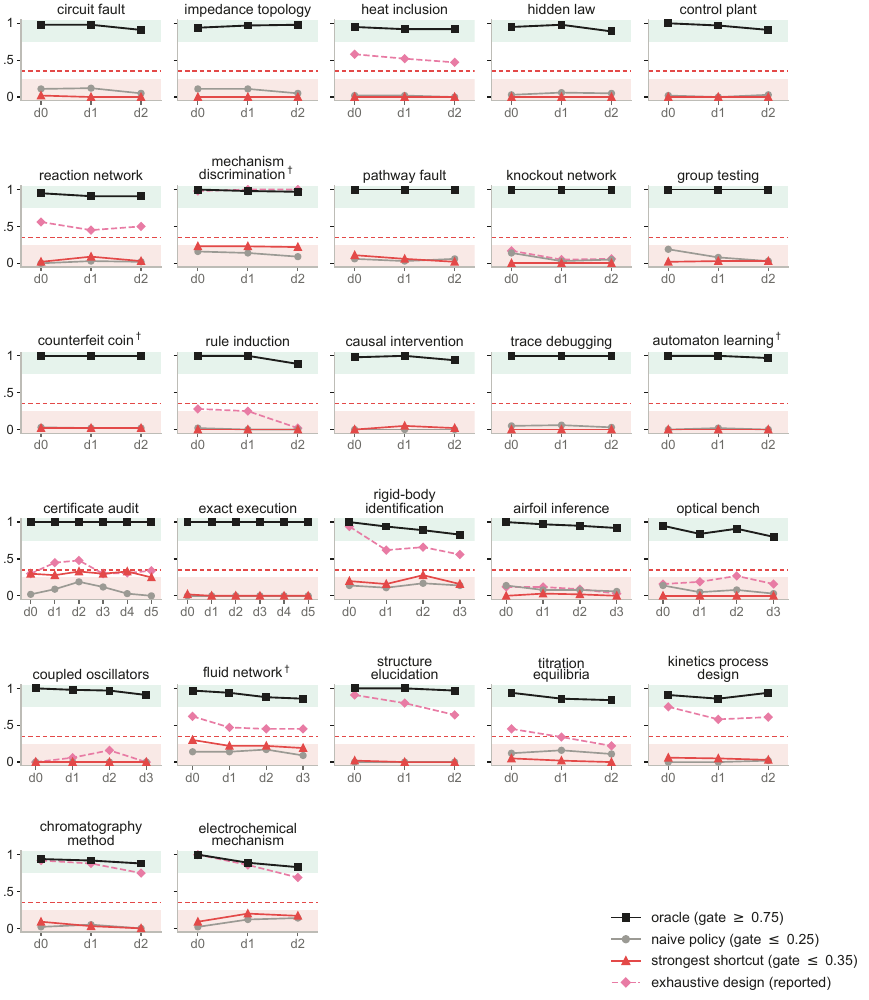}
\caption{\capt{Calibration of all 27 families per core level,} 64 development instances per level (first of two blocks for the PhysicsWorld and ChemistryWorld families). Shaded bands mark the oracle and naive gates; the dashed line is the shortcut gate (0.35). $^\dagger$ held out of all training.}
\label{fig:calibration}
\end{figure}

\Cref{fig:calibration} and \cref{tab:calibration,tab:calibration-micro,tab:calibration-new} give the calibration of every core level, and \cref{tab:frontier-gate} that of every frontier rung.
Two diagnostics outside the gate refine the grouping of \cref{sec:calibration}: a fixed sweep of a control plant with the oracle's fit scores 0.62--0.75, so analysis carries much of that family, and knockout network is solved by a plan fixed in advance from the interaction map, so its skill is deducing that plan rather than reacting to outcomes.
Rigid-body identification's exhaustive design reaches 0.94 at d0 and 0.56--0.66 above it.

\begin{table}[H]
\centering
\caption{\capt{Calibration of the discovery worlds' core levels} on 64 development instances per level (d0--d2). Shortcut: the strongest shortcut policy, including the answer prior. Exhaustive: every allowed experiment in a fixed order, then the oracle's analysis, where the family defines it. $^\dagger$ held out of all training.}
\label{tab:calibration}
\scriptsize
\renewcommand{\arraystretch}{0.95}
\setlength{\tabcolsep}{2.6pt}
\begin{tabular*}{\linewidth}{@{\extracolsep{\fill}}l ccc ccc ccc ccc@{}}
\toprule
 & \multicolumn{3}{c}{\textbf{\sffamily Oracle}} & \multicolumn{3}{c}{\textbf{\sffamily Naive}} & \multicolumn{3}{c}{\textbf{\sffamily Shortcut}} & \multicolumn{3}{c}{\textbf{\sffamily Exhaustive}} \\
\cmidrule(lr){2-4}\cmidrule(lr){5-7}\cmidrule(lr){8-10}\cmidrule(lr){11-13}
\textbf{\sffamily Family} & d0 & d1 & d2 & d0 & d1 & d2 & d0 & d1 & d2 & d0 & d1 & d2 \\
\midrule
circuit fault & 0.98 & 0.98 & 0.91 & 0.11 & 0.12 & 0.05 & 0.02 & 0.00 & 0.00 & \multicolumn{3}{c}{--} \\
impedance topology & 0.94 & 0.97 & 0.98 & 0.11 & 0.11 & 0.05 & 0.00 & 0.00 & 0.00 & \multicolumn{3}{c}{--} \\
heat inclusion & 0.95 & 0.92 & 0.92 & 0.02 & 0.02 & 0.00 & 0.00 & 0.00 & 0.00 & 0.58 & 0.52 & 0.47 \\
hidden law & 0.95 & 0.98 & 0.89 & 0.03 & 0.06 & 0.05 & 0.00 & 0.00 & 0.00 & \multicolumn{3}{c}{--} \\
control plant & 1.00 & 0.97 & 0.91 & 0.02 & 0.00 & 0.03 & 0.00 & 0.00 & 0.00 & \multicolumn{3}{c}{--} \\
reaction network & 0.95 & 0.91 & 0.91 & 0.00 & 0.03 & 0.02 & 0.02 & 0.09 & 0.03 & 0.56 & 0.45 & 0.50 \\
mechanism discrimination$^\dagger$ & 1.00 & 0.98 & 0.97 & 0.16 & 0.14 & 0.09 & 0.23 & 0.23 & 0.22 & 0.98 & 1.00 & 1.00 \\
pathway fault & 1.00 & 1.00 & 1.00 & 0.06 & 0.03 & 0.06 & 0.11 & 0.06 & 0.02 & \multicolumn{3}{c}{--} \\
knockout network & 1.00 & 1.00 & 1.00 & 0.14 & 0.03 & 0.05 & 0.00 & 0.00 & 0.00 & 0.17 & 0.05 & 0.06 \\
group testing & 1.00 & 1.00 & 1.00 & 0.19 & 0.08 & 0.03 & 0.02 & 0.03 & 0.03 & \multicolumn{3}{c}{--} \\
counterfeit coin$^\dagger$ & 1.00 & 1.00 & 1.00 & 0.03 & 0.02 & 0.02 & 0.02 & 0.02 & 0.02 & \multicolumn{3}{c}{--} \\
rule induction & 1.00 & 1.00 & 0.89 & 0.02 & 0.00 & 0.00 & 0.00 & 0.00 & 0.00 & 0.28 & 0.25 & 0.02 \\
causal intervention & 0.98 & 1.00 & 0.94 & 0.00 & 0.00 & 0.00 & 0.00 & 0.05 & 0.02 & \multicolumn{3}{c}{--} \\
trace debugging & 1.00 & 1.00 & 1.00 & 0.05 & 0.06 & 0.03 & 0.00 & 0.00 & 0.00 & \multicolumn{3}{c}{--} \\
automaton learning$^\dagger$ & 1.00 & 1.00 & 0.97 & 0.00 & 0.02 & 0.00 & 0.00 & 0.00 & 0.00 & \multicolumn{3}{c}{--} \\
\bottomrule
\end{tabular*}
\end{table}

\begin{table}[H]
\centering
\caption{\capt{Calibration of the two MicroWorld families} on 64 development instances per level of their six-level ladders. Policies as in \cref{tab:calibration}; exact execution defines no exhaustive design.}
\label{tab:calibration-micro}
\scriptsize
\renewcommand{\arraystretch}{0.95}
\setlength{\tabcolsep}{4pt}
\begin{tabular*}{0.72\linewidth}{@{\extracolsep{\fill}}ll cccccc@{}}
\toprule
\textbf{\sffamily Family} & \textbf{\sffamily Policy} & d0 & d1 & d2 & d3 & d4 & d5 \\
\midrule
certificate audit & oracle & 1.00 & 1.00 & 1.00 & 1.00 & 1.00 & 1.00 \\
 & naive & 0.02 & 0.09 & 0.19 & 0.12 & 0.03 & 0.00 \\
 & shortcut & 0.30 & 0.28 & 0.33 & 0.30 & 0.33 & 0.25 \\
 & exhaustive & 0.30 & 0.45 & 0.48 & 0.30 & 0.31 & 0.34 \\
\addlinespace
exact execution & oracle & 1.00 & 1.00 & 1.00 & 1.00 & 1.00 & 1.00 \\
 & naive & 0.00 & 0.00 & 0.00 & 0.00 & 0.00 & 0.00 \\
 & shortcut & 0.02 & 0.00 & 0.00 & 0.00 & 0.00 & 0.00 \\
\bottomrule
\end{tabular*}
\end{table}

\begin{table}[H]
\centering
\caption{\capt{Calibration of the PhysicsWorld (d0--d3) and ChemistryWorld (d0--d2) families} on the first of two blocks of 64 development instances per level; every cell also passes on the second block. $^\dagger$ held out of all training.}
\label{tab:calibration-new}
\scriptsize
\renewcommand{\arraystretch}{0.95}
\setlength{\tabcolsep}{2.0pt}
\begin{tabular*}{\linewidth}{@{\extracolsep{\fill}}l cccc cccc cccc cccc@{}}
\toprule
 & \multicolumn{4}{c}{\textbf{\sffamily Oracle}} & \multicolumn{4}{c}{\textbf{\sffamily Naive}} & \multicolumn{4}{c}{\textbf{\sffamily Shortcut}} & \multicolumn{4}{c}{\textbf{\sffamily Exhaustive}} \\
\cmidrule(lr){2-5}\cmidrule(lr){6-9}\cmidrule(lr){10-13}\cmidrule(lr){14-17}
\textbf{\sffamily Family} & d0 & d1 & d2 & d3 & d0 & d1 & d2 & d3 & d0 & d1 & d2 & d3 & d0 & d1 & d2 & d3 \\
\midrule
rigid-body identification & 1.00 & 0.94 & 0.89 & 0.83 & 0.14 & 0.11 & 0.17 & 0.14 & 0.20 & 0.16 & 0.28 & 0.16 & 0.94 & 0.62 & 0.66 & 0.56 \\
airfoil inference & 1.00 & 0.97 & 0.95 & 0.92 & 0.14 & 0.08 & 0.08 & 0.06 & 0.00 & 0.03 & 0.02 & 0.00 & 0.12 & 0.12 & 0.09 & 0.03 \\
optical bench & 0.95 & 0.84 & 0.91 & 0.80 & 0.14 & 0.05 & 0.08 & 0.03 & 0.00 & 0.00 & 0.00 & 0.00 & 0.16 & 0.19 & 0.27 & 0.16 \\
coupled oscillators & 1.00 & 0.98 & 0.97 & 0.91 & 0.00 & 0.00 & 0.00 & 0.00 & 0.00 & 0.00 & 0.00 & 0.00 & 0.00 & 0.06 & 0.16 & 0.00 \\
fluid network$^\dagger$ & 0.97 & 0.94 & 0.88 & 0.86 & 0.14 & 0.14 & 0.17 & 0.09 & 0.30 & 0.22 & 0.22 & 0.19 & 0.62 & 0.47 & 0.45 & 0.45 \\
\addlinespace
structure elucidation & 1.00 & 1.00 & 0.97 & -- & 0.00 & 0.00 & 0.00 & -- & 0.02 & 0.00 & 0.00 & -- & 0.91 & 0.80 & 0.64 & -- \\
titration equilibria & 0.94 & 0.86 & 0.84 & -- & 0.12 & 0.16 & 0.11 & -- & 0.05 & 0.02 & 0.00 & -- & 0.45 & 0.34 & 0.22 & -- \\
kinetics process design & 0.91 & 0.86 & 0.94 & -- & 0.00 & 0.00 & 0.02 & -- & 0.06 & 0.05 & 0.03 & -- & 0.75 & 0.58 & 0.61 & -- \\
chromatography method & 0.94 & 0.92 & 0.88 & -- & 0.02 & 0.05 & 0.00 & -- & 0.09 & 0.03 & 0.00 & -- & 0.92 & 0.88 & 0.75 & -- \\
electrochemical mechanism & 1.00 & 0.89 & 0.83 & -- & 0.02 & 0.12 & 0.14 & -- & 0.09 & 0.20 & 0.17 & -- & 1.00 & 0.86 & 0.69 & -- \\
\bottomrule
\end{tabular*}
\end{table}

\begin{table}[H]
\centering
\caption{\capt{Admission of the frontier rungs} on two blocks of 64 development instances per rung. For each rung: the oracle's lower block (O), the naive policy's higher block (N) and the strongest gated shortcut's higher block (S). Gated shortcuts include the next easier rung's strategy on H rungs, the stage-by-stage policy on L rungs and the oracle without notes on L3. Reported policies are not gated and not shown: they apply the oracle's exact analysis to a simpler design and are deductive, except one control-plant habit described in \cref{sec:pipeline}. A dash marks a rung that is not admitted. $^\dagger$ held out of all training.}
\label{tab:frontier-gate}
\scriptsize
\renewcommand{\arraystretch}{0.95}
\setlength{\tabcolsep}{1.7pt}
\begin{tabular*}{\linewidth}{@{\extracolsep{\fill}}l ccc ccc ccc ccc ccc ccc@{}}
\toprule
 & \multicolumn{3}{c}{\textbf{\sffamily H1}} & \multicolumn{3}{c}{\textbf{\sffamily H2}} & \multicolumn{3}{c}{\textbf{\sffamily H3}} & \multicolumn{3}{c}{\textbf{\sffamily L1}} & \multicolumn{3}{c}{\textbf{\sffamily L2}} & \multicolumn{3}{c}{\textbf{\sffamily L3}} \\
\cmidrule(lr){2-4}\cmidrule(lr){5-7}\cmidrule(lr){8-10}\cmidrule(lr){11-13}\cmidrule(lr){14-16}\cmidrule(lr){17-19}
\textbf{\sffamily Family} & O & N & S & O & N & S & O & N & S & O & N & S & O & N & S & O & N & S \\
\midrule
circuit fault & 0.91 & 0.00 & 0.17 & 0.84 & 0.00 & 0.08 & 0.86 & 0.00 & 0.20 & 0.83 & 0.00 & 0.27 & 0.81 & 0.00 & 0.12 & 0.89 & 0.00 & 0.12 \\
impedance topology & 0.84 & 0.12 & 0.31 & 0.86 & 0.05 & 0.12 & 0.89 & 0.09 & 0.23 & 0.89 & 0.08 & 0.00 & 0.78 & 0.00 & 0.02 & 0.86 & 0.00 & 0.03 \\
heat inclusion & 0.89 & 0.00 & 0.19 & 0.83 & 0.00 & 0.17 & 0.81 & 0.00 & 0.06 & 0.81 & 0.00 & 0.16 & 0.78 & 0.00 & 0.03 & 0.92 & 0.00 & 0.27 \\
hidden law & 0.88 & 0.00 & 0.05 & 0.84 & 0.00 & 0.02 & 0.81 & 0.00 & 0.23 & 0.92 & 0.00 & 0.12 & 0.89 & 0.00 & 0.16 & 0.84 & 0.00 & 0.02 \\
control plant & 0.83 & 0.00 & 0.08 & 0.88 & 0.00 & 0.00 & 0.92 & 0.00 & 0.00 & 0.94 & 0.00 & 0.27 & 0.86 & 0.00 & 0.27 & 0.86 & 0.00 & 0.22 \\
reaction network & 0.80 & 0.02 & 0.17 & 0.84 & 0.00 & 0.00 & 0.84 & 0.03 & 0.25 & 0.86 & 0.00 & 0.20 & 0.86 & 0.00 & 0.08 & 0.81 & 0.00 & 0.00 \\
mechanism discrimination$^\dagger$ & 0.83 & 0.09 & 0.30 & 0.78 & 0.00 & 0.28 & 0.84 & 0.00 & 0.00 & 0.86 & 0.00 & 0.23 & 0.84 & 0.00 & 0.19 & 0.83 & 0.00 & 0.03 \\
pathway fault & -- & -- & -- & -- & -- & -- & -- & -- & -- & 0.88 & 0.05 & 0.23 & -- & -- & -- & 0.88 & 0.00 & 0.02 \\
knockout network & 1.00 & 0.02 & 0.12 & 1.00 & 0.00 & 0.05 & 1.00 & 0.00 & 0.03 & 1.00 & 0.00 & 0.05 & 1.00 & 0.00 & 0.02 & 1.00 & 0.00 & 0.00 \\
group testing & 0.92 & 0.03 & 0.06 & 0.97 & 0.00 & 0.09 & 0.88 & 0.00 & 0.08 & 1.00 & 0.02 & 0.05 & 1.00 & 0.00 & 0.05 & 0.97 & 0.00 & 0.03 \\
counterfeit coin$^\dagger$ & 1.00 & 0.14 & 0.09 & 0.97 & 0.11 & 0.11 & 1.00 & 0.08 & 0.12 & 1.00 & 0.00 & 0.27 & 1.00 & 0.00 & 0.22 & 1.00 & 0.00 & 0.09 \\
rule induction & 0.91 & 0.00 & 0.19 & 0.89 & 0.03 & 0.14 & 0.84 & 0.03 & 0.20 & 1.00 & 0.08 & 0.02 & 1.00 & 0.00 & 0.00 & 1.00 & 0.00 & 0.00 \\
causal intervention & 0.94 & 0.00 & 0.00 & 0.92 & 0.00 & 0.08 & 0.92 & 0.00 & 0.08 & 0.91 & 0.00 & 0.00 & 0.95 & 0.00 & 0.12 & 0.86 & 0.00 & 0.09 \\
trace debugging & 1.00 & 0.00 & 0.09 & 1.00 & 0.02 & 0.03 & 1.00 & 0.00 & 0.05 & 1.00 & 0.00 & 0.19 & 1.00 & 0.00 & 0.02 & 1.00 & 0.00 & 0.02 \\
automaton learning$^\dagger$ & 0.84 & 0.00 & 0.28 & 0.88 & 0.00 & 0.00 & 0.88 & 0.00 & 0.00 & 0.91 & 0.00 & 0.02 & 0.92 & 0.00 & 0.00 & 0.94 & 0.00 & 0.00 \\
\bottomrule
\end{tabular*}
\end{table}

\paragraph{Robustness of the calibration.}
\label{app:robustness}
A second audit re-runs the calibration, determinism and shortcut checks on the distributed build of the suite rather than the research code.
Replaying the original calibration instances reproduces 88 of the 90 oracle and naive entries of the fifteen discovery worlds exactly, and the other two differ by a single instance in a simulator-backed family, through floating-point differences between processor architectures.
On a separate development split, sixteen families pass every gate at every audited level, 88 family levels in all with entry and harder levels included; 50 of those were run on both macOS arm64 and Linux x86\_64 and pass on both, and every update replays each previously audited level byte for byte.
On the ten PhysicsWorld and ChemistryWorld families the audit passes 34 of 35 levels; the exception is a rigid-body level whose naive policy reached 17 of 64 on that split, one episode above the gate, and which passes on both calibration blocks.
The frontier rungs are admitted on the machine that plays their final instances.

\paragraph{Simulators and determinism.}
Most families compute their readings exactly: nodal analysis of resistor networks, complex impedance, closed-form kinetics checked against a numerical ODE solver, an exact discretisation of a control loop with dead time, Boolean reachability in signalling networks, charge-balance equilibria, modal analysis, and interpreters for programs and machines.
Heat inclusion's finite-volume solver agrees with a finite-element reference (scikit-fem; \citealp{gustafsson2020scikitfem}) to within \heatMaxMK{}\,mK over all \heatCandidates{} candidate inclusions, under a quarter of the noise at every core level (at least \heatNoiseMinMK{}\,mK), and the rigid-body integrator agrees with MuJoCo \citep{todorov2012mujoco} to within $\rigidMaxDeg^\circ$ on \rigidReleases{} fixed release comparisons.
The hydraulic solver implements the global-gradient method of EPANET \citep{rossman2000epanet}; pathway fault simulates curated BioModels kinetic models \citep{malik2020biomodels} with libRoadRunner \citep{welsh2023libroadrunner}; and structure elucidation draws from an isomer bank enumerated with RDKit \citep{rdkit}.
External simulators run out of process in pinned environments, all randomness is seeded, and every admission run records a replay digest over the truth, every prompt, every oracle action and the return.

\paragraph{Contamination controls.}
Every distributed data file carries a canary string, every observation an instance hash, and a deduplication tool checks a training corpus against a split.
The exact-execution generator rejects every program whose structure matches a protected program-execution benchmark; the protected list exists only as keyed digests under a key that is never distributed, so it cannot be enumerated from the code, and a copy of the generator without the key draws from a different instance space.
Comparing 3,000 training truths with 64 development truths per level finds no repeated truth in 22 families.
Four families draw their truths from a finite bank by construction and show one opening observation per level, so what can be learned there is the mapping from observations to answers; in pathway fault, the published kinetic models recur across splits, so its splits separate instances, not models.

\section{Evaluation settings}
\label{app:eval}

\paragraph{Local models.}
Every evaluation of \ours{}, \nar{}, \fin{} and \qbig{} uses thinking on, a 4,096-token thinking cap per decision with a forced close, temperature 1.0, top-$p$ 0.95 and top-$k$ 20, under the protocol of \cref{sec:episode}.
After a forced close, the answer budget is just large enough for the answer, so that the model cannot keep reasoning past the thinking cap.

\paragraph{API model.}
\flash{} is the \snapshot{}, served under the \texttt{deepseek-flash} API name since 17 October 2026; every run reported here used it with thinking on, and the default-effort suite runs used a limit of 16,384 tokens per call. The maximum-effort comparison of \cref{tab:flashmax} used reasoning effort ``max'' with up to 393,216 tokens per call and no re-ask (\cref{tab:settings}).
On the core tier, including PhysicsWorld and ChemistryWorld, a turn cut at the limit returns no action, and two such turns in a row end the episode, as for any two replies without a legal action.
On the frontier tier a cut turn gets one re-ask of the same messages with thinking off and the episode's answer budget, asking for the action line only, and the reply counts as the turn's action; the re-ask carries the cut turn's reasoning, up to its last 90,000 characters.
Its runs on the laboratory worlds of \cref{sec:labrl} use the same API name with the effort settings and caps given in \cref{app:labrl}, not the 16,384-token limit above.
Every response is logged with its exact messages, so every episode can be replayed without new calls.

\paragraph{Pilot model.}
The pilot (\solModel{}) plays the instance with ordinal $o = 0$ of every family and frontier rung, and of every core level of the discovery worlds, PhysicsWorld and ChemistryWorld.
Each decision is one stateless call that carries exactly the messages of the API protocol; the observation already contains the history (on L3, its window and notebook), every tool is disabled except the collaboration (sub-agent) tools that this CLI build offers regardless, which no stored call record shows in use, and the model's final message is passed to the action parser unchanged.
The output length is not ours to cap, so the frontier re-ask does not apply.
The pilot runs the same evaluation code on a second machine of a different processor architecture. A replay feeds a recorded episode's actions to the environment again; its outcome matches when it gives the same return (and, for Flash, the same ending). On that machine, \replayFlashSame{} of \replayFlashN{} replayed Flash episodes and \replayQSame{} of \replayQN{} replayed 27B episodes match (\replayFlashBytes{} of Flash's are byte-identical at every turn; the others differ only after a model-written Python run, whose printed output differs in its last digits across architectures). The one that does not match is a pathway-fault L1 instance that draws a different hidden model, as the hosted-service paragraph below describes for that family, and is excluded from every pilot comparison. Two rigid-body instances show small numerical differences in some joint readings and are kept.
The oracle results in \cref{tab:pilot} were played on that machine.

\paragraph{Hosted service.}
The service plays the core tier's final instances under the protocol of \cref{sec:episode}, so a model run there with the settings above is directly comparable to \cref{tab:results,tab:baselines}.
One family, pathway fault, relies on a numerical library whose floating-point results differ across processor architectures; the service runs on x86\_64, like the evaluation here, and its scripted-policy rates match.
Generator internals and the composition of the training mixture are withheld.

\paragraph{GPQA Diamond.}
\label{app:gpqa}
All 198 questions with identical presented options, question order, per-question seeds and native thinking-mode prompt for every model; temperature 1.0, top-$p$ 0.95, top-$k$ 20, presence penalty 1.5; a 32,768-token reasoning allowance followed, when needed, by a forced close with at most 512 answer tokens; one response per question per model, paired by question and scored on the last boxed answer (\cref{tab:gpqa}).

\begin{table}[H]
\centering
\caption{\capt{GPQA Diamond,} all 198 questions, one response per question per model with identical prompts, options, question order and seeds, scored on the last boxed answer. Discordant: questions answered correctly by only this model / only \ours{}.}
\label{tab:gpqa}
\footnotesize
\setlength{\tabcolsep}{8pt}
\begin{tabular}{@{}lccc@{}}
\toprule
 & \textbf{\sffamily \ours{}} & \textbf{\sffamily \nar{}} & \textbf{\sffamily \fin{}} \\
\midrule
correct & 155 & 157 & 156 \\
discordant with \ours{} & -- & 14 / 12 & 9 / 8 \\
median generated tokens & 11,349 & 11,035 & 11,094 \\
\bottomrule
\end{tabular}
\end{table}

\paragraph{Family-level bootstrap.}
\label{app:cluster}
Main-text intervals resample paired episodes, treating those of one family as independent.
\Cref{tab:cluster} repeats each headline comparison with a bootstrap that resamples whole families with all their paired episodes, computed exactly by enumerating every resample (no Monte Carlo error); with all $m^m$ ordered draws of the $m$ families equally likely, its bounds are the smallest means with cumulative probability above 0.025 and at least 0.975.
The family-level intervals are at least as wide, and all stay resolved except Flash minus the 27B on L3 alone and the held-out training gains of the 9B and the 27B, which rest on three families; automaton learning does not change, so a resample that draws only that family shows no gain.
On the pilot's 84 frontier instances, one per family and rung in 14 families, every pilot row stays resolved except the pilot model's gap to the oracle over all rungs, \solMinusOracleFam{}; on the long-horizon rungs that gap stays resolved, \solLongMinusOracleFam{}.
\enlargethispage{\baselineskip}

\begin{table}[H]
\centering
\caption{\capt{Paired differences with episode-level and family-level bootstrap intervals} (Flash: \flash{}, \snapshot{}; Sol: \solname{}, the pilot model; 27B: \qbig{}). Episode bootstrap: 10,000 resamples of paired episodes; family bootstrap: whole families drawn with replacement, each with all of its paired episodes, computed exactly by enumerating every resample (no Monte Carlo error); both are 95\% percentile intervals. Families: the number of families resampled; episodes: the paired episodes they contain. Pilot rows: the pilot's frontier instances (\cref{tab:pilot}), with \flash{} at its default effort under the 16,384-token cap and at its maximum effort (\cref{tab:flashmax}).}
\label{tab:cluster}
\footnotesize
\setlength{\tabcolsep}{5pt}
\begin{tabular}{@{}lrrccc@{}}
\toprule
\textbf{\sffamily Comparison} & \textbf{\sffamily Families} & \textbf{\sffamily Episodes} & \textbf{\sffamily Difference} & \textbf{\sffamily Episode bootstrap} & \textbf{\sffamily Family bootstrap} \\
\midrule
Core tier, \flash{} $-$ \qbig{} & 15 & 360 & +0.225 & [+0.169, +0.278] & [+0.111, +0.347] \\
Frontier tier, \flash{} $-$ \qbig{} & 15 & 680 & +0.249 & [+0.213, +0.284] & [+0.163, +0.335] \\
\quad harder rungs & 14 & 336 & +0.292 & [+0.241, +0.342] & [+0.167, +0.432] \\
\quad long-horizon rungs & 15 & 344 & +0.206 & [+0.160, +0.256] & [+0.119, +0.296] \\
\quad L3 only & 14 & 112 & +0.071 & [+0.009, +0.134] & [\ensuremath{-}0.009, +0.152] \\
Trained families, \fin{} $-$ \nar{} & 12 & 837 & +0.045 & [+0.019, +0.072] & [+0.017, +0.080] \\
Trained families, \fin{} $-$ \ours{} & 12 & 837 & +0.059 & [+0.031, +0.085] & [+0.019, +0.121] \\
Held-out families, \fin{} $-$ \ours{} & 3 & 216 & +0.083 & [+0.023, +0.144] & [+0.000, +0.153] \\
Held-out families, \fin{} $-$ \nar{} & 3 & 216 & +0.102 & [+0.051, +0.157] & [+0.000, +0.167] \\
Held-out families, 27B after two rounds $-$ base & 3 & 216 & +0.120 & [+0.046, +0.194] & [+0.000, +0.306] \\
Invalid endings, \fin{} $-$ \nar{} & 27 & 2,181 & \ensuremath{-}0.043 & [\ensuremath{-}0.054, \ensuremath{-}0.032] & [\ensuremath{-}0.071, \ensuremath{-}0.020] \\
Invalid endings, \fin{} $-$ \ours{} & 27 & 2,181 & \ensuremath{-}0.074 & [\ensuremath{-}0.087, \ensuremath{-}0.061] & [\ensuremath{-}0.119, \ensuremath{-}0.035] \\
\addlinespace
Pilot, Flash at max $-$ default effort & 14 & 84 & +0.298 & [+0.191, +0.405] & [+0.155, +0.440] \\
Pilot, Flash at max effort $-$ oracle & 14 & 84 & \ensuremath{-}0.262 & [\ensuremath{-}0.357, \ensuremath{-}0.179] & [\ensuremath{-}0.417, \ensuremath{-}0.143] \\
Pilot, Sol $-$ Flash at default effort & 14 & 84 & +0.464 & [+0.357, +0.571] & [+0.310, +0.619] \\
Pilot, Sol $-$ Flash at max effort & 14 & 84 & +0.167 & [+0.071, +0.262] & [+0.060, +0.286] \\
Pilot, Sol $-$ oracle & 14 & 84 & \ensuremath{-}0.095 & [\ensuremath{-}0.179, \ensuremath{-}0.012] & [\ensuremath{-}0.190, +0.000] \\
\quad long-horizon rungs & 14 & 42 & \ensuremath{-}0.167 & [\ensuremath{-}0.286, \ensuremath{-}0.071] & [\ensuremath{-}0.286, \ensuremath{-}0.071] \\
\bottomrule
\end{tabular}
\end{table}

\section{Training details}
\label{app:training}

\begin{table}[H]
\centering
\caption{\capt{The two training phases.} Kept episodes are successful episodes per round, pooled over all trained families.}
\label{tab:schedule}
\footnotesize
\setlength{\tabcolsep}{5pt}
\begin{tabular}{@{}lcccccc@{}}
\toprule
\textbf{\sffamily Phase} & \textbf{\sffamily Rounds} & \textbf{\sffamily Families} & \textbf{\sffamily Own episodes kept} & \textbf{\sffamily Teacher episodes kept} & \textbf{\sffamily Updates} & \textbf{\sffamily Target tokens} \\
\midrule
narrow & 3 & 12 & 48, 47, 84 & -- & 41 & 3.1M \\
broad & 4 & 23 & 333, 423, 468, 475 & 81, 45, 24, 17 & 323 & 25.7M \\
\bottomrule
\end{tabular}
\end{table}

\paragraph{Optimisation.}
Both phases fine-tune all parameters for one epoch per round on every decision of the kept episodes (\cref{tab:schedule}), at a learning rate of order $10^{-6}$ with batches of a few dozen sequences (exact values withheld).
Rollouts use a 4,096-token thinking cap per decision; the narrow phase samples at temperature 1.0 without top-$p$/top-$k$ truncation, and the broad phase samples like the evaluation.
In the broad phase's training rollouts, pooled over all trained families, invalid endings fell from 0.118 in the first round to 0.026, 0.020 and 0.017.

\paragraph{Pre-specified criteria.}
The four criteria of \cref{sec:method} were fixed before the broad phase's evaluation: (i) fewer invalid endings than \nar{} over all final instances; (ii) higher success than \nar{} on the twelve trained discovery worlds, with a paired lower bound above zero; (iii) the same on the six families where \nar{} succeeded on fewer than 0.10; and (iv) GPQA Diamond within eight questions of \ours{}, with median length at least 70\% of \ours{}'s.
All four pass: (i) \invBN{}; (ii) \trBN{}; (iii) \weakBN{}, mostly control plant; (iv) 156 against 155.
Because criterion (iii) selects families on the same instances it tests, part of that gain may be regression to the mean.

\paragraph{Entry levels.}
In the last round's training rollouts, knockout network succeeded on about half of its entry-level episodes and heat inclusion on about a third, with at most two successes at any other level.
Control plant had the only resolved gain among the weak families; it first rose on an upper level of its training ladder, and its final success rose from 0 for \nar{} to 0.097 for \fin{}, with seven episodes gained and none lost.

\paragraph{Reinforcement learning on laboratory worlds.}
\label{app:labrl}
\Cref{tab:labrlsettings} gives the settings of both runs.
Before each step the trainer recomputes the log-probability of every sampled token and rejects the batch if it differs from the sampler's beyond fixed tolerances; a token whose two log-probabilities differ by more than \lrMaskNats{} nats is dropped from the loss, which happened to \lrMaskedTokens{} tokens in all, both in \mixrl{}.
Every update of both runs committed, no episode was excluded, and gradient clipping did not bind in any update.

\begin{table}[H]
\centering
\caption{\capt{Settings of the two reinforcement-learning runs} of \cref{sec:labrl}, which differ only in their training worlds.}
\label{tab:labrlsettings}
\footnotesize
\setlength{\tabcolsep}{5pt}
\begin{tabular}{@{}L{3.3cm}L{9.6cm}@{}}
\toprule
\textbf{\sffamily Setting} & \textbf{\sffamily Value} \\
\midrule
Starting model & \warm{}: \ours{} after a short supervised warm-up on its own turns in training replication worlds \\
Policy update & advantage = reward $-$ mean reward of the episode's group; clipped importance ratio, $\epsilon = \lrClip$; dual clip at $\lrDualClip$ for negative advantages; per-token penalty on the divergence from \warm{}, weight \lrRefCoef{} \\
Reward & binary and terminal, from the family's verifier; no length shaping \\
Group size & \lrG{} episodes per world \\
Worlds per update & \lrWorldsPerUpdate{}, none used twice; \labrl{}: replication only; \mixrl{}: \lrMcLabPerUpdate{} of the same replication worlds and \lrMcMcPerUpdate{} method comparison \\
Episodes & \lrEpisodesPerUpdate{} per update, \lrEpisodesRun{} per run \\
Updates & \lrUpdates{}, one optimiser step each, on all parameters \\
Optimiser & Adam, $(\beta_1, \beta_2) = (\lrAdamBetaOne, \lrAdamBetaTwo)$; learning rate $\lrLR$ after a linear warm-up over \lrWarmupUpdates{} updates; no weight decay; gradient norm clipped at \lrGradClip{} \\
Sampling & temperature 1.0, no top-$p$ or top-$k$ truncation; at most \lrTurnCap{} tokens per turn and \lrEpisodeCap{} generated tokens per episode; context \lrContext{} tokens \\
Loss normalisation & sum over the batch's sampled tokens, divided by \lrEpisodesPerUpdate{} $\times$ \lrEpisodeCap{} \\
Hardware and time & one B200 GPU; \lrHoursLab{} h for \labrl{}, \lrHoursMix{} h for \mixrl{} \\
\bottomrule
\end{tabular}
\end{table}

\paragraph{The held-out panel.}
The panel's worlds come from each family's development split, one world per scenario, drawn at random within each stratum with a fixed seed.
We checked that they are disjoint from training by world and scenario identifiers, by the study each world derives from, by a hash of its full public task content (brief and data tables) and, for method comparison, by its hidden analyser: none of the \lrPanelWorlds{} panel worlds matches any of the \lrPoolWorlds{} training-split worlds, which hold the full training pools of both families, every world drawn for either run (\lrTrainRepWorldsLab{} replication worlds for \labrl{}; \lrTrainRepWorldsMix{} replication and \lrMcWorldsMix{} method-comparison worlds for \mixrl{}) and the \lrWarmWorlds{} worlds behind the warm-up's rows.
Every system plays each panel world with at most \lrTurnCap{} tokens per turn (\lrPanelTurnCapMc{} in method comparison, as the panel's fixed configuration sets for every system), \lrEpisodeCap{} generated tokens per episode, and sampling as in training.

\paragraph{Flash and the broader evaluations.}
On the panel, \flash{} runs under the API name of \cref{app:eval} with thinking on, at reasoning effort ``low'' (the reference setting) or ``high'' (the default).
It receives the same system and user prompts, tool schema, statistics engine, budgets, parser and verifier as the local models, and its output in a turn is cut at the panel's per-turn cap, counted in the local models' tokenizer.
On P+X and the workflow set it runs at its default effort with up to \lrFlashMaxTokens{} output tokens per call; the 9B models run there with thinking on, temperature \lrEvalTemp{}, top-$p$ \lrEvalTopP{} and at most \lrEvalMaxTokens{} tokens per call.
The registration reports P+X both with and without its power-calculation items, which need a pooled standard deviation that no evaluation tool reports: the \lrPowerSteps{} power steps of the projects and the \lrPowerBins{} sample-size decisions.
Without them, a project scores 1 when all of its other steps pass and the X suite keeps its other items (\lrPXnpItems{} items in all); the macro score is \lrPXnpStart{} for \warm{}, \lrPXnpLabTwelve{} and \lrPXnpLab{} for \labrl{} after 12 and 16 updates and \lrPXnpMixTwelve{} and \lrPXnpMix{} for \mixrl{}, and the differences from \warm{} are \lrPXnpLabTwelveMinusStart{}, \lrPXnpLabMinusStart{}, \lrPXnpMixTwelveMinusStart{} and \lrPXnpMixMinusStart{}, against \lrPXLabTwelveMinusStart{}, \lrPXLabMinusStart{}, \lrPXMixTwelveMinusStart{} and \lrPXMixMinusStart{} with them.
The workflow set has \lrWFItems{} workflows with \lrWFSteps{} steps from \lrWFClusters{} source experiments; a step passes when its decisions are right, its values match the engine's analysis and its numbers come from a tool result or the materials, and a workflow scores its share of passed steps.

\section{Per-family results}
\label{app:baselines}

\begin{table}[H]
\centering
\caption{\capt{The core tier on all 1,080 final instances of the discovery worlds} (24 per level, 72 per family); the scripted policies played the same instances. Invalid: share of episodes ending on replies without a legal action. Headroom: oracle minus \qbig{}. Rows sorted by headroom. $^\dagger$ held out of all training.}
\label{tab:baselines}
\footnotesize
\setlength{\tabcolsep}{4.5pt}
\begin{tabular}{@{}lccccccc@{}}
\toprule
 & \multicolumn{2}{c}{\textbf{\sffamily \ours{}}} & \multicolumn{2}{c}{\textbf{\sffamily \qbig{}}} & & & \\
\cmidrule(lr){2-3}\cmidrule(lr){4-5}
\textbf{\sffamily Family} & success & invalid & success & invalid & \textbf{\sffamily Oracle} & \textbf{\sffamily Naive} & \textbf{\sffamily Headroom} \\
\midrule
heat inclusion & 0.000 & 0.069 & 0.000 & 0.000 & 0.917 & 0.014 & 0.917 \\
rule induction & 0.000 & 0.139 & 0.028 & 0.000 & 0.917 & 0.000 & 0.889 \\
impedance topology & 0.000 & 0.028 & 0.125 & 0.014 & 0.958 & 0.111 & 0.833 \\
pathway fault & 0.097 & 0.014 & 0.167 & 0.014 & 0.986 & 0.014 & 0.819 \\
automaton learning$^\dagger$ & 0.000 & 0.000 & 0.194 & 0.125 & 1.000 & 0.000 & 0.806 \\
knockout network & 0.000 & 0.097 & 0.250 & 0.000 & 1.000 & 0.097 & 0.750 \\
control plant & 0.083 & 0.000 & 0.250 & 0.000 & 0.986 & 0.000 & 0.736 \\
counterfeit coin$^\dagger$ & 0.097 & 0.292 & 0.333 & 0.000 & 1.000 & 0.069 & 0.667 \\
circuit fault & 0.097 & 0.028 & 0.389 & 0.000 & 0.972 & 0.167 & 0.583 \\
reaction network & 0.097 & 0.000 & 0.375 & 0.028 & 0.931 & 0.014 & 0.556 \\
causal intervention & 0.111 & 0.000 & 0.403 & 0.000 & 0.958 & 0.000 & 0.556 \\
trace debugging & 0.167 & 0.347 & 0.500 & 0.000 & 1.000 & 0.028 & 0.500 \\
mechanism discrimination$^\dagger$ & 0.153 & 0.000 & 0.653 & 0.000 & 0.986 & 0.069 & 0.333 \\
hidden law & 0.181 & 0.014 & 0.806 & 0.000 & 0.944 & 0.028 & 0.139 \\
group testing & 0.347 & 0.514 & 1.000 & 0.000 & 1.000 & 0.250 & 0.000 \\
\midrule
twelve trained families & 0.098 & 0.104 & 0.358 & 0.005 & 0.964 & 0.060 & 0.606 \\
three held-out families$^\dagger$ & 0.083 & 0.097 & 0.394 & 0.042 & 0.995 & 0.046 & 0.602 \\
\bottomrule
\end{tabular}
\end{table}

\begin{table}[H]
\centering
\caption{\capt{The core tier outside the discovery worlds,} on all final instances (24 per family and level), except Flash. Flash: \flash{} (\snapshot{}) in the core tier's setting (thinking on, 16,384 tokens per call, no re-ask), on the instances in column $n_F$: the first \fpPhysRounds{} final instances of each PhysicsWorld level and all 24 of each ChemistryWorld level. Paired with the other systems on the same instances, Flash minus the oracle is \fpPhysMinusOracle{} on PhysicsWorld and \fpChemMinusOracle{} on ChemistryWorld; minus \qbig{}, \fpPhysMinusQ{} and \fpChemMinusQ{}; minus \ours{}, \fpPhysMinusConcise{} and \fpChemMinusConcise{}. Oracle: the gate's scripted oracle on the same final instances of the PhysicsWorld and ChemistryWorld families, pooled over their levels (\oraclePhysFinal{} and \oracleChemFinal{} per group); it was not played on the MicroWorld final instances. $^\dagger$ held out of all training; $^\ddagger$ includes the two evaluation-only levels.}
\label{tab:othergroups}
\footnotesize
\setlength{\tabcolsep}{6pt}
\begin{tabular}{@{}lcccccccc@{}}
\toprule
\textbf{\sffamily Family} & \textbf{\sffamily $n$} & \textbf{\sffamily \ours{}} & \textbf{\sffamily \nar{}} & \textbf{\sffamily \fin{}} & \textbf{\sffamily \qbig{}} & \textbf{\sffamily $n_F$} & \textbf{\sffamily Flash} & \textbf{\sffamily Oracle} \\
\midrule
rigid-body identification & 96 & 0.010 & 0.000 & 0.010 & 0.021 & 64 & 0.203 & 0.906 \\
airfoil inference & 96 & 0.021 & 0.000 & 0.010 & 0.052 & 64 & 0.250 & 0.990 \\
optical bench & 96 & 0.010 & 0.021 & 0.010 & 0.021 & 64 & 0.078 & 0.802 \\
coupled oscillators & 96 & 0.000 & 0.000 & 0.000 & 0.000 & 64 & 0.031 & 0.958 \\
fluid network$^\dagger$ & 96 & 0.010 & 0.010 & 0.000 & 0.062 & 64 & 0.078 & 0.885 \\
\addlinespace
structure elucidation & 72 & 0.014 & 0.028 & 0.069 & 0.333 & 72 & 0.806 & 1.000 \\
titration equilibria & 72 & 0.069 & 0.000 & 0.028 & 0.125 & 72 & 0.167 & 0.931 \\
kinetics process design & 72 & 0.000 & 0.014 & 0.028 & 0.014 & 72 & 0.014 & 0.931 \\
chromatography method & 72 & 0.014 & 0.014 & 0.000 & 0.056 & 72 & 0.292 & 0.944 \\
electrochemical mechanism & 72 & 0.056 & 0.069 & 0.153 & 0.347 & 72 & 0.347 & 0.944 \\
\addlinespace
certificate audit & 144 & 0.049 & 0.132 & 0.188 & 0.764 & -- & -- & -- \\
exact execution$^\ddagger$ & 144 & 0.618 & 0.562 & 0.611 & 0.743 & -- & -- & -- \\
\bottomrule
\end{tabular}
\end{table}

\label{app:frontier}
\begin{table}[H]
\centering
\caption{\capt{The harder rungs per family:} success of \flash{} (F; \snapshot{}), \qbig{} (Q) and the oracle (O) on the final instances: F and O on 24 per family (\roundsHthree{} on H3), Q on 8. $^\dagger$ held out of all training.}
\label{tab:frontier-h}
\scriptsize
\renewcommand{\arraystretch}{0.95}
\setlength{\tabcolsep}{3pt}
\begin{tabular*}{0.86\linewidth}{@{\extracolsep{\fill}}l ccc ccc ccc@{}}
\toprule
 & \multicolumn{3}{c}{\textbf{\sffamily H1}} & \multicolumn{3}{c}{\textbf{\sffamily H2}} & \multicolumn{3}{c}{\textbf{\sffamily H3}} \\
\cmidrule(lr){2-4}\cmidrule(lr){5-7}\cmidrule(lr){8-10}
\textbf{\sffamily Family} & F & Q & O & F & Q & O & F & Q & O \\
\midrule
circuit fault & 0.17 & 0.00 & 0.96 & 0.17 & 0.00 & 0.88 & 0.21 & 0.00 & 0.84 \\
impedance topology & 0.25 & 0.00 & 0.92 & 0.00 & 0.00 & 0.96 & 0.00 & 0.00 & 0.89 \\
heat inclusion & 0.08 & 0.00 & 0.83 & 0.04 & 0.00 & 0.79 & 0.00 & 0.00 & 0.74 \\
hidden law & 0.46 & 0.00 & 0.96 & 0.21 & 0.00 & 0.92 & 0.16 & 0.00 & 0.89 \\
control plant & 0.58 & 0.00 & 0.71 & 0.71 & 0.00 & 0.88 & 0.47 & 0.00 & 1.00 \\
reaction network & 0.71 & 0.38 & 0.83 & 0.46 & 0.00 & 0.96 & 0.84 & 0.50 & 1.00 \\
mechanism discrimination$^\dagger$ & 0.62 & 0.12 & 0.92 & 0.79 & 0.12 & 0.79 & 0.42 & 0.12 & 0.79 \\
pathway fault & -- & -- & -- & -- & -- & -- & -- & -- & -- \\
knockout network & 1.00 & 0.25 & 1.00 & 0.92 & 0.00 & 1.00 & 0.84 & 0.00 & 1.00 \\
group testing & 0.92 & 0.88 & 0.96 & 0.58 & 0.12 & 1.00 & 0.37 & 0.12 & 0.95 \\
counterfeit coin$^\dagger$ & 0.88 & 0.25 & 1.00 & 0.42 & 0.25 & 1.00 & 0.37 & 0.00 & 1.00 \\
rule induction & 0.08 & 0.00 & 0.92 & 0.08 & 0.00 & 0.92 & 0.00 & 0.00 & 0.79 \\
causal intervention & 0.96 & 0.38 & 1.00 & 0.00 & 0.00 & 0.92 & 0.00 & 0.00 & 1.00 \\
trace debugging & 0.12 & 0.00 & 1.00 & 0.12 & 0.00 & 1.00 & 0.00 & 0.00 & 1.00 \\
automaton learning$^\dagger$ & 0.04 & 0.00 & 0.96 & 0.00 & 0.00 & 0.75 & 0.05 & 0.00 & 0.84 \\
\midrule
\textbf{pooled} & \textbf{0.49} & \textbf{0.16} & \textbf{0.93} & \textbf{0.32} & \textbf{0.04} & \textbf{0.91} & \textbf{0.27} & \textbf{0.05} & \textbf{0.91} \\
\bottomrule
\end{tabular*}
\end{table}

\begin{table}[H]
\centering
\caption{\capt{The long-horizon rungs per family,} as in \cref{tab:frontier-h}: \flash{} (F; \snapshot{}), \qbig{} (Q) and the oracle (O); F and O on 24 per family (18 on L3), Q on 8.}
\label{tab:frontier-l}
\scriptsize
\renewcommand{\arraystretch}{0.95}
\setlength{\tabcolsep}{3pt}
\begin{tabular*}{0.86\linewidth}{@{\extracolsep{\fill}}l ccc ccc ccc@{}}
\toprule
 & \multicolumn{3}{c}{\textbf{\sffamily L1}} & \multicolumn{3}{c}{\textbf{\sffamily L2}} & \multicolumn{3}{c}{\textbf{\sffamily L3}} \\
\cmidrule(lr){2-4}\cmidrule(lr){5-7}\cmidrule(lr){8-10}
\textbf{\sffamily Family} & F & Q & O & F & Q & O & F & Q & O \\
\midrule
circuit fault & 0.00 & 0.00 & 0.75 & 0.00 & 0.00 & 0.83 & 0.00 & 0.00 & 1.00 \\
impedance topology & 0.21 & 0.00 & 0.92 & 0.00 & 0.00 & 0.71 & 0.22 & 0.00 & 0.83 \\
heat inclusion & 0.04 & 0.00 & 0.96 & 0.00 & 0.00 & 0.83 & 0.00 & 0.00 & 0.89 \\
hidden law & 0.92 & 0.12 & 0.92 & 0.54 & 0.00 & 0.92 & 0.17 & 0.12 & 0.89 \\
control plant & 0.33 & 0.00 & 1.00 & 0.25 & 0.00 & 0.96 & 0.11 & 0.00 & 1.00 \\
reaction network & 0.75 & 0.25 & 0.96 & 0.71 & 0.00 & 0.83 & 0.11 & 0.00 & 0.83 \\
mechanism discrimination$^\dagger$ & 0.79 & 0.38 & 0.79 & 0.71 & 0.25 & 0.96 & 0.56 & 0.25 & 0.67 \\
pathway fault & 0.29 & 0.00 & 0.96 & -- & -- & -- & -- & -- & -- \\
knockout network & 0.12 & 0.00 & 1.00 & 0.00 & 0.00 & 1.00 & 0.00 & 0.00 & 1.00 \\
group testing & 1.00 & 1.00 & 1.00 & 0.96 & 1.00 & 1.00 & 0.06 & 0.25 & 0.94 \\
counterfeit coin$^\dagger$ & 0.79 & 0.00 & 1.00 & 0.54 & 0.25 & 1.00 & 0.06 & 0.00 & 1.00 \\
rule induction & 0.62 & 0.00 & 1.00 & 0.62 & 0.00 & 1.00 & 0.22 & 0.00 & 1.00 \\
causal intervention & 0.67 & 0.00 & 0.92 & 0.08 & 0.00 & 0.96 & 0.00 & 0.00 & 0.94 \\
trace debugging & 0.29 & 0.00 & 1.00 & 0.04 & 0.00 & 1.00 & 0.00 & 0.00 & 1.00 \\
automaton learning$^\dagger$ & 0.17 & 0.00 & 0.96 & 0.00 & 0.00 & 0.92 & 0.00 & 0.00 & 0.89 \\
\midrule
\textbf{pooled} & \textbf{0.47} & \textbf{0.12} & \textbf{0.94} & \textbf{0.32} & \textbf{0.11} & \textbf{0.92} & \textbf{0.11} & \textbf{0.04} & \textbf{0.92} \\
\bottomrule
\end{tabular*}
\end{table}

\end{document}